\documentclass[5p,number,sort&compress,times,final,nopreprintline]{elsarticle} 

\usepackage[utf8]{inputenc}
\usepackage[hyphens,spaces,obeyspaces]{url}
\usepackage{amsmath}
\usepackage[export]{adjustbox}
\usepackage{calc}

\usepackage{xcolor}
\newcommand\colorsquare[2][RGB]{%
	\textcolor[#1]{#2}{\rule{0.5\baselineskip}{0.5\baselineskip}}
} 

\usepackage{csquotes}

\newcommand{\beginappendix}{%
	\setcounter{table}{0}
	\renewcommand{\thetable}{A.\arabic{table}}%
	\setcounter{figure}{0}
	\renewcommand{\thefigure}{A.\arabic{figure}}%
	\setcounter{algorithm}{0}
	\renewcommand{\thealgorithm}{A.\arabic{algorithm}}%
}

\usepackage[binary-units]{siunitx}
\usepackage[super]{nth}
\newcommand{\CI}[3]{\SI{#1}{\percent} (\SI{95}{\percent} $\text{\glsentryshort{CI}} = \SIrange[range-phrase ={,}\,]{#2}{#3}{\percent}$)}
\newcommand{\CIpp}[3]{\SI{#1}{\pp} (\SI{95}{\percent} $\text{\glsentryshort{CI}} = \SIrange[range-phrase ={,}\,]{#2}{#3}{\pp}$)}

\usepackage[
	textsize=footnotesize,
	colorinlistoftodos,
	obeyFinal
	]{todonotes}
	
\presetkeys{todonotes}{inline,backgroundcolor=green!50!white}{}

\usepackage{nameref}
\makeatletter
\newcommand*{\thissectiontitle}{\@currentlabelname}
\makeatother

\usepackage[
	acronym,
	nomain,
	nogroupskip,
	symbols,
	nonumberlist,
	nopostdot,
	automake,
	style=super]{glossaries} 
\usepackage{glossary-mcols}
\newglossary[slg]{symbolslist}{syi}{syg}{Symbols}
\newglossary[vlg]{variablelist}{vyi}{vyg}{Variables}

\makeglossaries

\usepackage{float}
\usepackage{placeins}
\usepackage{wrapfig}
\usepackage[figuresleft]{rotating}

\usepackage{graphicx}
\usepackage[space]{grffile}

\usepackage{caption}
\DeclareCaptionLabelFormat{continued}{#1~#2 (continuation)}
\usepackage[skip=1mm]{subcaption}

\newlength{\figurewidth}
\newlength{\imagewidth}
\newlength{\spacerwidth}
\newlength{\spacerheight}
\newcount\nImageColumns

\usepackage{array}
\usepackage{booktabs}
\usepackage{multirow}
\usepackage{threeparttable}

\usepackage{tabularx}
\newcolumntype{Y}{>{\centering\bfseries\arraybackslash}X}
\newcolumntype{Z}{>{\centering\arraybackslash}X}
\newcolumntype{L}{>{\raggedright\arraybackslash}X}
\newcolumntype{R}{>{\raggedleft\arraybackslash}X}
\newcolumntype{C}{>{\centering\bfseries\arraybackslash}m{4.2cm}}
\usepackage{tabu}

\usepackage[noend]{algpseudocode}
\usepackage{algorithm}
\usepackage{algorithmicx}

\algrenewcommand\algorithmicrequire{\textbf{Input:}}
\algrenewcommand\algorithmicreturn{\State \textbf{return}}

\algnewcommand{\algorithmicgoto}{\textbf{go to}}%
\algnewcommand{\Goto}[1]{\algorithmicgoto~\cref{#1}}%

\newcommand{\name}[1]{#1}
\usepackage{textcomp}
\newcommand{\textapprox}{\raisebox{0.5ex}{\texttildelow}}

\usepackage[hidelinks,draft=false]{hyperref}
\usepackage[nameinlink,capitalize,noabbrev]{cleveref}

\usepackage{nameref,zref-xr}
\zxrsetup{toltxlabel}

\makeatletter

\@ifdefinable{\mrcnn}{\def\mrcnn/{\name{Mask \glsentryshort{R-CNN}}}}

\@ifdefinable{\krcnn}{\def\krcnn/{\name{Keypoint \glsentryshort{R-CNN}}}}

\@ifdefinable{\frcnn}{\def\frcnn/{\name{Fibe\glsentryshort{R-CNN}}}}
\@ifdefinable{\ctfire}{\def\ctfire/{\name{\glsentryshort{CT}-\glsentryshort{FIRE}}}}

\@ifdefinable{\identifierlegend}{\def\identifierlegend/{\texttt{+}/\texttt{-}/\texttt{?}: yes, no, random; \texttt{l}/\texttt{c}/\texttt{o}: loops, clutter, overlaps}}

\makeatother
\newacronym{GPU}{GPU}{graphics processing unit}
\newacronym{CPU}{CPU}{central processing unit}
\newacronym{ReLU}{ReLU}{rectified linear unit}
\newacronym{MAPE}{MAPE}{mean absolute percentage error}
\newacronym[first=\glstext{SGDM}]{SGDM}{SGDM}{stochastic gradient descent with momentum}
\newacronym{COCO}{COCO}{\name{Common Objects in Context}}
\newacronym{CNT}{CNT}{carbon nanotube}
\newacronym[firstplural=regions of interest (ROIs)]{ROI}{ROI}{region of interest}
\newacronym{SEM}{SEM}{scanning electron microscope}
\newacronym{CNN}{CNN}{convolutional neural network}
\newacronym{R-CNN}{R-CNN}{region-based convolutional neural network}
\newacronym[first=\glstext{synthPIC}]{synthPIC}{synthPIC}{synthetic particle image creator}
\newacronym{BIC}{BIC}{\name{Bayes}ian information criterion}
\newacronym{MSE}{MSE}{mean squared error}
\newacronym{SSR}{SSR}{sum of squared residuals}
\newacronym{IoU}{IoU}{intersection over union}
\newacronym{AP}{AP}{average precision}
\newacronym{mAP}{mAP}{mean average precision}
\newacronym{CI}{CI}{confidence interval}
\newacronym{CT}{CT}{curvelet transform}
\newacronym{FIRE}{FIRE}{fiber extraction}
\newacronym{TP}{TP}{true positive}
\newacronym{FP}{FP}{false positive}
\newacronym{TN}{TN}{true negative}
\newacronym{FN}{FN}{false negative}

\newglossaryentry{symb:loss}{
	name=\ensuremath{\mathcal{L}},
	description={overall loss},
	sort=L, type=symbolslist
}

\newglossaryentry{symb:lossWidth}{
	name=\ensuremath{\mathcal{L}_\text{fw}},
	description={fiber width regression head loss},
	sort=Lfw, type=symbolslist
}

\newglossaryentry{symb:lossWeightWidth}{
	name=\ensuremath{w_\text{fw}},
	description={weight of the fiber width regression head loss},
	sort=wfw, type=symbolslist
}

\newglossaryentry{symb:lossWeightLength}{
	name=\ensuremath{w_\text{fl}},
	description={weight of the fiber length regression head loss},
	sort=wfl, type=symbolslist
}
	
\newglossaryentry{symb:lossLength}{
	name=\ensuremath{\mathcal{L}_\text{fl}},
	description={fiber length regression head loss},
	sort=Lfl, type=symbolslist
}

\newglossaryentry{symb:lossClass}{
	name=\ensuremath{\mathcal{L}_\text{cls}},
	description={instance classification head loss},
	sort=Lcls, type=symbolslist
}

\newglossaryentry{symb:lossMask}{
	name=\ensuremath{\mathcal{L}_\text{mask}},
	description={mask segmentation head loss},
	sort=Lmask, type=symbolslist
}

\newglossaryentry{symb:lossBox}{
	name=\ensuremath{\mathcal{L}_\text{box}},
	description={bounding box regression head loss},
	sort=Lbox, type=symbolslist
}

\newglossaryentry{symb:lossKeypoint}{
	name=\ensuremath{\mathcal{L}_\text{kp}},
	description={keypoint regression head loss},
	sort=Lkp, type=symbolslist
}
	
\newglossaryentry{symb:numberOfDates}{
	name=\ensuremath{n},
	description={number of dates},
	sort=n, type=symbolslist
}

\newglossaryentry{symb:numberOfKeypoints}{
	name=\ensuremath{k},
	description={number of keypoints},
	sort=k, type=symbolslist
}

\newglossaryentry{symb:prediction}{
	name=\ensuremath{y},
	description={prediction},
	sort=y, type=symbolslist
}

\newglossaryentry{symb:target}{
	name=\ensuremath{t},
	description={target},
	sort=t, type=symbolslist
}

\newglossaryentry{symb:index}{
	name=\ensuremath{i},
	description={index},
	sort=i, type=symbolslist
}

\newglossaryentry{symb:index2}{
	name=\ensuremath{j},
	description={index},
	sort=j, type=symbolslist
}

\newglossaryentry{symb:meanSquaredError}{
	name=\ensuremath{\text{MSE}},
	description={mean squared error},
	sort=MSE, type=symbolslist
}

\newglossaryentry{symb:sumOfSquaredResiduals}{
	name=\ensuremath{\text{SSR}},
	description={sum of squared residuals},
	sort=SSR, type=symbolslist
}

\newglossaryentry{symb:bayesianInformationCriterion}{
	name=\ensuremath{\text{BIC}},
	description={\name{Bayes}ian information criterion},
	sort=BIC, type=symbolslist
}

\newglossaryentry{symb:baseLearningRate}{
	name=\ensuremath{\alpha_\text{base}},
	description={base learning rate},
	sort=abase, type=symbolslist
}

\newglossaryentry{symb:intersectionoverUnion}{
	name=\ensuremath{\text{IoU}},
	description={intersection over union},
	sort=IoU, type=symbolslist
}

\newglossaryentry{symb:meanAveragePrecision}{
	name=\ensuremath{\text{mAP}},
	description={mean average precision at \ensuremath{\text{IoU}=\text{0.5:0.05:0.95}}},
	sort=mAP, type=symbolslist
}

\newglossaryentry{symb:averagePrecision50}{
	name=\ensuremath{\text{AP}_{50}},
	description={average precision at \ensuremath{\text{IoU}=0.5}},
	sort=AP50, type=symbolslist
}

\newglossaryentry{symb:averagePrecision75}{
	name=\ensuremath{\text{AP}_{75}},
	description={average precision at \ensuremath{\text{IoU}=0.75}},
	sort=AP75, type=symbolslist
}

\newglossaryentry{symb:meanAbsolutePercentageError}{
	name=\ensuremath{\text{MAPE}},
	description={mean absolute percentage error},
	sort=MAPE, type=symbolslist
}

\newglossaryentry{symb:percentageError}{
	name=\ensuremath{\Delta y_{\%}},
	description={percentage error},
	sort=Deltaypercentage, type=symbolslist
}

\newglossaryentry{symb:kullbackLeiblerDivergence}{
	name=\ensuremath{D_\text{KL}},
	description={\name{Kullback}–\name{Leibler} divergence},
	sort=DKL, type=symbolslist
}

\newglossaryentry{symb:probabilityDistributionP}{
	name=\ensuremath{P},
	description={probability distribution},
	sort=P, type=symbolslist
}

\newglossaryentry{symb:probabilityDistributionQ}{
	name=\ensuremath{Q},
	description={probability distribution},
	sort=Q, type=symbolslist
}

\newglossaryentry{var:fiberwidth}{
	name=\ensuremath{fiber\_width},
	description={fiber width},
	sort=w, type=variablelist
}

\newglossaryentry{var:fiberlength}{
	name=\ensuremath{fiber\_length},
	description={fiber length},
	sort=l, type=variablelist
}

\newglossaryentry{var:keypoints}{
	name=\ensuremath{keypoints},
	description={fiber keypoints},
	sort=kpts, type=variablelist
}

\newglossaryentry{var:keypoint}{
	name=\ensuremath{keypoint},
	description={fiber keypoint},
	sort=k, type=variablelist
}

\newglossaryentry{var:mask}{
	name=\ensuremath{mask},
	description={mask},
	sort=m, type=variablelist
}

\newglossaryentry{var:splinemask}{
	name=\ensuremath{spline\_mask},
	description={mask},
	sort=m, type=variablelist
}

\newglossaryentry{var:iou}{
	name=\ensuremath{iou},
	description={iou},
	sort=iou, type=variablelist
}

\newglossaryentry{var:segment}{
	name=\ensuremath{segment},
	description={seg},
	sort=seg, type=variablelist
}

\newglossaryentry{var:segments}{
	name=\ensuremath{segments},
	description={seg},
	sort=seg, type=variablelist
}

\newglossaryentry{var:splinelength}{
	name=\ensuremath{spline\_length},
	description={splinelength},
	sort=seg, type=variablelist
}

\newglossaryentry{var:splinelengtherror}{
	name=\ensuremath{spline\_length\_error},
	description={splinelengtherror},
	sort=seg, type=variablelist
}

\newglossaryentry{var:numberofkeypoints}{
	name=\ensuremath{number\_of\_keypoints},
	description={number_of_keypoints},
	sort=seg, type=variablelist
}

\DeclareSIUnit{\px}{px}
\DeclareSIUnit{\pp}{\textup{pp}}
\DeclareSIUnit{\percentagepoints}{\textup{percentage points}}

\begin{document}
	
	\begin{frontmatter}
		\journal{
}	 	 
		\title{
\frcnn/: Expanding \mrcnn/ to Improve Image-Based Fiber Analysis}	
	 	 
\author{M. Frei\corref{mycorrespondingauthor}}
\cortext[mycorrespondingauthor]{Corresponding author. Tel.: +49 203 379--3621}
\ead{max.frei@uni-due.de}

\author{F. E. Kruis\corref{}}

\address{Institute of Technology for Nanostructures (NST) and Center for Nanointegration Duisburg-Essen (CENIDE)\\University of Duisburg-Essen, Duisburg, D-47057, Germany}
	 	 
		\begin{abstract}
Fiber-shaped materials (e.g. carbon nano tubes) are of great relevance, due to their unique properties but also the health risk they can impose. Unfortunately, image-based analysis of fibers still involves manual annotation, which is a time-consuming and costly process. 

We therefore propose the use of \glspl{R-CNN} to automate this task. \mrcnn/, the most widely used \gls{R-CNN} for semantic segmentation tasks, is prone to errors when it comes to the analysis of fiber-shaped objects. Hence, a new architecture -- \frcnn/ -- is introduced and validated. \frcnn/ combines two established \gls{R-CNN} architectures (\name{Mask} and \krcnn/) and adds additional network heads for the prediction of fiber widths and lengths. As a result, \frcnn/ is able to surpass the mean average precision of \mrcnn/ by \SI{33}{\percent} (\SI{11}{\percentagepoints}) on a novel test data set of fiber images. 
\\\\
Source code available online.
		\end{abstract}

	\end{frontmatter}

	
	
	\graphicspath{{./figs/introduction/}}
\section{Introduction}
\label{sec:Introduction}
Fiber-shaped materials, such as asbestos, \glspl{CNT} and fiberglass, possess a janiform character. On the one hand, they exhibit unique and attractive material properties~\cite{Rajak.2019}, so that they are of great scientific and commercial interest. On the other hand, they can have a severe toxicological potential, which impedes their environmental compatibility, especially when being delivered as aerosols~\cite{Mohanta.2019}. Since both, the desired and the undesired properties, are often shape and size dependent, image-based fiber analysis is an important tool for the exploration and assessment of risks and chances of fiber-shaped materials~\cite{Brossell.2019,VanOrden.2006,Bowen.2002,Wen.2009,Chen.2013}. 

There already exist automated online measurement methods for the determination of the length and width distributions of \glspl{CNT}, e.g. via differential mobility analysis~\cite{Pease.2009}. However, these methods always require a validation based on some form of image-based analysis. Unfortunately, automated annotation algorithms for images of fiber-shaped particles are scarce, especially for overlapping and occluded fibers. 

The most notable and accessible of these algorithms is \ctfire/~\cite{Bredfeldt.2014}, which is based on classical image processing methods and fast discrete \gls{CT}~\cite{Candes.2006}, combined with the \gls{FIRE}~\cite{Stein.2008}. 

\begin{figure}
	\centering
	\includegraphics[width=\figurewidth]{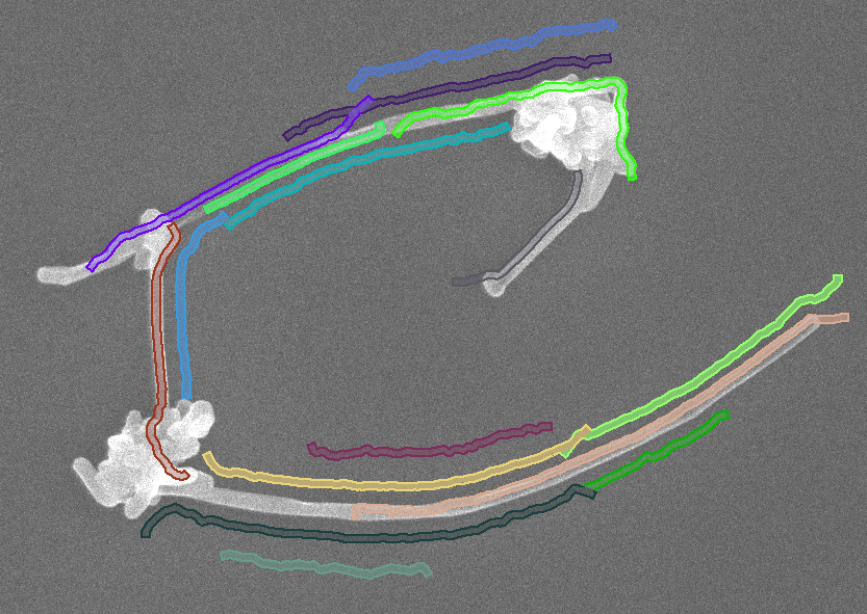}
	\caption{Example of the insufficient detection quality of \ctfire/, when being applied to an image from the test data at hand. To reduce the number of false-positives, detections featuring a short fiber length were filtered with a threshold of \SI{150}{px} (pixels). The remaining false-positive detections result from artifacts induced by the fast discrete \gls{CT} used by \ctfire/ and could not be reduced any further through tuning of the associated parameters.}
	\label{fig:CtFireExampleDetection}
\end{figure}

While having been successfully applied to high-contrast images of collagen fibers~\cite{Bredfeldt.2014} and perfectly straight glass fibers~\cite{Giusti.2018}, for more difficult data like the test data at hand (see \cref{sec:Data-DataSets}), \ctfire/ produces insufficient results (see \cref{fig:CtFireExampleDetection}), even if supplied with a priori information about the sample (e.g. minimum fiber length). Furthermore, it is extremely slow (\textapprox\SI{5}{\minute} per image on a single \glsentryshort{CPU} core; see also \cref{app:tab:Hardware}) and features many parameters that need to be carefully tuned -- often on a per-image basis -- to optimize the results.
Due to these reasons, analyses of fiber length and width distributions still often have to be carried out manually. This practice is not only laborious, expensive and repetitive but also error-prone, due to the subjectivity and exhaustion of the operators~\cite{Allen.2003}.

To solve these problems, we propose a new approach to fully automated imaging fiber analysis, with help of \glspl{CNN}. Recently, neural networks in general and \glspl{CNN} in particular, have been applied successfully to particle measurement problems, such as the characterization of particle shapes and their size distribution~\cite{Frei.2020,Frei.2018,Heisel.2017,Heisel.2019,Wu.2020} as well as the classification of the chiral indices of \glspl{CNT}~\cite{Forster.2020}. They are therefore promising candidates for the solution of the problem at hand. The main advantage of \glspl{CNN} is that they require no user-tunable parameters, once they have been trained. Also, they are outstandingly robust to changes in imaging conditions. However, they require a set of already annotated samples for the training~\cite{Frei.2020}. 

\begin{figure}
	\centering
	\includegraphics[width=\figurewidth]{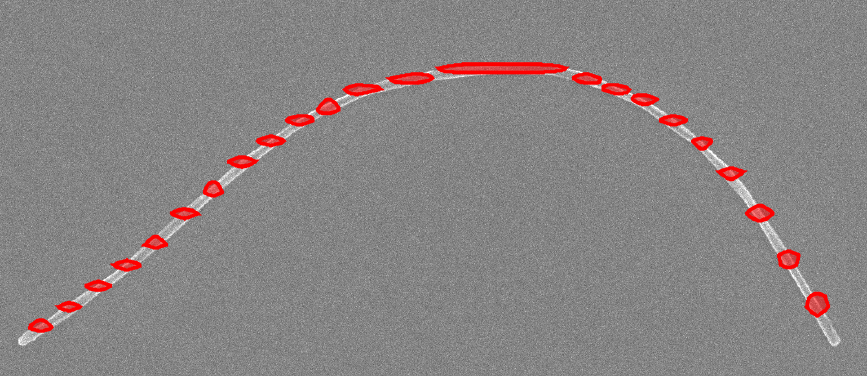}
	\caption{Example of the poor detection quality of \mrcnn/, when being trained on and applied to fiber images from the test data at hand.}
	\label{fig:Motivation}
\end{figure}

Our previously presented, \mrcnn/-based, particle analysis method~\cite{Frei.2020} works well for spherical instances\footnote{In computer vision, an instance refers to an object of a certain class, e.g. a particle.}, i.e. particles, even if they exhibit large amounts of occlusion and sintering. However, \mrcnn/ yields quite ragged instance masks\footnote{An instance mask assigns a binary value to each pixel of an input image: \emph{false} for background pixels and \emph{true} for foreground, i.e. instance pixels.}, when being trained on and applied to fiber images (see \cref{fig:Motivation}) and elongated objects in general~\cite{Looi.2019}. 
The \mrcnn/ architecture is region-based. Contrarily to spheres, fibers contribute only little information to the extracted \gls{ROI} feature maps, due to their thinness and curvature and therefore small area relative to the area of their associated \glspl{ROI}. Apparently, the extracted features do not suffice to reliably reconstruct the instance masks of the fibers directly. 

However, we hypothesize that the features may be meaningful enough to extract keypoints, as well as widths and lengths of fibers. While the latter are often sought-after measurands themselves, their combination with the extracted keypoints also allows a more complete reconstruction of instance masks.

In this publication, we therefore propose, implement and validate an extension of the \mrcnn/ architecture, hereby named \frcnn/, to extract keypoints, widths and lengths of fibers from images, thereby improving automatic fiber shape analysis.

\enlargethispage{\baselineskip}

	\graphicspath{{./figs/data/}}
\section{Training and Test Data}
\label{sec:Data}
The fiber images used in this publication are courtesy of the \name{Institute of Energy and Environmental Technology e.V. (IUTA)} and were created using a \name{JEOL JSM-7500F} field emission \gls{SEM}. The pictured fibers are \glspl{CNT}, deposited from the gas phase (see \cref{fig:DataSetComparison}).
\nImageColumns4\relax 
\setlength{\spacerwidth}{1mm}
\setlength{\imagewidth}{(\textwidth-\spacerwidth*\nImageColumns)/\nImageColumns}
\setlength{\spacerheight}{1cm}
\newcommand{\hspacer}{\hspace{\spacerwidth}\hspace{\spacerwidth}}
\newcommand{\image}[1]{\includegraphics[width=\imagewidth]{#1}}
\graphicspath{{./figs/data/dataset-comparison/}}
\begin{figure*}
	\setlength{\tabcolsep}{0mm} %
	\centering
	\begin{tabularx}{\textwidth}{XXcXX}
		\image{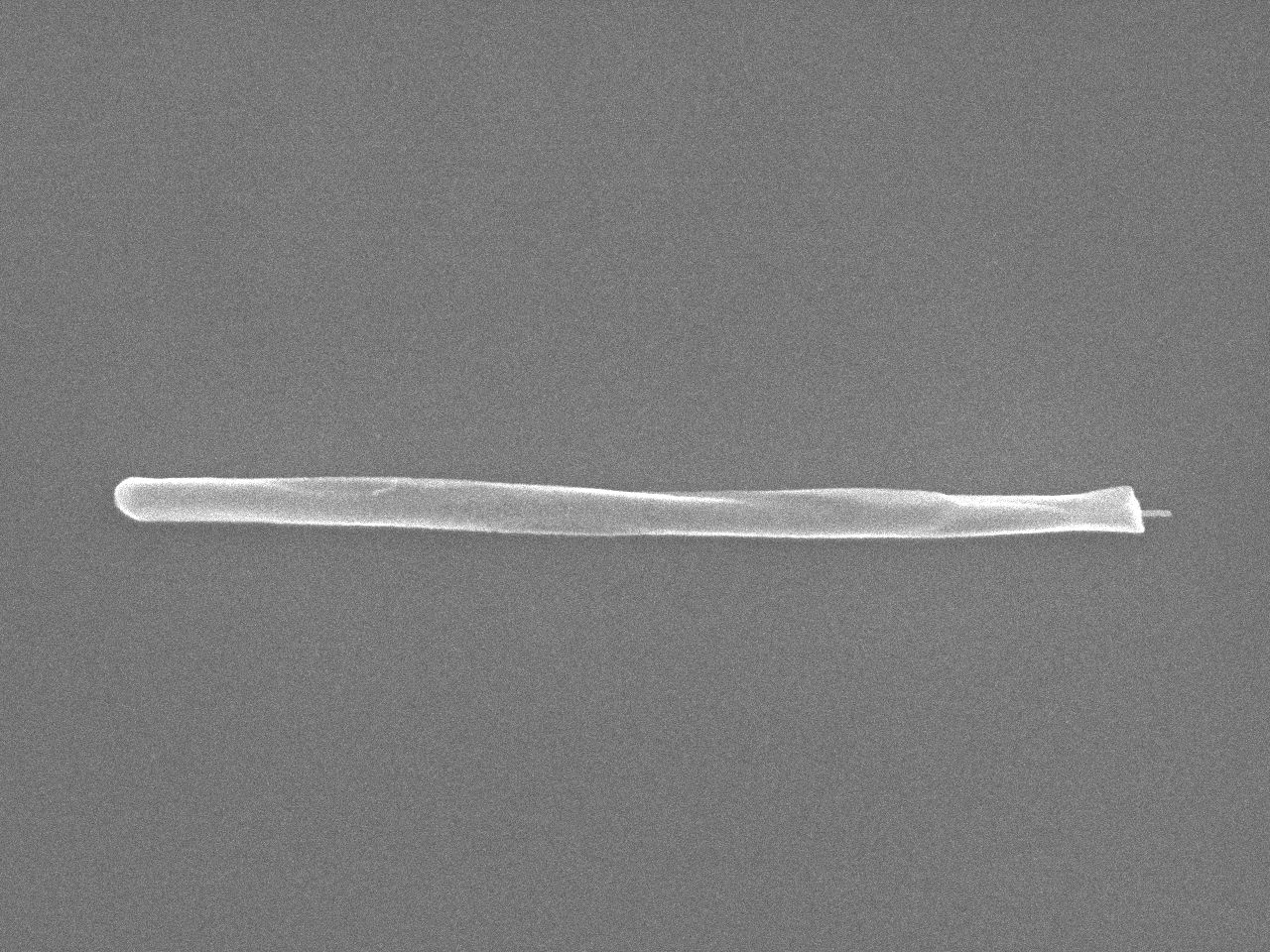}             & \image{_loops_-clutter_-overlaps_1}             & \hspacer & \image{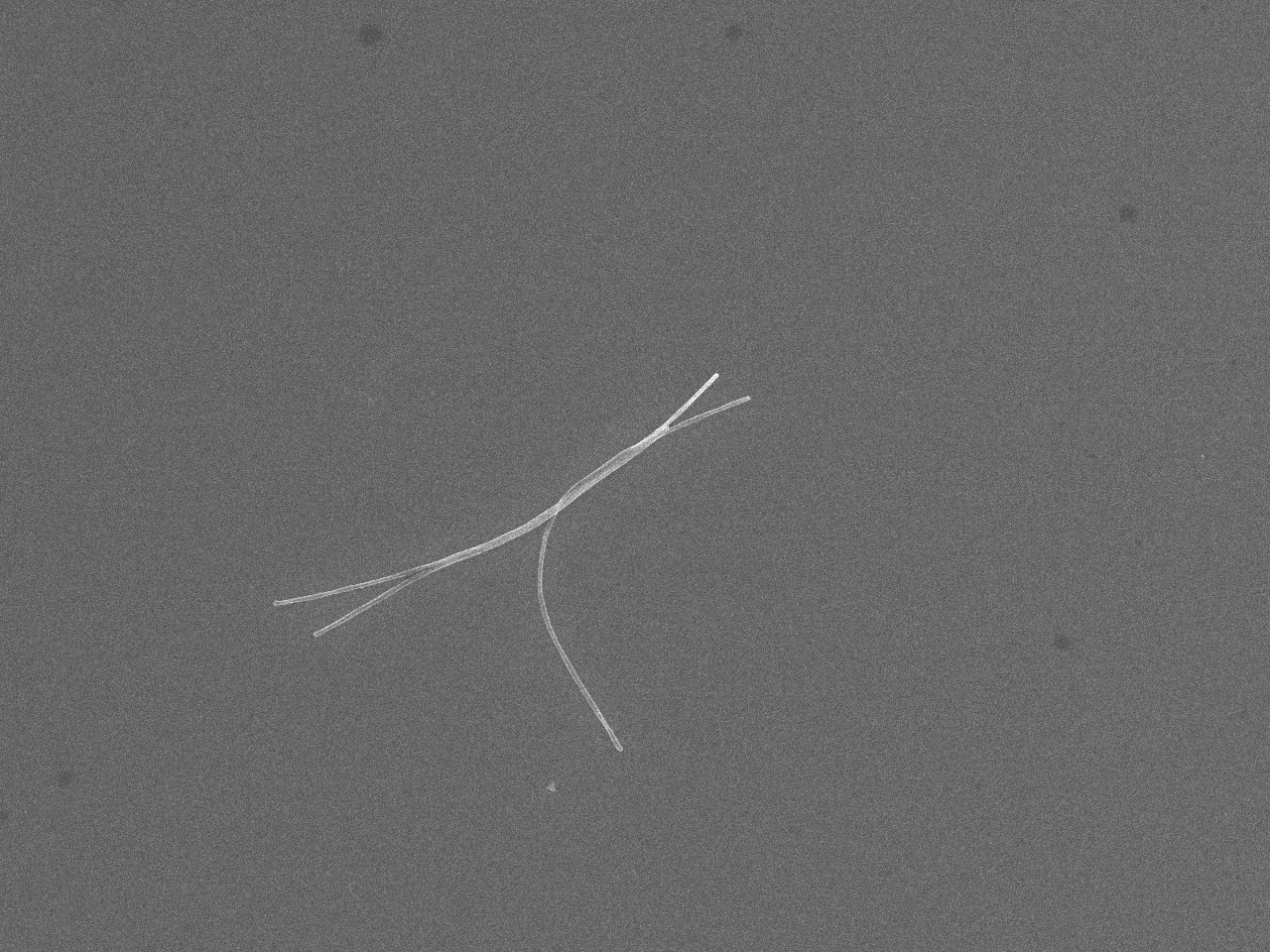}             & \image{_loops_-clutter_+overlaps_1}             \\
		                             \multicolumn{2}{c}{\texttt{[-l|-c|-o]}}                              &          &                              \multicolumn{2}{c}{\texttt{[-l|-c|+o]}}                              \\
		\image{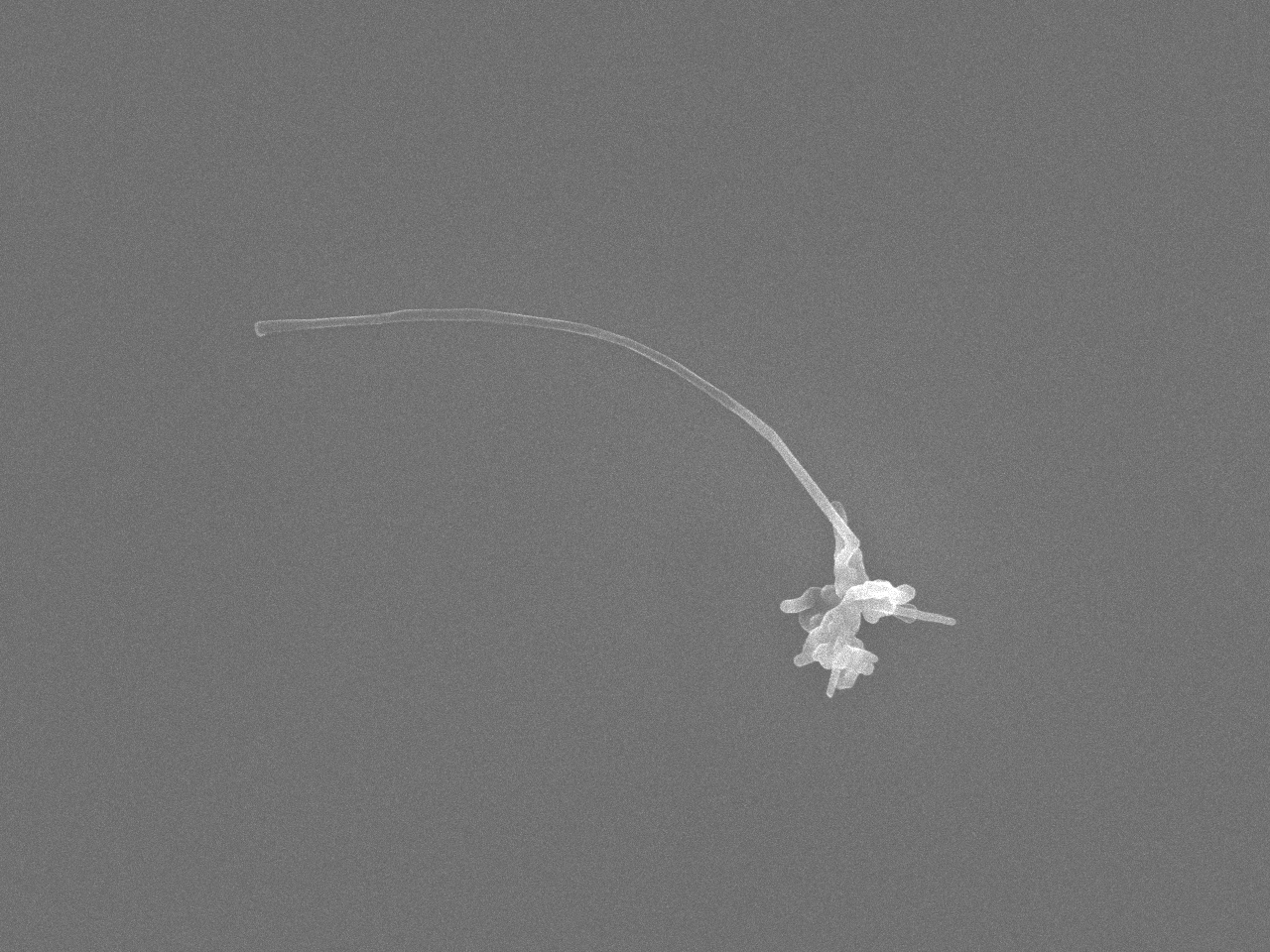}             & \image{_loops_+clutter_-overlaps_1}             &          & \image{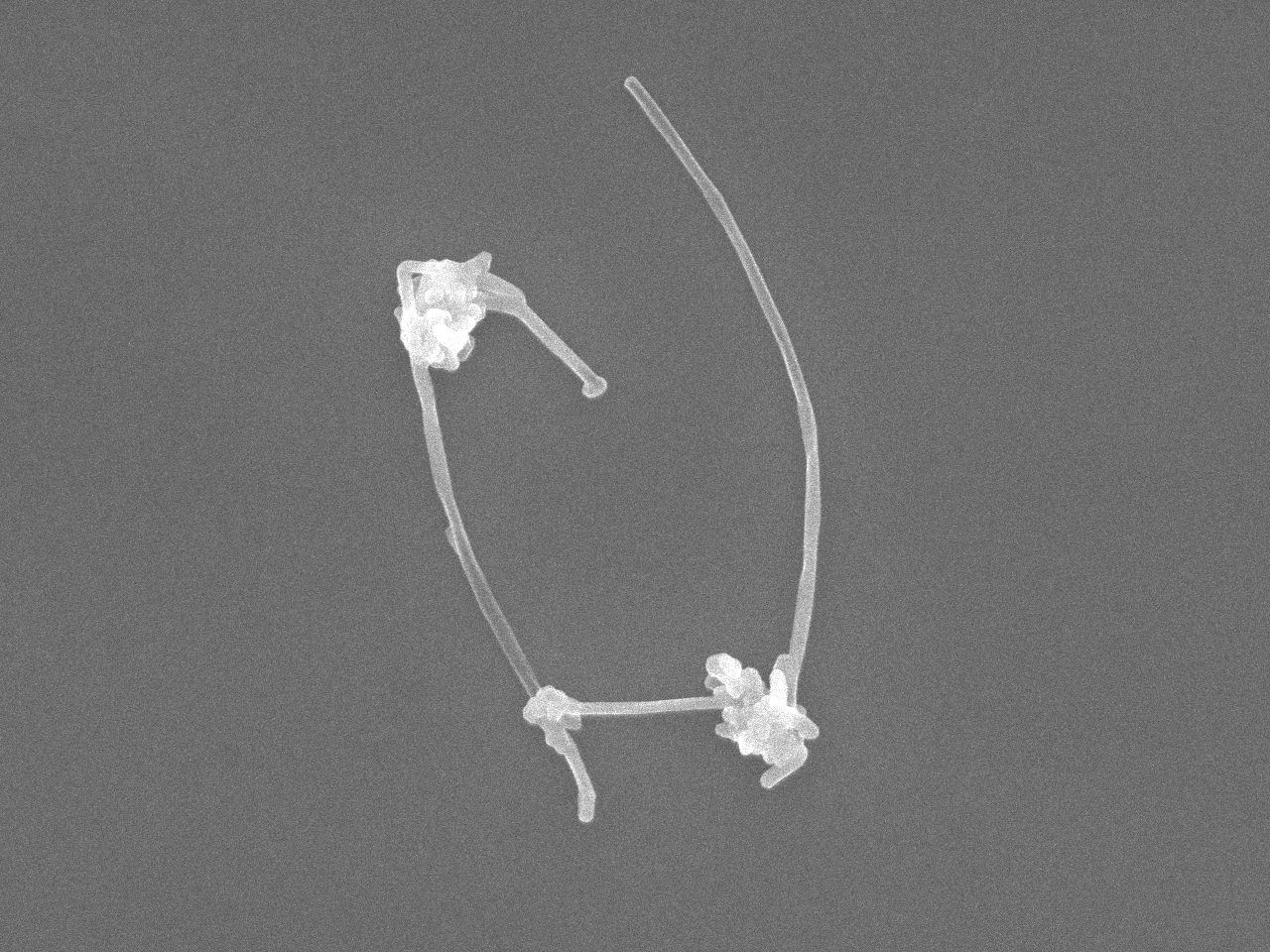}             & \image{_loops_+clutter_+overlaps_1}             \\
		                             \multicolumn{2}{c}{\texttt{[-l|+c|-o]}}                              &          &                              \multicolumn{2}{c}{\texttt{[-l|+c|+o]}}                              \\
		\image{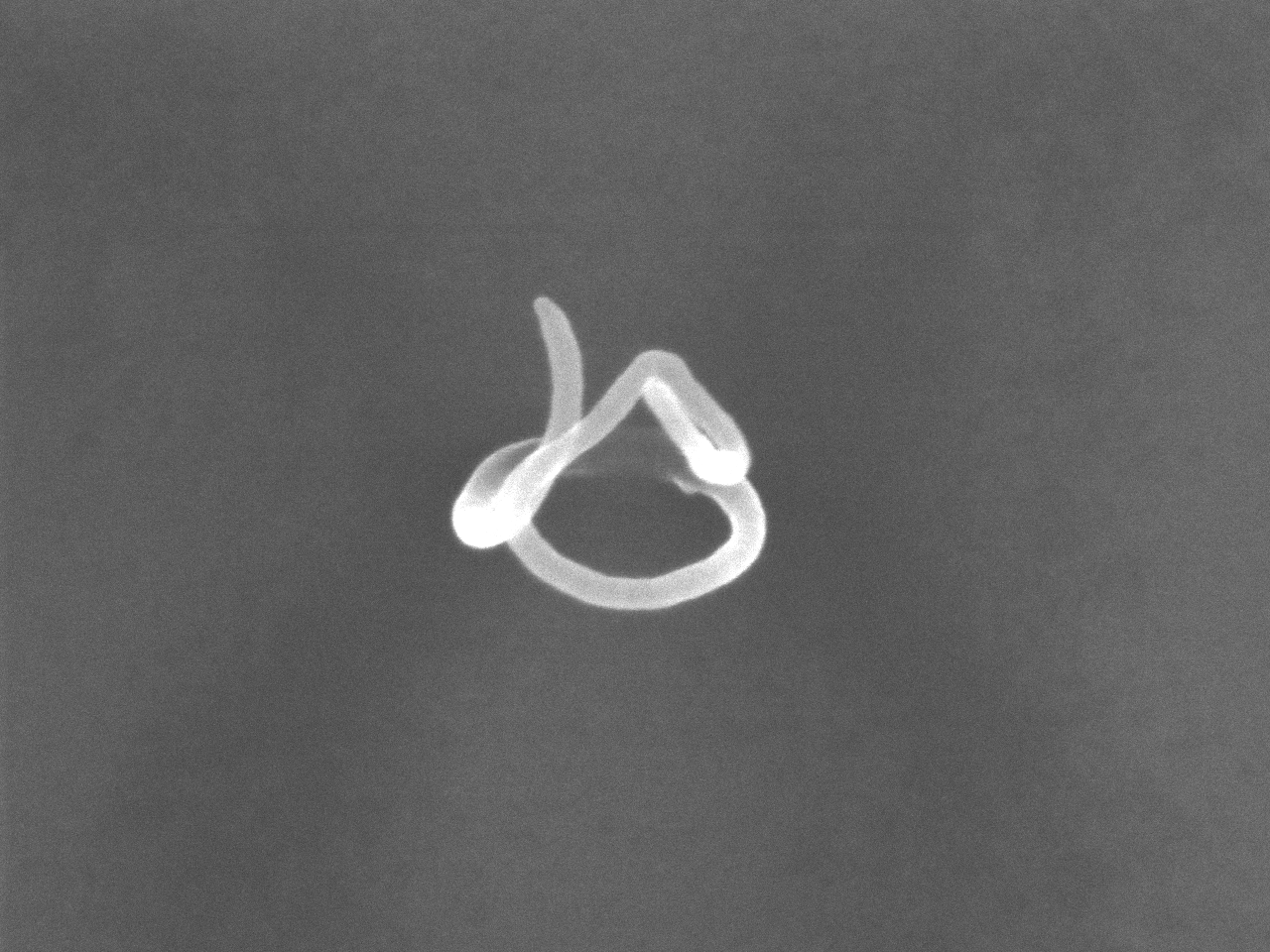}             & \image{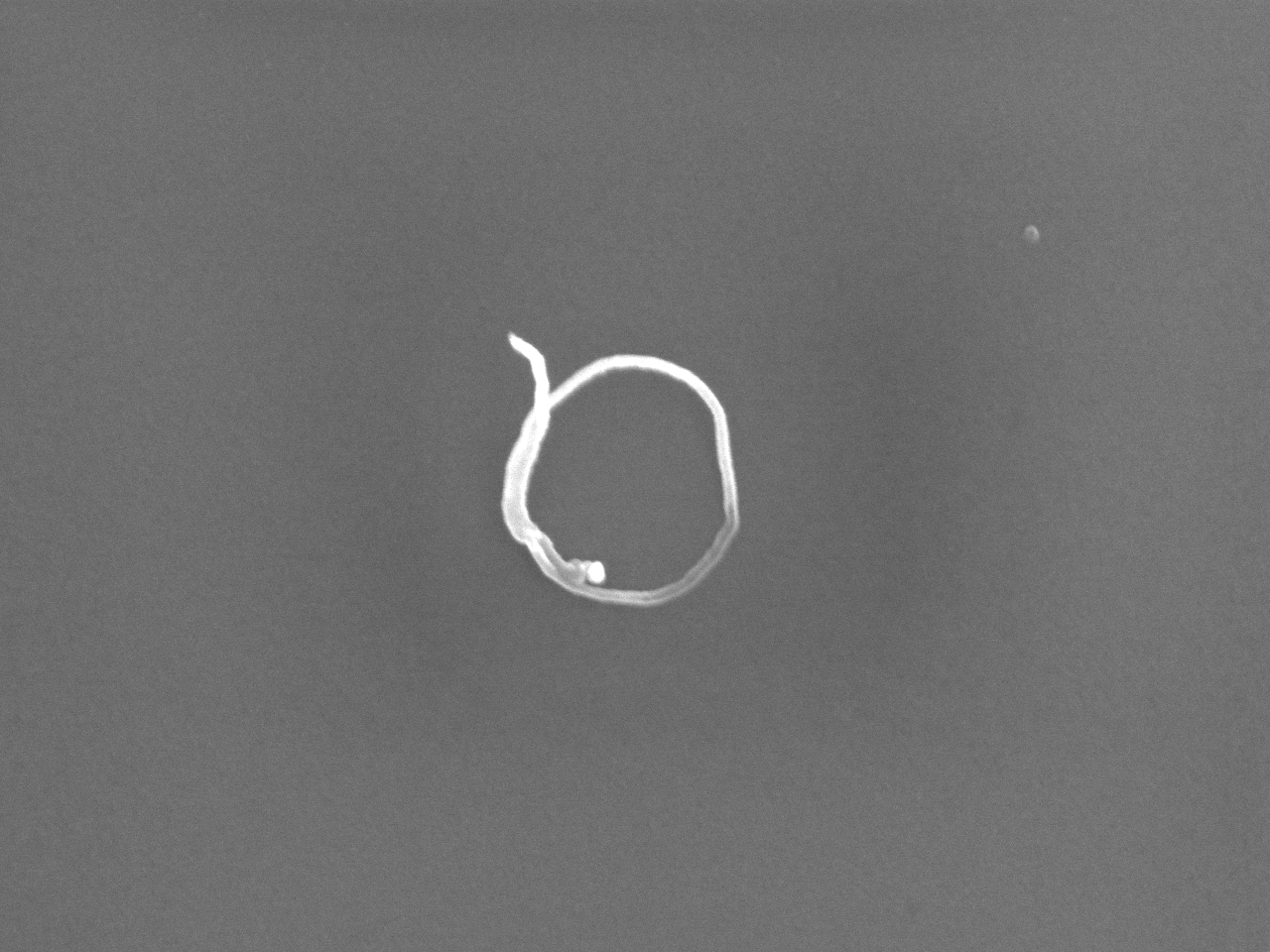}             &          & \image{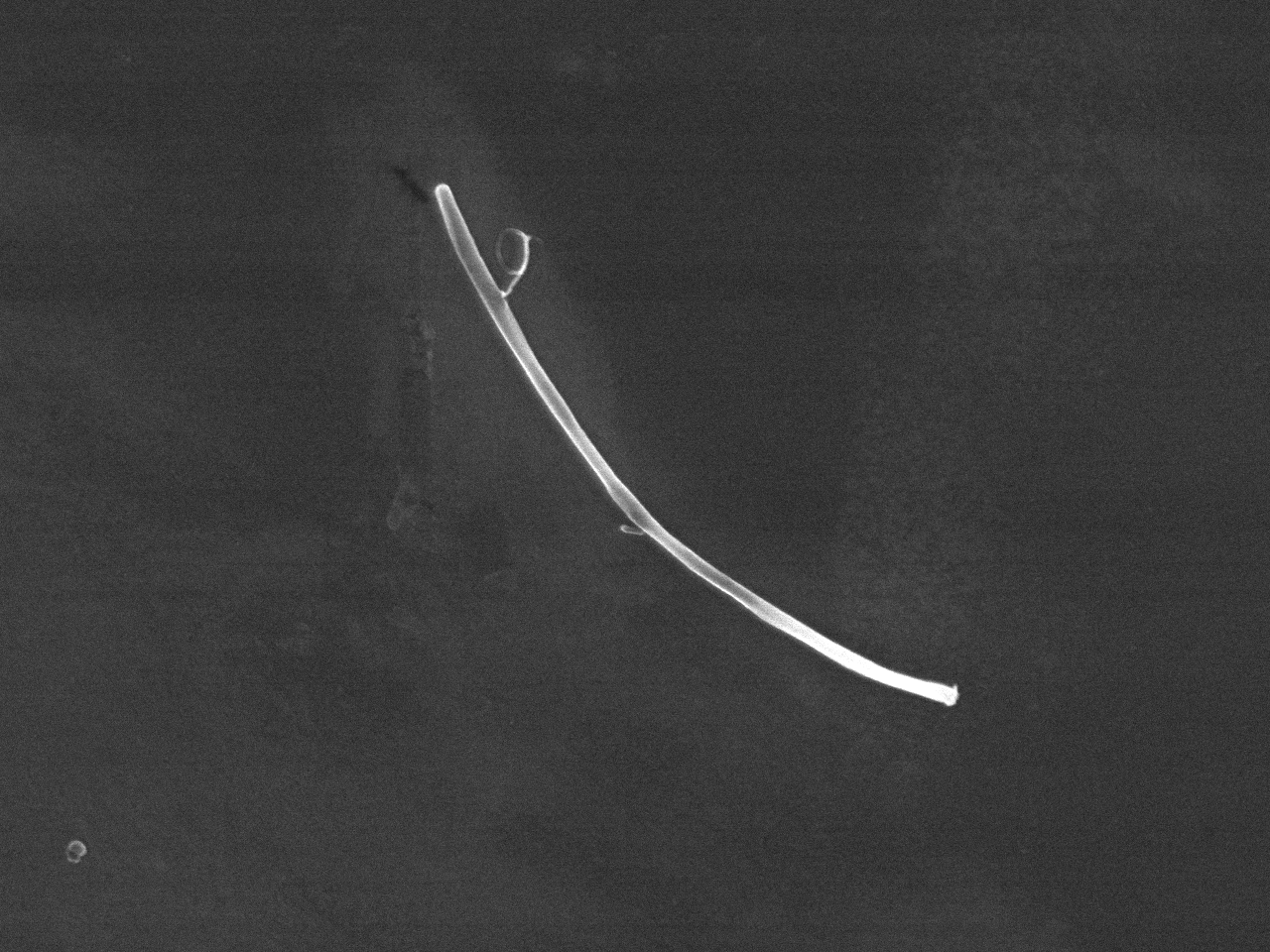}             & \image{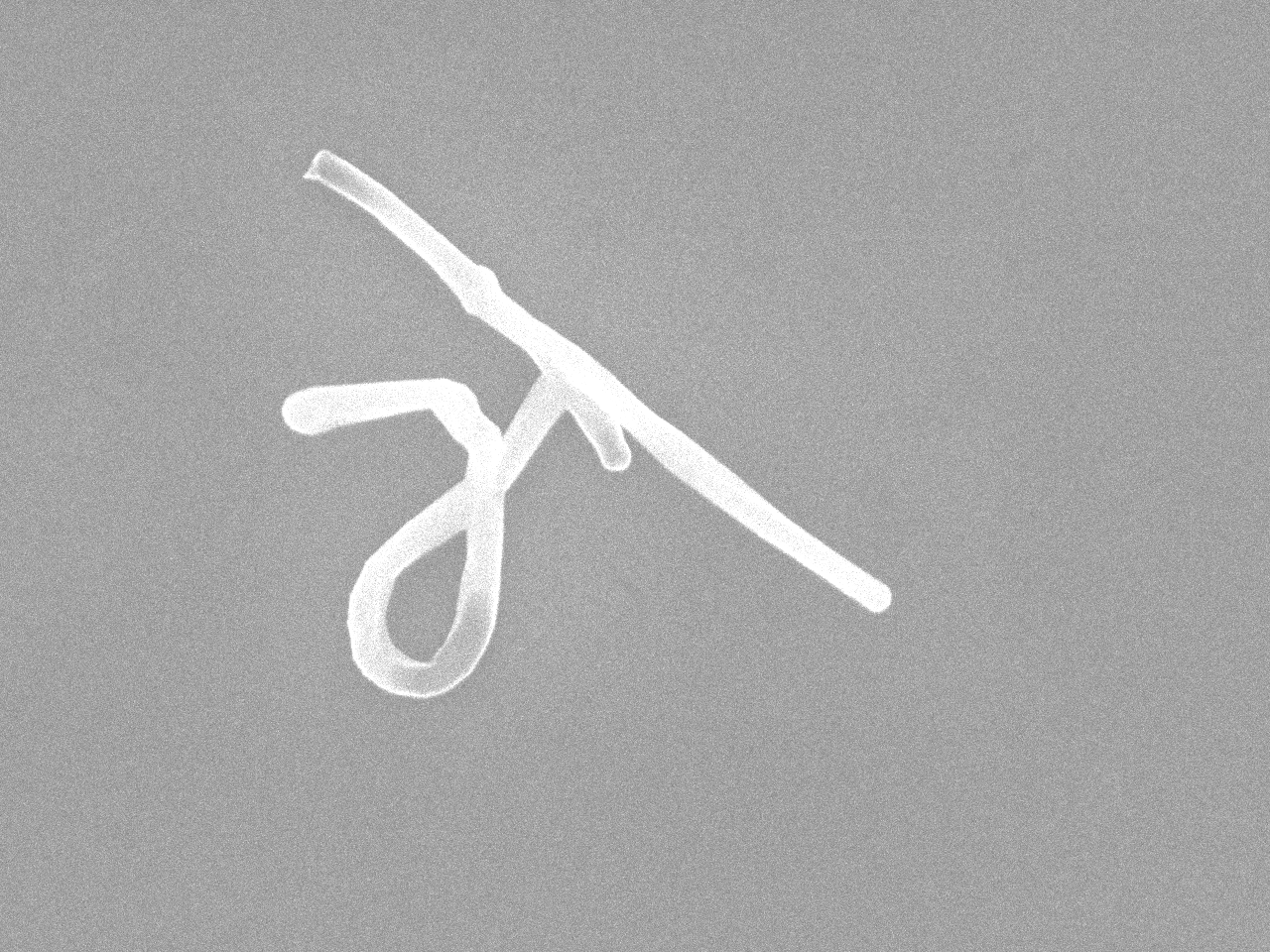}             \\
		                             \multicolumn{2}{c}{\texttt{[+l|-c|-o]}}                              &          &                             \multicolumn{2}{c}{\texttt{[+l|-c|+o]} }                              \\
		\image{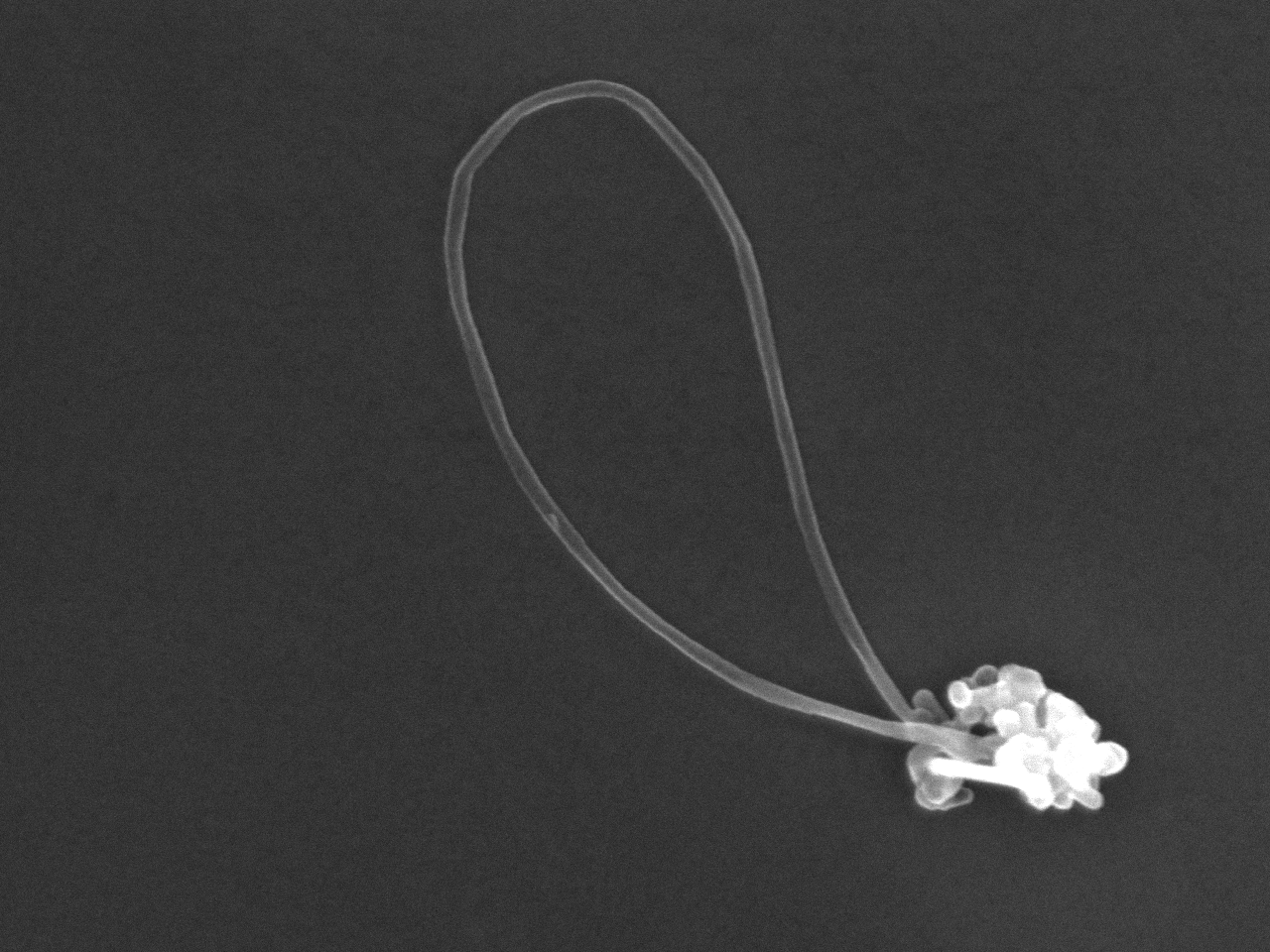}             & \image{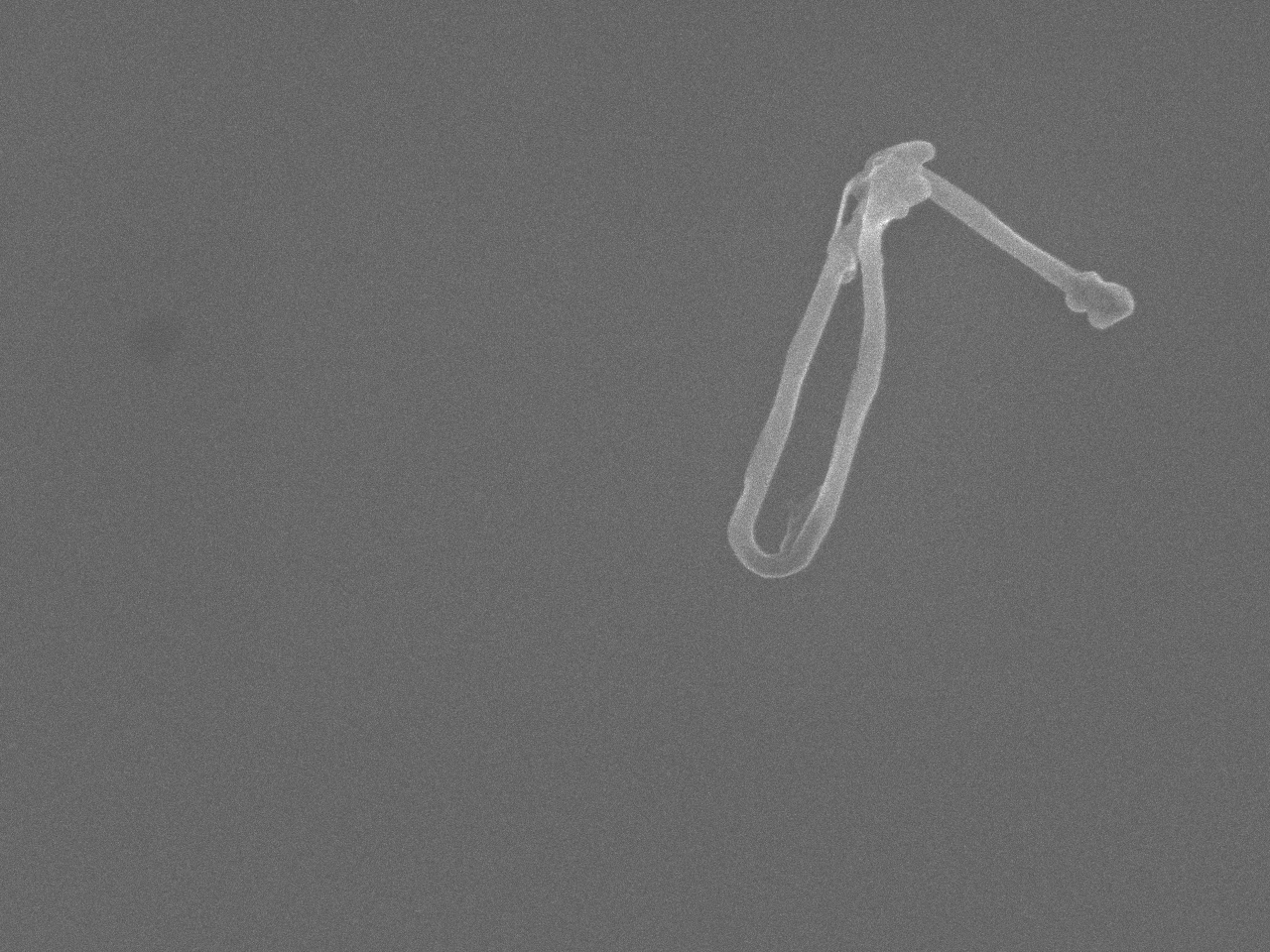}             &          & \image{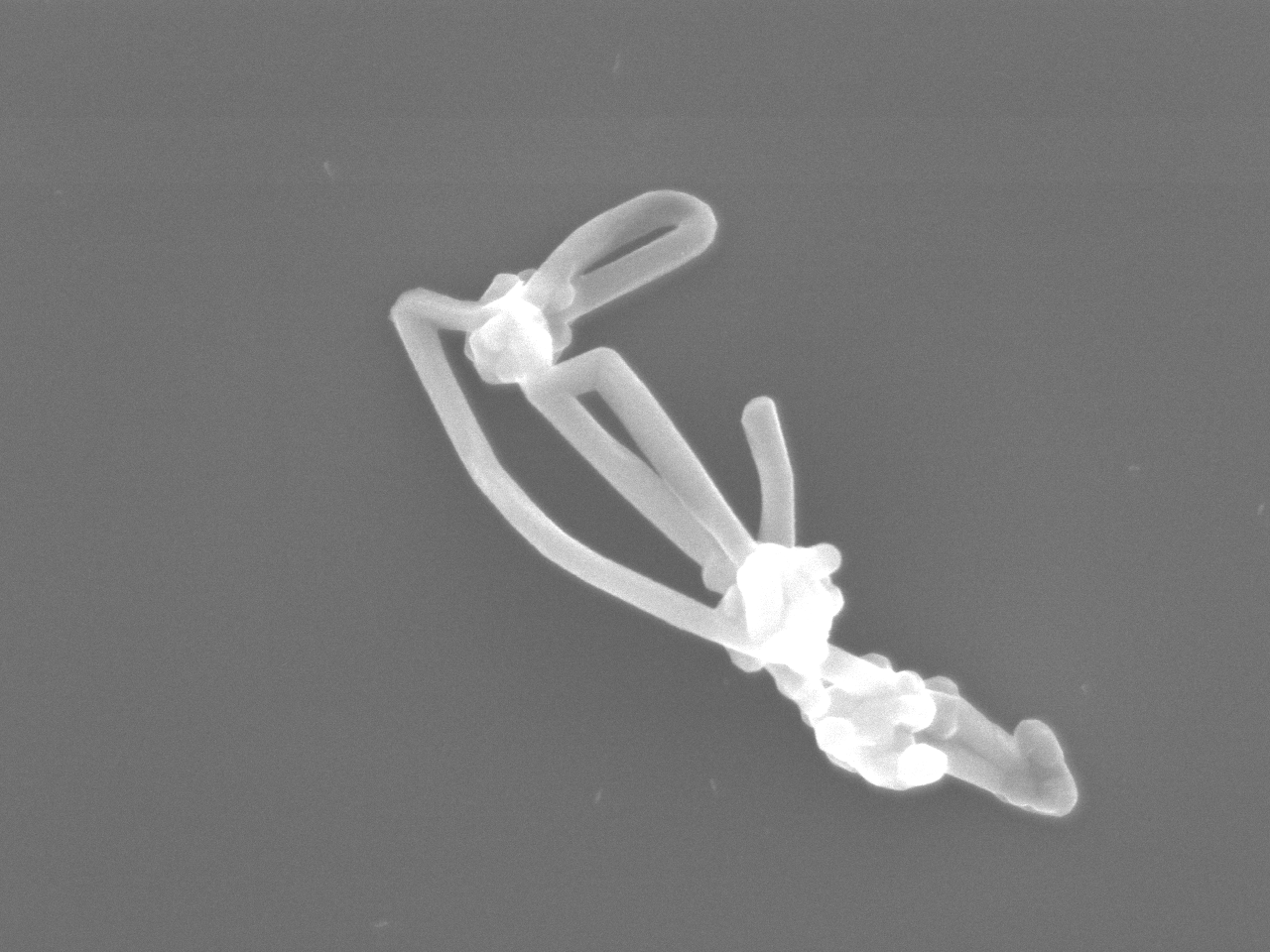}             & \image{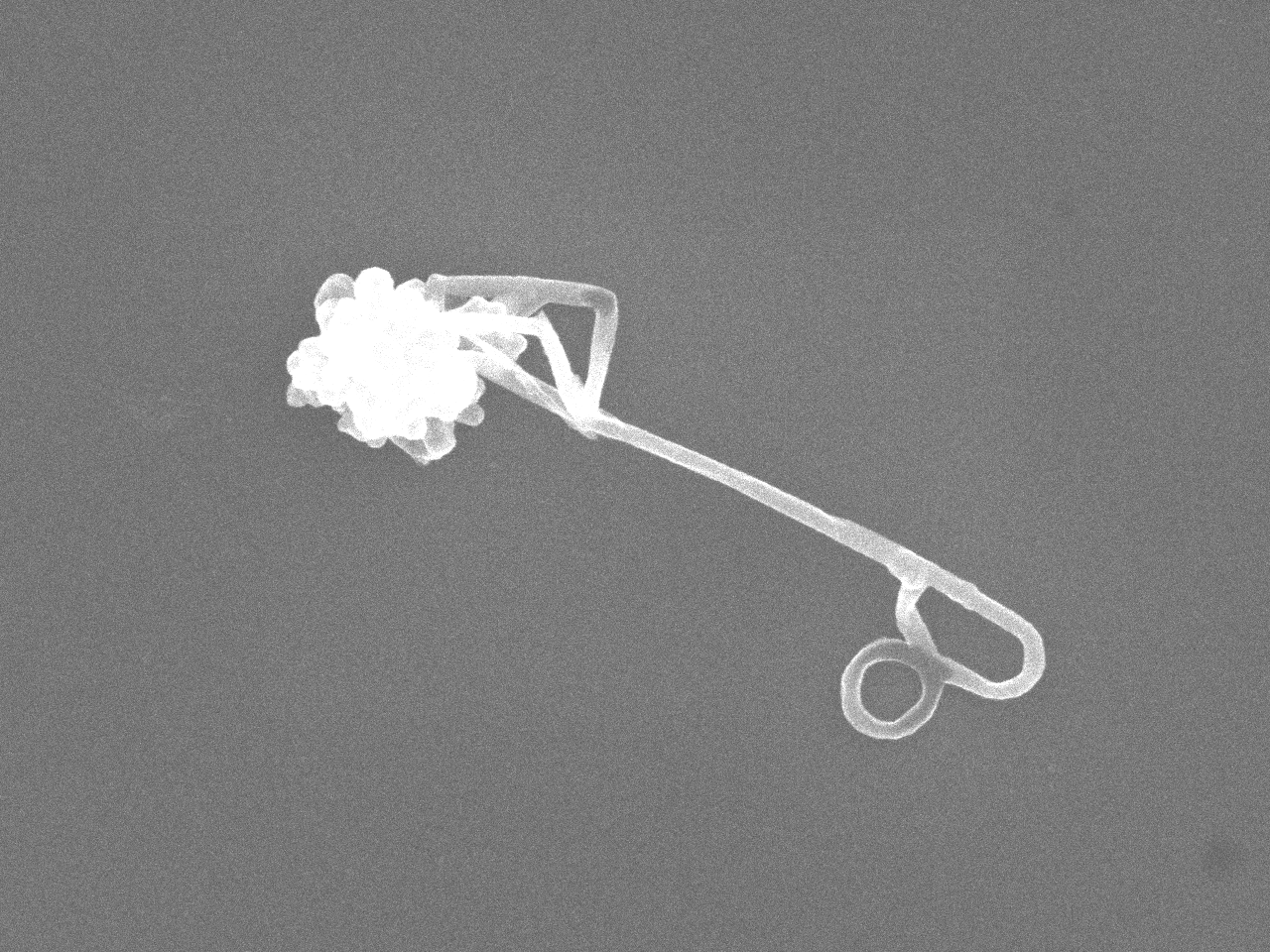}             \\
		                             \multicolumn{2}{c}{\texttt{[+l|+c|-o]}}                              &          &                             \multicolumn{2}{c}{\texttt{[+l|+c|+o]} }                              \\
		\image{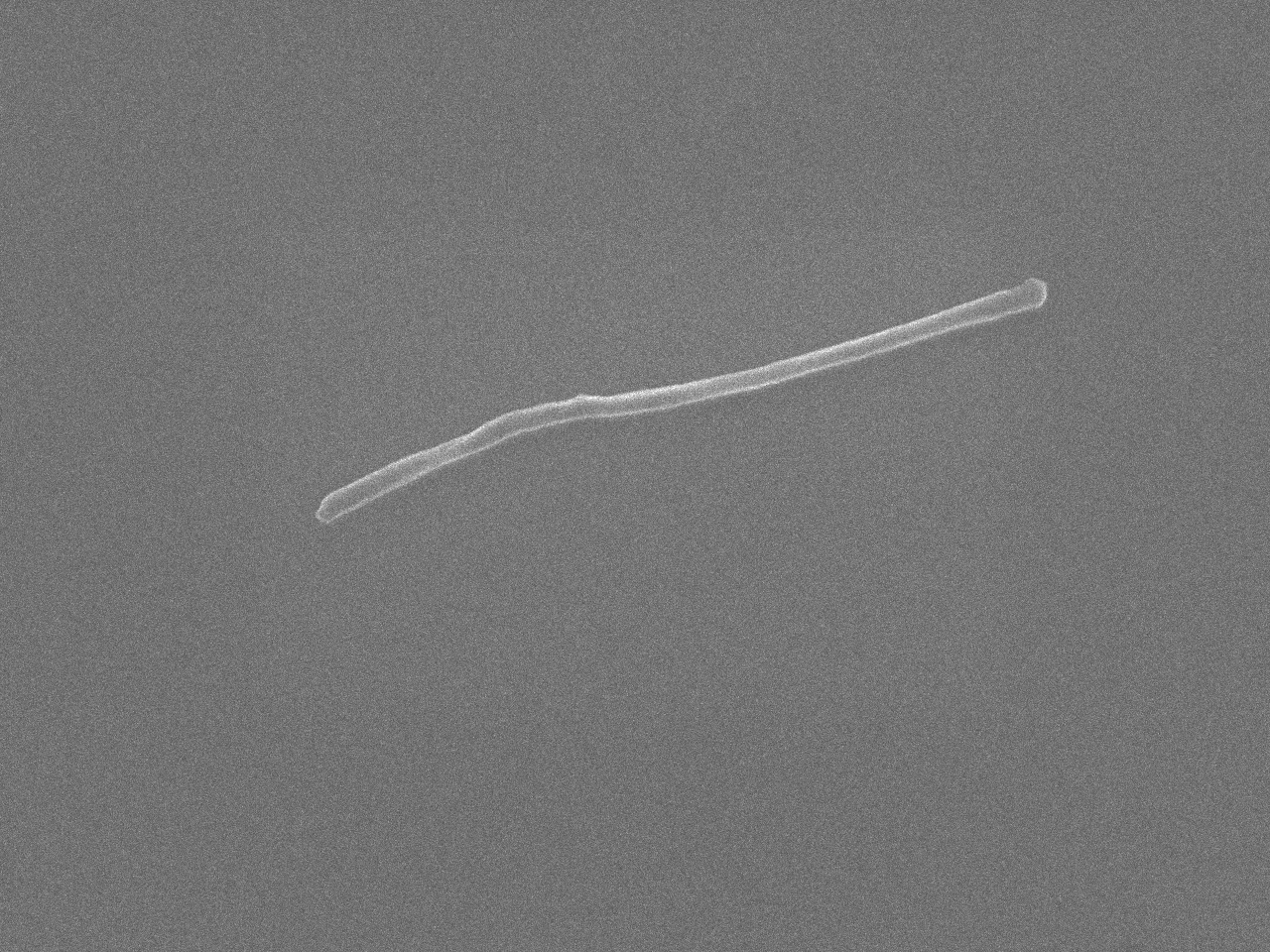} & \image{_loops_-clutter_-overlaps__automatic__1} &          & \image{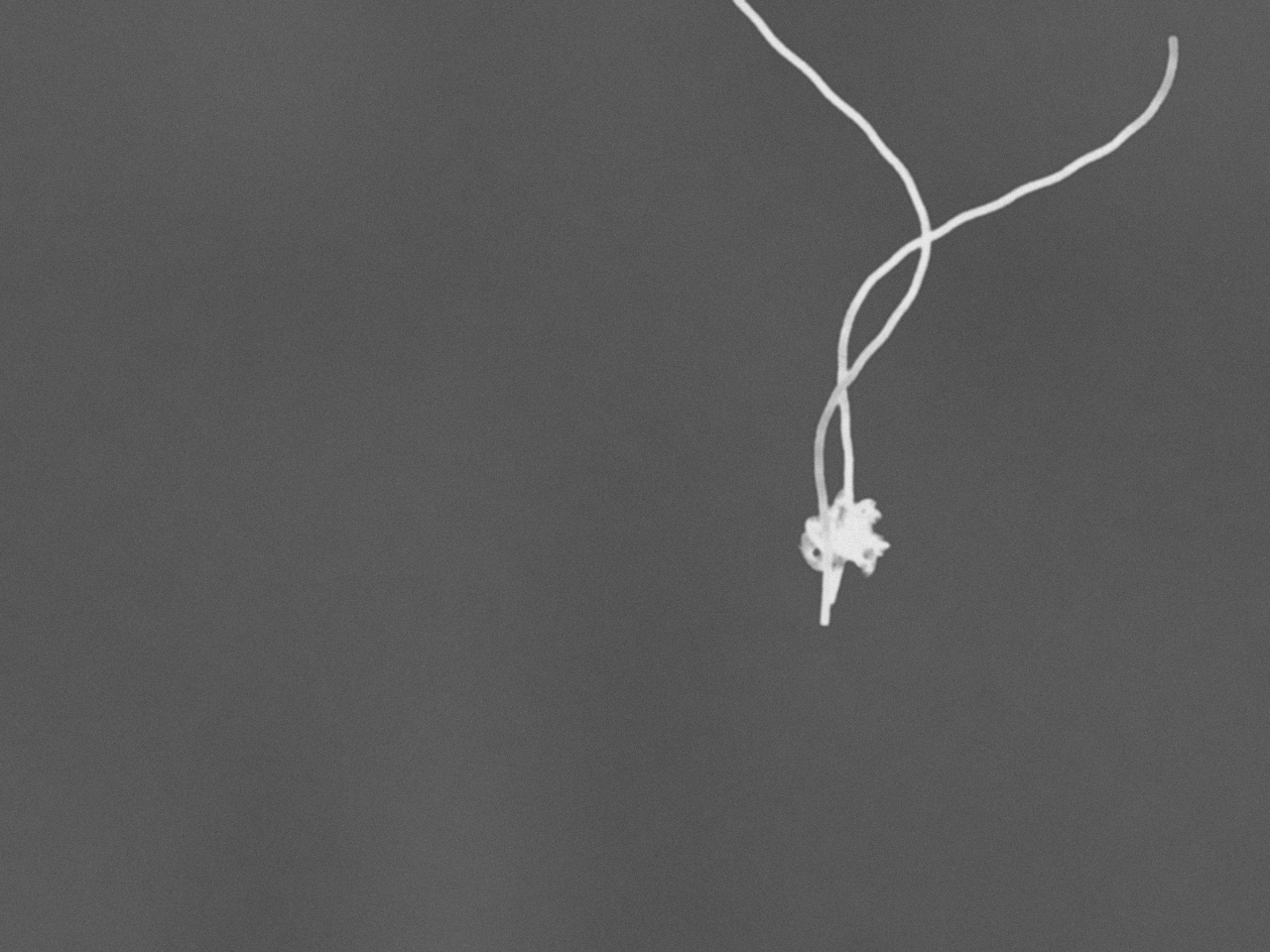} & \image{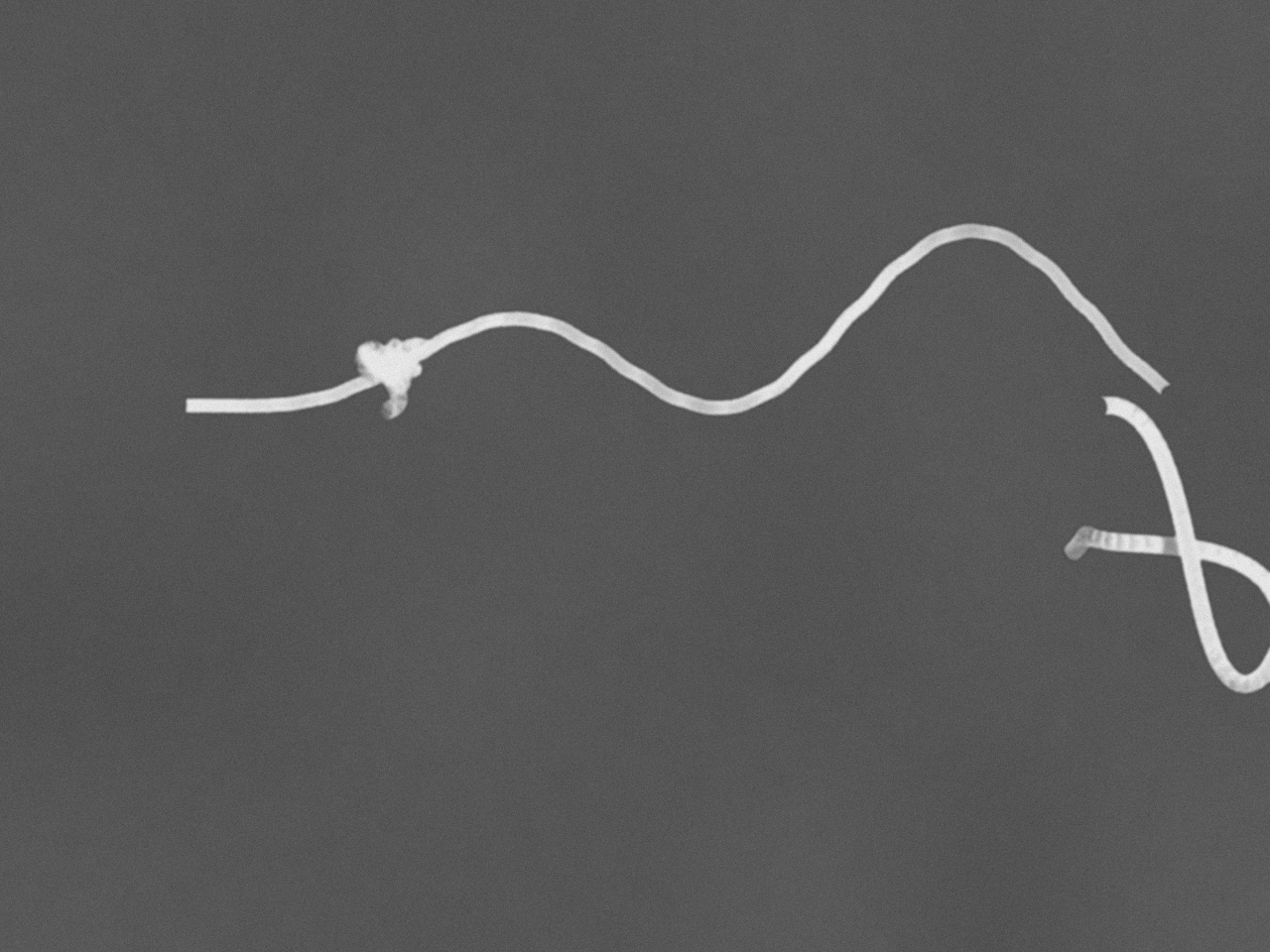} \\
		                      \multicolumn{2}{c}{\texttt{[-l|-c|-o]}$_\text{auto.}$}                       &          &                      \multicolumn{2}{c}{\texttt{[?l|?c|?o]}$_\text{synth.}$}
	\end{tabularx}
	\caption{Example images of the utilized data sets (\identifierlegend/).}
	\label{fig:DataSetComparison}
\end{figure*}
\graphicspath{{./figs/data/}}

\subsection{Ground Truth Generation}
\label{sec:Data-GroundTruthGeneration}
The ground truths\footnote{In machine learning, the ground truth, while not necessarily being perfect, is the best available data to test predictions of an algorithm.}, used to train and test the \glspl{CNN} utilized within this publication, have three origins: manual annotation, semiautomatic annotation and image synthesis.

\subsubsection{Manual Annotation}
\label{sec:Data-ManualAnnotation}
\begin{figure}
	\centering
	\includegraphics[width=\figurewidth]{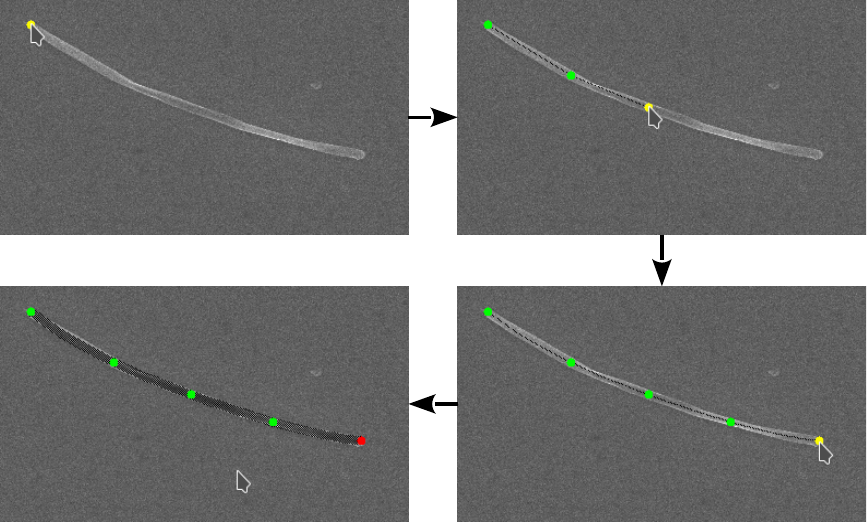}
	\caption{Illustration of the manual annotation process.}
	\label{fig:ManualAnnotation}
\end{figure}

A total of \num{1075} images, featuring \num{1935} instances (i.e. fibers), were annotated manually, using an ad hocly implemented annotation tool\footnote{Available at: \url{https://github.com/maxfrei750/FiberAnnotator}}. The manual annotation was done by selecting keypoints for each fiber that were interpolated using cubic splines and adjusting the fiber width until an optimal coverage was achieved (see \cref{fig:ManualAnnotation}).

\subsubsection{Semiautomatic Annotation}
\label{sec:Data-SemiautomaticAnnotation}
For basic fiber images, featuring neither clutter, loops nor overlaps (see \cref{sec:Data-DataSets}), a semiautomatic annotation can be carried out to avoid the laborious task of manual annotation.
\begin{figure}
	\centering
	\includegraphics[width=\figurewidth]{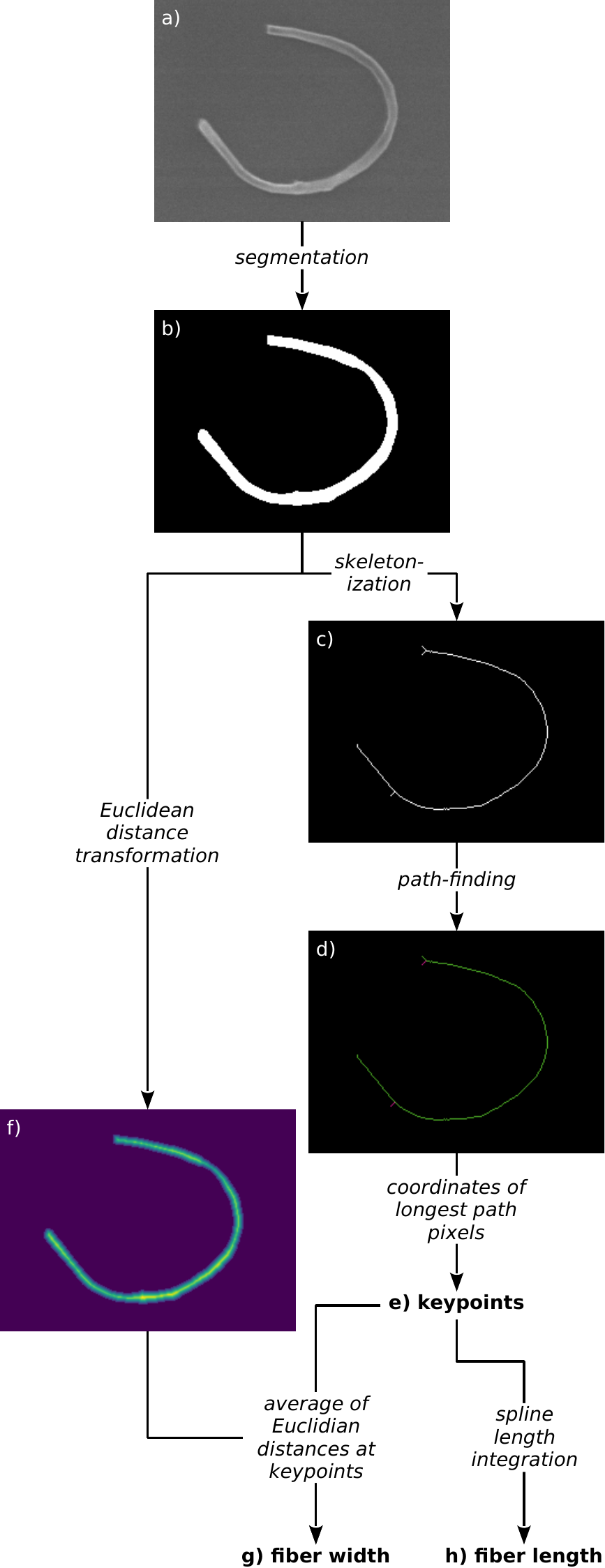}
	\caption{Illustration of the semiautomatic annotation process (see \cref{sec:Data-SemiautomaticAnnotation}).}
	\label{fig:SemiautomaticAnnotation}
\end{figure} 

For the use case at hand, the semiautomatic annotation was implemented as follows (see also \cref{fig:SemiautomaticAnnotation}):
\begin{enumerate}
	\item The original image (see \cref{fig:SemiautomaticAnnotation}a) is segmented using denoising and thresholding, yielding an instance mask (see \cref{fig:SemiautomaticAnnotation}b).
	\item The instance mask (see \cref{fig:SemiautomaticAnnotation}b) is skeletonized\footnote{During a skeletonization, the outmost \emph{true} pixels of a binary mask are removed repeatedly, until a further removal would separate previously connected regions. Effectively, a skeletonization reduces the thickness of a mask to \SI{1}{\px}.} (see \cref{fig:SemiautomaticAnnotation}c).
	\item To remove artifacts resulting from the skeletonization and to determine a correct order of keypoints, the longest connected path in the skeleton is identified via a path-finding method and all other pixels are discarded (see \cref{fig:SemiautomaticAnnotation}d; \colorsquare{77,172,38} pixels: kept, \colorsquare{208,28,139} pixels: discarded).
	\item The pixels of the longest connected path (see \cref{fig:SemiautomaticAnnotation}d; \colorsquare{77,172,38}\,pixels) are converted into keypoint coordinates (see \cref{fig:SemiautomaticAnnotation}e).
	\item To determine the fiber width (see \cref{fig:SemiautomaticAnnotation}g), an \name{Euclid}ian distance map\footnote{In a \name{Euclid}ian distance map, which results from the \name{Euclid}ian distance transformation of a mask, each pixel represents the \name{Euclid}ian distance of said pixel to the next background, i.e. \emph{false}, pixel in the input mask.} (see \cref{fig:SemiautomaticAnnotation}f) of the instance mask (see \cref{fig:SemiautomaticAnnotation}b) is calculated. Subsequently, the \name{Euclid}ian distances of the previously determined keypoints (see \cref{fig:SemiautomaticAnnotation}e) are looked up, their average is calculated and the resulting value is multiplied by a factor of \num{2} to yield the fiber width.
	\item The fiber length (see \cref{fig:SemiautomaticAnnotation}h) is determined by integrating a cubic spline interpolation of the keypoints (see \cref{fig:SemiautomaticAnnotation}e).
	\item Finally, annotations with faulty keypoints or fiber widths are removed manually.
\end{enumerate}
A total of \num{579} images, featuring \num{583} instances (i.e. fibers), were annotated semiautomatically.

\subsubsection{Image Synthesis}
\label{sec:Data-ImageSynthesis}
Using our \gls{synthPIC} toolbox\footnote{Available at: \url{https://github.com/maxfrei750/synthPIC4Python}}, \num{500} images, featuring 760 instances (i.e. fibers), were synthesized. The purpose of the synthetic images is to survey, whether they can be used to supplement or even replace real training data, thereby obliterating the need for a manual annotation.

\subsection{Data Sets}
\label{sec:Data-DataSets}
So far, three data sets were distinguished: manually annotated, semiautomatically annotated and synthetic fiber images (see \cref{sec:Data-GroundTruthGeneration}). However, the set of manually annotated images can be partitioned into subsets once again, based on the presence of potentially inhibiting factors for the automatic detection of fibers. A survey of the available images yielded three such factors: 
\begin{itemize}
	\item \emph{Loops}: Self-overlapping fibers.
	\item \emph{Clutter}: Agglomerates or aggregates of non-fiber particles which stick to fibers, e.g. nuclei that did not grow into long fibers.
	\item \emph{Overlaps}: Multiple fibers which overlap each other. Fibers which are connected only by clutter are not considered overlapping.
\end{itemize}
The set of manually annotated images was therefore subdivided into eight subsets, representing all possible combinations of the three inhibiting factors (see \cref{fig:DataSetComparison,tab:DataSetComparisonCombined}), to study their impact on the detection quality. Next, each real data set was partitioned once again, to yield training and test sets ($\lfloor \SI{85}{\percent}\rfloor/\lceil \SI{15}{\percent}\rceil$) for the proposed method (see \cref{tab:DataSetComparisonCombined}, right side). Ultimately, due to the small number of images for data sets featuring loops, all loop data sets (\texttt{[+l|...]}; see \cref{tab:DataSetComparisonCombined}, gray rows) were aggregated into a ninth data set (\texttt{[+l|?c|?o]}).


\begin{table*}
	\centering
	\caption{Data set properties (\identifierlegend/). Gray rows represent data sets before aggregation.}	
	\label{tab:DataSetComparisonCombined}
	\begin{tabu}{ccccc|ccc|ccc}
		\toprule\toprule
		                                             \multicolumn{5}{c|}{}                                               &   \multicolumn{3}{c|}{\textbf{Number of Images}}   &  \multicolumn{3}{c}{\textbf{Number of Fibers}}  \\
		\textbf{Loops} & \textbf{Clutter} & \textbf{Overlaps} & \textbf{Annotation} &        \textbf{Identifier}         & \textbf{Total} & \textbf{Training} & \textbf{Test} & \textbf{Total} & \textbf{Training} & \textbf{Test} \\
		      no       &        no        &        no         &       manual        &        \texttt{[-l|-c|-o]}         &   \num{236}    &     \num{200}     &   \num{36}    &   \num{250}    &     \num{211}     &   \num{39}    \\
		      no       &        no        &        yes        &       manual        &        \texttt{[-l|-c|+o]}         &   \num{107}    &     \num{90}      &   \num{17}    &   \num{227}    &     \num{192}     &   \num{35}    \\
		      no       &       yes        &        no         &       manual        &        \texttt{[-l|+c|-o]}         &   \num{351}    &     \num{298}     &   \num{53}    &   \num{444}    &     \num{379}     &   \num{65}    \\
		      no       &       yes        &        yes        &       manual        &        \texttt{[-l|+c|+o]}         &   \num{337}    &     \num{286}     &   \num{51}    &   \num{1014}   &     \num{855}     &   \num{159}   \\ \midrule
		     yes       &      random      &      random       &       manual        &        \texttt{[+l|?c|?o]}         &    \num{44}    &     \num{35}      &    \num{9}    &    \num{75}    &     \num{59}      &   \num{16}    \\ \rowfont{\color{gray}}
		     yes       &        no        &        no         &       manual        &        \texttt{[+l|-c|-o]}         &    \num{22}    &     \num{18}      &    \num{4}    &    \num{23}    &     \num{19}      &    \num{4}    \\
		     \rowfont{\color{gray}}
		     yes       &        no        &        yes        &       manual        &        \texttt{[+l|-c|+o]}         &    \num{2}     &      \num{1}      &    \num{1}    &    \num{5}     &      \num{3}      &    \num{2}    \\
		     \rowfont{\color{gray}}
		     yes       &       yes        &        no         &       manual        &        \texttt{[+l|+c|-o]}         &    \num{11}    &      \num{9}      &    \num{2}    &    \num{13}    &     \num{11}      &    \num{2}    \\
		     \rowfont{\color{gray}}
		     yes       &       yes        &        yes        &       manual        &        \texttt{[+l|+c|+o]}         &    \num{9}     &      \num{7}      &    \num{2}    &    \num{34}    &     \num{26}      &    \num{8}    \\ \midrule
		      no       &        no        &        no         &    semiautomatic    & \texttt{[-l|-c|-o]}$_\text{auto.}$  &   \num{579}    &     \num{492}     &   \num{87}    &   \num{583}    &     \num{496}     &   \num{87}    \\
		    random     &      random      &      random       &      synthetic      & \texttt{[?l|?c|?o]}$_\text{synth.}$ &   \num{425}    &     \num{425}     &   --   &   \num{645}    &     \num{645}     &   --  \\	    
 \bottomrule
	\end{tabu}
\end{table*}%

	\graphicspath{{./figs/method/}}
\section{Method}
\label{sec:Method}
The focus of the proposed method lies on the modification of already existing \gls{R-CNN} architectures (see \cref{sec:Method-NetworkArchitecture,sec:Method-LossFunction}) and training schedules (see \cref{sec:Method-Training}), to meet the requirements of imaging fiber analysis. Furthermore, the proposed extensions require changes with respect to the preparation of the utilized input data (see \cref{sec:Method-DataTransformations}) and allow for custom-designed error detection and correction strategies (see \cref{sec:Method-ErrorDetectionAndCorrection}). 

\subsection{Network Architecture}
\label{sec:Method-NetworkArchitecture}
\begin{figure}
	\centering
	\includegraphics[width=\figurewidth]{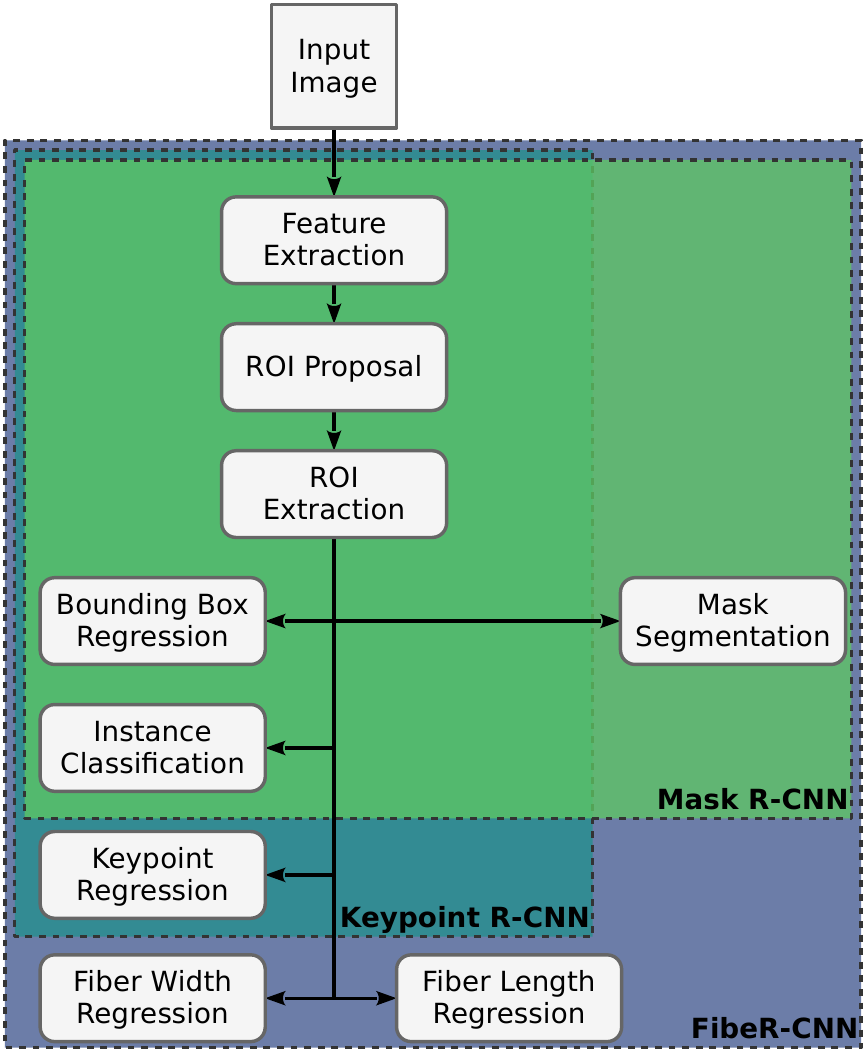}
	\caption{Architectures of \mrcnn/, \krcnn/ and \frcnn/.}
	\label{fig:RcnnOverview}
\end{figure}

The \frcnn/ architecture, presented within this publication, is an extension of the well-known \mrcnn/ architecture~\cite{He.2017} (see \cref{fig:RcnnOverview}; \colorsquare{97,181,115}\,box). It is therefore imperative to briefly elaborate upon the structure and general principles of \mrcnn/\footnote{For a more detailed, yet plain explanation please refer to~\cite{Frei.2020}. An in-depth explanation can be found in~\cite{He.2017}.}. Subsequently, it will be expanded in two steps: Firstly, by adding a head\footnote{Neural networks can consist of multiple branches, which perform independent tasks. The final part of a branch, which produces an output meaningful to the user, is referred to as head.} for keypoint regression\footnote{In a machine learning context, the term regression refers to the prediction of continuous values, e.g. keypoint coordinates.} (see \cref{fig:RcnnOverview}; \colorsquare{51,139,147}\,box), thereby yielding the \krcnn/ architecture~\cite{He.2017} and secondly, by adding two heads performing fiber width and length regressions, which ultimately yields the \frcnn/ architecture (see \cref{fig:RcnnOverview}; \colorsquare{108,125,168}\,box).
As codebase for the implementation, the \name{detectron2} framework~\cite{Wu.2019} was used, which features PyTorch~\cite{Paszke.2019} implementations of \mrcnn/ and \krcnn/. 

\subsubsection{Region-Based Convolutional Neural Networks}
\label{sec:Method-RCNNs}
\begin{figure}
	\centering
	\includegraphics[width=\figurewidth]{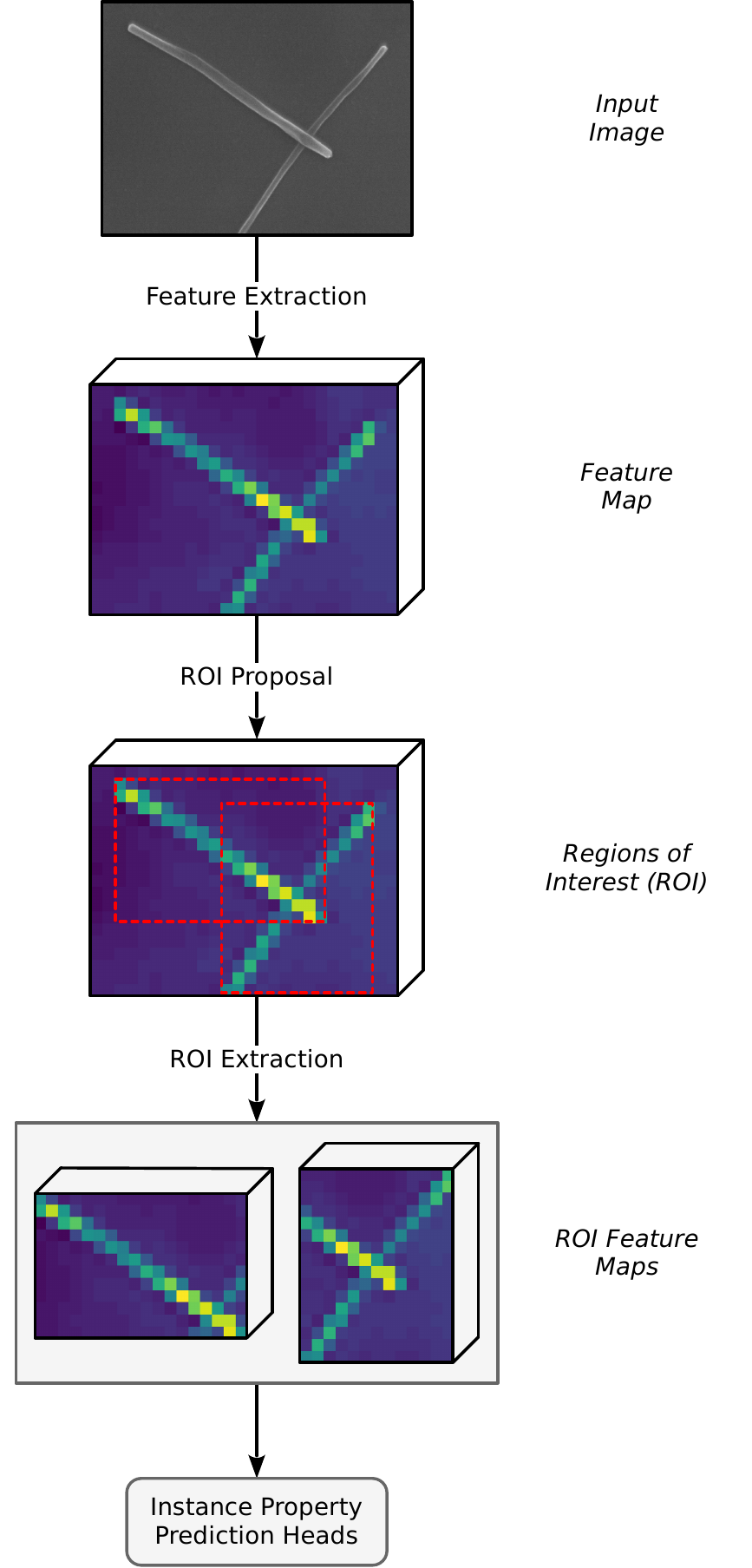}
	\caption{Illustration of the feature extraction as well as the \gls{ROI} proposal and extraction, which are shared by most modern \gls{R-CNN} architectures.}
	\label{fig:RcnnArchitecture}
\end{figure}

Modern \Glspl{R-CNN} consist of three conceptual stages (see \cref{fig:RcnnArchitecture}): feature extraction, \gls{ROI} proposal/extraction and instance property prediction.

\paragraph{Feature Extraction} The input image is processed by a \gls{CNN}, referred to as backbone, thereby extracting a map\footnote{Contrarily to ordinary maps, this map has more than two dimensions.} of prominent features over the entirety of the input image. Compared to the other architecture parts, the backbone is usually a much deeper network, i.e. it has more layers. Therefore, the majority of calculations takes place within the backbone. It is easily interchangeable to adjust the number of operations and thereby the network speed. In this publication, the convolutional blocks \numrange{2}{5} of the \name{ResNet-50} network~\cite{He.2015} are used as backbone.\footnote{For an elaboration upon the reasoning behind this design choice, please refer to~\cite{Frei.2020}.}

\paragraph{\glsentryshort{ROI} Proposal and Extraction} \glspl{ROI} encompassing instances are identified and extracted from the feature map. The \glspl{ROI} are selected so that each \gls{ROI} represents exactly one instance. Additionally, for each \gls{ROI} an objectness score, which quantifies the likelihood of the \gls{ROI} to encompass an object, is output. Subsequently, the set of instance feature maps is passed to each of the downstream heads, each of which predicts a desired instance property (e.g. class, bounding box, instance mask, keypoints, etc.).

\paragraph{Instance Property Prediction} \Glspl{R-CNN} can be distinguished based on the presence of characteristic instance property prediction heads (e.g. the mask segmentation or the keypoint regression head). However, most modern \Glspl{R-CNN} share at least two such heads: The bounding box regression head, which determines a refined bounding box for the instance in each \gls{ROI} and the instance classification\footnote{In a machine learning context, the term classification refers to the prediction of a discrete value, i.e. a class.} head, which determines the class of said instance. For the given application, the latter head is obsolete, because there is only a single class of instances. However, due to its negligible computational cost, it was not removed to facilitate future multi-class applications. 

All instance property prediction heads operate on the same shared set of instance feature maps. Therefore, the computational cost of adding additional heads is small compared to the backbone's computational cost. This is beneficial for the use-case at hand, since all extensions of \mrcnn/ within this publication come in the form of additional instance property prediction heads.

\subsubsection{\mrcnn/}
\label{sec:Method-MaskRCNN}
\begin{figure*}
	\centering
	\includegraphics[width=2\figurewidth]{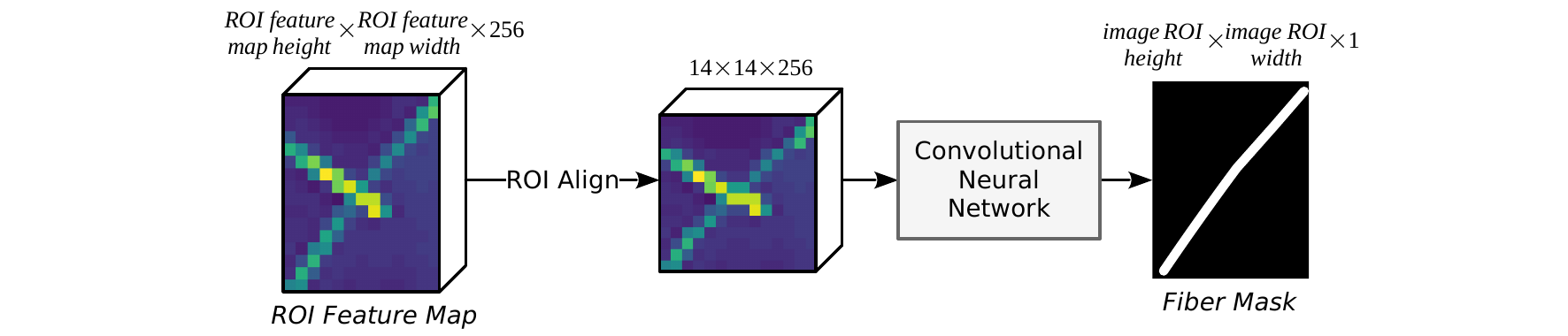}
	\caption{Illustration of the mask segmentation head.}
	\label{fig:MaskHead}
\end{figure*}

The characteristic instance property prediction head of \mrcnn/ is the mask segmentation head (see \cref{fig:RcnnOverview}; \colorsquare{97,181,115}\,box), which computes a binary mask representing the instance pixels, i.e. it answers the question, which pixels of the input image belong to a certain instance and which pixels belong to the image background or another instance. \cref{fig:MaskHead} illustrates the functionality of the mask segmentation: Initially, each \gls{ROI} feature map, resulting from the \gls{ROI} extraction, is resized using \gls{ROI} align\footnote{The details of \gls{ROI} align are beyond the scope of this publication. For an in-depth explanation, please refer to \cite{He.2017}.}. Subsequently, a \gls{CNN} upsamples the low-resolution, high-depth feature map to a high-resolution, low-depth binary mask.

\subsubsection{\krcnn/}
\label{sec:Method-KeyPointRCNN}
\begin{figure*}
	\centering
	\includegraphics[width=2\figurewidth]{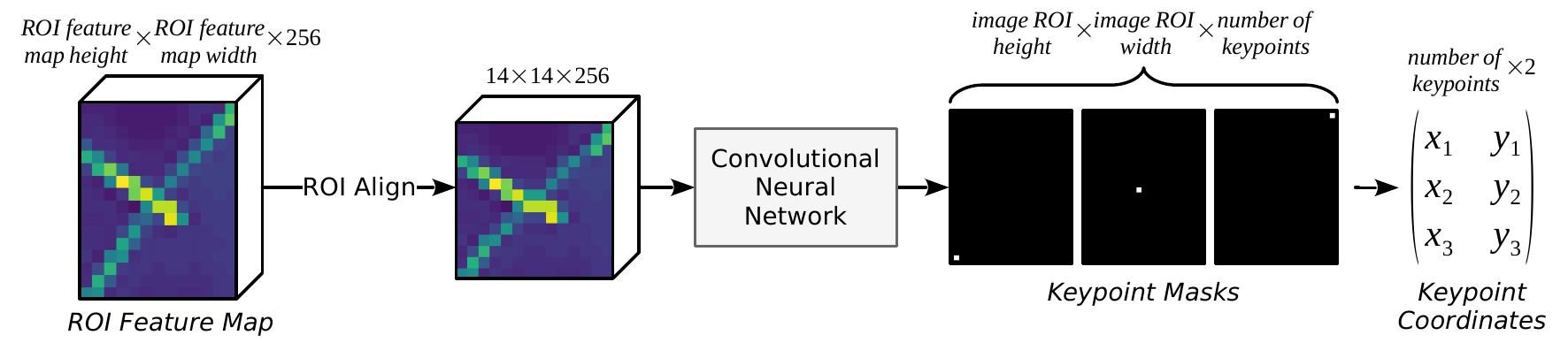}
	\caption{Illustration of the keypoint regression head (white pixels in keypoint masks are oversized).}
	\label{fig:KeypointHead}
\end{figure*}

The \krcnn/ architecture was proposed along with \mrcnn/ by \name{He et al.}~\cite{He.2017}, with the task of human pose estimation in mind. Instead of a mask segmentation head, it features a keypoint regression head (see \cref{fig:RcnnOverview}; \colorsquare{51,139,147}\,box). The functionality of this head (see \cref{fig:KeypointHead}) is closely related to that of the mask segmentation head (see \cref{fig:MaskHead}), with the key difference being that multiple (keypoint) masks per instance are predicted, instead of just a single mask. In each keypoint mask, there exists only a single \emph{true} pixel, which represents the keypoint position.

In the original implementation, the keypoint regression head outputs the coordinates of \num{17} keypoints, which is insufficient to describe the shapes of long and/or strongly curved fibers. Therefore, in the \frcnn/ architecture, the keypoint regression head was altered to output 40 keypoint coordinates (see also \cref{sec:Method-NumberOfKeypoints}), by increasing the respective dimension of its last layer.

\subsubsection{\frcnn/}
\label{sec:Method-FibeRCNN}
\frcnn/ expands \mrcnn/ beyond \krcnn/ by adding two additional instance property prediction heads (see \cref{fig:RcnnOverview}; \colorsquare{108,125,168}\,box): the fiber width and length regression heads.

\begin{figure*}
	\centering
	\includegraphics[width=2\figurewidth]{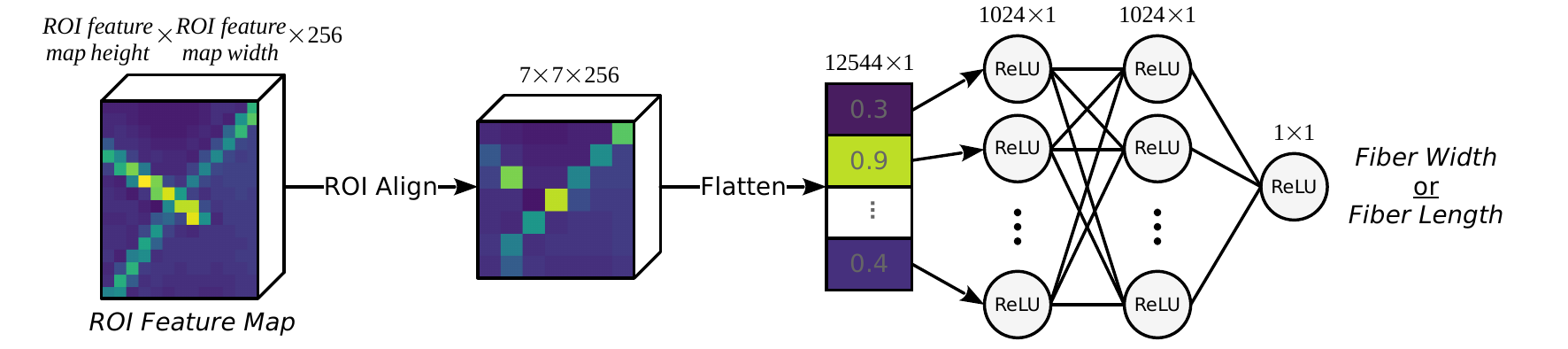}
	\caption{Architecture of the fiber width and fiber length regression head, respectively.}
	\label{fig:FiberHeads}
\end{figure*}

The architectures of these heads were inspired by the bounding box regression head of \mrcnn/~\cite{He.2017}, i.e. they are implemented as fully connected neural networks\footnote{The term fully connected neural networks is used to distinguish simple neural networks with scalar weights, where each neuron of a layer is connected to all neurons of the previous and the following layer, from \glspl{CNN}.}, which each consist of three \gls{ReLU} layers\footnote{Artificial neurons usually use non-linear activation functions to calculate their output. The \glsentryfull{ReLU} function is the simplest non-linear function: $\text{ReLU}(x)=\max(0,x)$} (see \cref{fig:FiberHeads}). As inputs for the fully connected neural networks, resized and flattened versions of the input \gls{ROI} feature maps are used. In contrast to the mask and keypoint prediction heads, the fiber width and length regression heads -- just like the bounding box regression head -- operate on lower-resolution versions of the utilized \gls{ROI} feature maps, to reduce the size and complexity of the utilized fully connected neural network. During the flattening, each multidimensional \gls{ROI} feature map is transformed into a vector by concatenating all of its elements. Subsequently, each element is being fed to a corresponding input neuron of the downstream fully connected neural network, which, as a whole, predicts the fiber width or length, respectively. 

Due to the fact that the fiber width, as well as the fiber length regression head both only output a single quantity, each of them only features a single output neuron. While the prediction of a single length per fiber is intuitive, the prediction of just a single width per fiber is arbitrary and tailored to the utilized data, which features only fibers with various, yet constant widths. However, for fibers with inconstant widths, the architecture could easily be expanded, by adding more output neurons to the fiber width regression branch, to predict an individual width at every keypoint. 

At first glance, the mask segmentation head inherited from the \mrcnn/ architecture and the fiber length head may seem obsolete. After all, a fiber mask can also be attained by performing a cubic spline interpolation of the detected fiber keypoints and drawing this spline with lines having the predicted fiber width. Similarly, the fiber length can be determined by performing an integration of the cubic spline approximation. However, actually, these redundancies are indeed useful, because they enable the application of error detection and correction strategies (see \cref{sec:Method-ErrorDetectionAndCorrection}), to improve the detection accuracy. 

\subsection{Loss Function}
\label{sec:Method-LossFunction}
The loss function -- sometimes also referred to as cost function -- is an essential element of many optimization problems, such as the training of neural networks. It is a means to quantify the quality of a model, based on the deviation of its predictions from the ground truth, i.e. the desired target outputs. The higher the deviation, the higher the loss. Therefore, the goal of the training is to minimize this loss.

\frcnn/ uses a multi-task loss \gls{symb:loss}, which is equal to the sum of the individual prediction head losses:
\begin{equation}
	\gls{symb:loss} = \gls{symb:lossClass} + \gls{symb:lossBox} + \gls{symb:lossMask} + \gls{symb:lossKeypoint} + \gls{symb:lossWidth} + \gls{symb:lossLength}
\end{equation}
\gls{symb:lossClass} and \gls{symb:lossBox} are the instance classification and bounding box regression head losses~\cite{Girshick.2015}, whereas \gls{symb:lossMask} and \gls{symb:lossKeypoint} are the mask segmentation and keypoint regression head losses~\cite{He.2017}.

The fiber width and length prediction head losses \gls{symb:lossWidth} and \gls{symb:lossLength} are both based on the \gls{MSE}:
\begin{equation}
	\gls{symb:lossWidth}(\gls{symb:prediction},\gls{symb:target}) = \gls{symb:lossWeightWidth} \cdot \gls{symb:meanSquaredError}(\gls{symb:prediction},\gls{symb:target})
\end{equation}
\begin{equation}
\gls{symb:lossLength}(\gls{symb:prediction},\gls{symb:target}) = \gls{symb:lossWeightLength} \cdot \gls{symb:meanSquaredError}(\gls{symb:prediction},\gls{symb:target})
\end{equation}
where
\begin{equation}
\gls{symb:meanSquaredError}(\gls{symb:prediction},\gls{symb:target}) = \frac{1}{\gls{symb:numberOfDates}}\sum_{\gls{symb:index}=1}^{\gls{symb:numberOfDates}}(\gls{symb:prediction}_{\gls{symb:index}} - \gls{symb:target}_{\gls{symb:index}})^2
\end{equation}
is the \glsentrylong{MSE}, \gls{symb:prediction} and \gls{symb:target} are the prediction and target vectors\footnote{\glspl{ROI} are usually processed in batches to take advantage of parallelization. Therefore, the properties of more than one instance are predicted simultaneously.} of the respective heads, \gls{symb:index} is the index of each instance and \gls{symb:numberOfDates} is the number of dates, i.e. instances. 

The main difference of the losses are their weights $\gls{symb:lossWeightWidth}=\num{e-3}$ and $\gls{symb:lossWeightLength}=\num{e-6}$, which where chosen so that \gls{symb:loss} is not dominated by the fiber width and length regression heads. In practice, the weights $\gls{symb:lossWeightWidth}$ and $\gls{symb:lossWeightLength}$  were adjusted, so that all prediction head losses have a similar magnitude at the beginning of the training:
\begin{equation}
\gls{symb:lossClass} \sim \gls{symb:lossBox} \sim \gls{symb:lossMask} \sim \gls{symb:lossKeypoint} \sim \gls{symb:lossWidth} \sim \gls{symb:lossLength}
\end{equation}

Otherwise, the neural network would focus primarily on the improvement of the fiber width and length prediction heads heads and neglect the other heads during the training. This could for instance result in a network which can very reliably predict fiber widths and lengths, but not bounding boxes or keypoints.

\subsection{Data Transformations}
\label{sec:Method-DataTransformations}
To homogenize or augment the input and ground truth data of \glspl{CNN}, it is often useful or even mandatory to apply transformations to it. 

\subsubsection{Number of Keypoints}
\label{sec:Method-NumberOfKeypoints}
As mentioned in \cref{sec:Method-KeyPointRCNN}, the number of keypoints per instance, predicted by the keypoint regression head of \frcnn/, had to be adjusted, to yield a high enough resolution of keypoints, to describe the shape of long and/or strongly curved fibers. Also, the calculation of the keypoint regression head loss requires a consistent number of keypoints in the ground truths. 

While a higher number of keypoints yields a better resolution, as with all statistical models, it is undesirable to introduce more degrees of freedom (i.e. keypoints) than necessary. Therefore, approximations of the ground truth, using varying numbers of keypoints were tested and the resulting approximation qualities were assessed using the \gls{BIC}\footnote{The \gls{BIC} is a commonly used metric for the evaluation of statistical models.}.

According to \name{Yaffee} and \name{McGee} \cite{Yaffee.2000}, the \gls{BIC} (omitting the \name{Bessel}'s correction~\cite{Radziwill.2017}) is defined as:
\begin{equation}
	\gls{symb:bayesianInformationCriterion}(\gls{symb:prediction},\gls{symb:target}) = \gls{symb:numberOfDates}\cdot\ln\left(\frac{1}{\gls{symb:numberOfDates}} \cdot \gls{symb:sumOfSquaredResiduals}(\gls{symb:prediction},\gls{symb:target})\right) + \gls{symb:numberOfKeypoints}\cdot\ln(\gls{symb:numberOfDates}),
\end{equation}
where, for the case at hand, \gls{symb:prediction} and \gls{symb:target} are the coordinate matrices of the approximated and ground truth keypoints, respectively, whereas \gls{symb:numberOfKeypoints} is the tested number of keypoints, \gls{symb:sumOfSquaredResiduals} is the \glsentrylong{SSR} and \gls{symb:numberOfDates} is the number of dates, i.e. the arbitrary number of sampled squared residuals (for the study at hand: $\gls{symb:numberOfDates}=\num{200}$).

\begin{figure}
	\centering
	\includegraphics[width=\figurewidth]{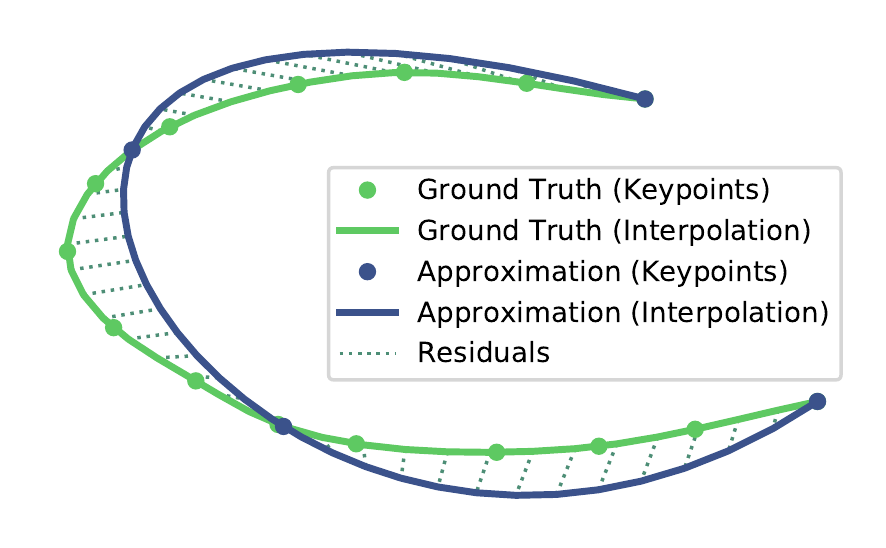}
	\caption{Illustration of the \glsentryfull{SSR} of a keypoint approximation versus the associated ground truth.}
	\label{fig:SplineResiduals}
\end{figure}

The \gls{SSR} quantifies the error of the tested approximation. It equals the sum of squared distances of pairs of points, sampled from two uniform cubic spline interpolations, one through the ground truth keypoints and the other one through the approximated keypoints (see \cref{fig:SplineResiduals}):
\begin{equation}
\gls{symb:sumOfSquaredResiduals}(\gls{symb:prediction},\gls{symb:target}) = \sum_{\gls{symb:index}=1}^{\gls{symb:numberOfDates}}\sum_{\gls{symb:index2}=1}^{2}(\gls{symb:prediction}_{\gls{symb:index}, \gls{symb:index2}} - \gls{symb:target}_{\gls{symb:index}, \gls{symb:index2}})^2, 
\end{equation}
where \gls{symb:index} is the sampling point index and \gls{symb:index2} is the coordinate index (i.e. whether the $x$- or the $y$-coordinate of the sampled point is used).
The \gls{BIC} rewards good approximations, while punishing the introduction of additional parameters. Therefore, lower \glspl{BIC} indicate better balanced models~\cite{Yaffee.2000}. 

\begin{figure}
	\centering
	\includegraphics[width=\figurewidth]{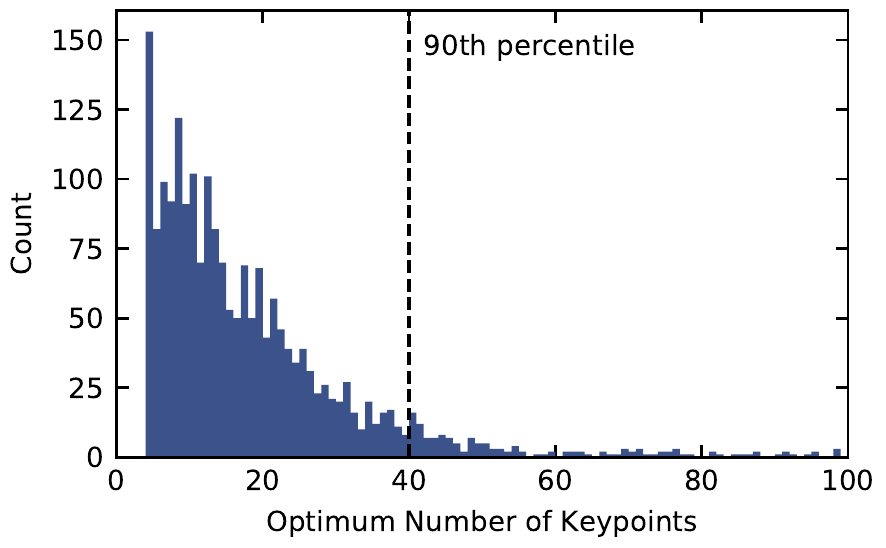}
	\caption{Distribution of the optimum number of keypoints for the surveyed fibers.}
	\label{fig:OptimumNumberOfKeypointsDistribution}
\end{figure}

To determine the optimum number of keypoints per fiber, the ground truth keypoints of each fiber of the non-synthetic training data sets\footnote{The test data sets were excluded to prevent a possible bias.} were interpolated using uniform cubic splines, having numbers of knots in the range from \numrange{4}{100}. Subsequently, the resulting \gls{BIC} was calculated for each interpolation. The optimum number of keypoints of each fiber, was defined as the number of knots yielding the minimum \gls{BIC} for each fiber. \cref{fig:OptimumNumberOfKeypointsDistribution} depicts the resulting distribution of optimum number of keypoints for the surveyed fibers. As overall optimum number of keypoints, the \nth{90} percentile of this distribution was chosen, which yields a number of \num{40} keypoints as optimum for the given data.
Accordingly, the last layer of the keypoint regression head of \frcnn/ was dimensioned to output the coordinates of \num{40} keypoints and all ground truths were transformed to have \num{40} keypoints, by using uniform cubic spline interpolation. 

\subsubsection{Keypoint Ordering}
\label{sec:Method-KeypointOrdering}

The order of keypoints is crucial for the keypoint regression head loss, because the position of a keypoint within the list of keypoints has an implicit meaning, in form of a label. In human pose estimation, these labels are i.a. \emph{left hand}, \emph{right hand}, \emph{head}, etc. Therefore, it is evident that a keypoint -- even if it has the perfect spacial location -- is plainly wrong, if it is mislabeled, i.e. it has the wrong position in the list of keypoints (see \cref{fig:MisplacedKeypoint}). A keypoint labeled as \emph{left hand} at the spacial location of a head is just as wrong as one labeled as \emph{left hand} at another, e.g. random, spacial location\footnote{Unless of course, it coincidentally \emph{is} at the spatial location of a left hand.}.

For the use-case at hand, this brings about a problem with respect to the training data annotation. While during the annotation of human poses, each keypoint is unique and even similar keypoints such as \emph{left hand} and \emph{right hand} can be distinguished reliably (even if the human does not face the camera), this is no longer true for the annotation of fibers. To the human eye, both ends of a fiber are indistinguishable. Therefore, the annotations are inconsistent, which severely impedes the performance of \frcnn/. To solve this problem by establishing consistency, the keypoints need to be ordered according to a rule. Due to the fact that the relative keypoint order is already correct, the rule only has to address the fiber end keypoints.

\begin{figure}
	\centering
	\includegraphics[width=\figurewidth]{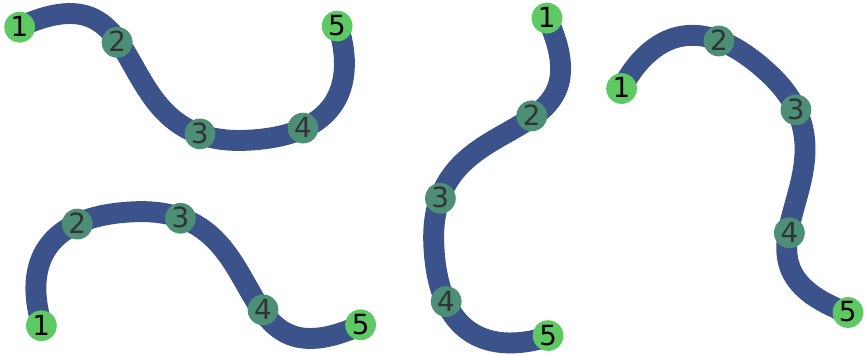}
	\caption{Examples of keypoint orders according to the \enquote{top to bottom, left to right} rule, resulting from different orientations of a fiber.}
	\label{fig:KeypointOrdering}
\end{figure}

A simple rule to order the fiber end keypoints, is to order them as if they were words in an English book, i.e. from top to bottom and from left to right (see also \cref{fig:KeypointOrdering}):
\begin{displayquote}
	Choose the topmost end point of a fiber as the first keypoint. If there are two candidates, choose the leftmost candidate as first keypoint.
\end{displayquote}

\subsubsection{Input Augmentation}
\label{sec:Method-InputAugmentation}

A widespread augmentation technique, to artificially increase the amount of training data, is the flipping of images. Since \krcnn/ is traditionally used for human pose estimation, it usually does only apply horizontal but no vertical flipping. However, since for microscopic images, there is no notion of up and down, \frcnn/ offers the possibility to apply vertical as well as horizontal flipping. To have a consistent order of keypoints, independent from the flipping, the flipping takes place before the keypoint ordering (see \cref{sec:Method-KeypointOrdering}).

To further increase the input data variance, the contrast and brightness of input images is varied randomly. An overview of the utilized input augmentation parameters is given in \cref{tab:InputAugmentationParameters}.
\begin{table}
	\centering
	\caption{Input augmentation parameters.}	
	\label{tab:InputAugmentationParameters}
	\begin{tabularx}{\linewidth}{CZ}
		\toprule\toprule
		Flip: left-right & \SI{50}{\percent} chance \\
		 Flip: up-down   & \SI{50}{\percent} chance \\
		    Contrast     & \numrange{0.5}{1.5}      \\
		   Brightness    & \numrange{0.5}{1.5}      \\ \bottomrule
	\end{tabularx}
\end{table}%

\subsection{Error Detection and Correction}
\label{sec:Method-ErrorDetectionAndCorrection}

\begin{figure}
	\centering
	\includegraphics[width=\figurewidth]{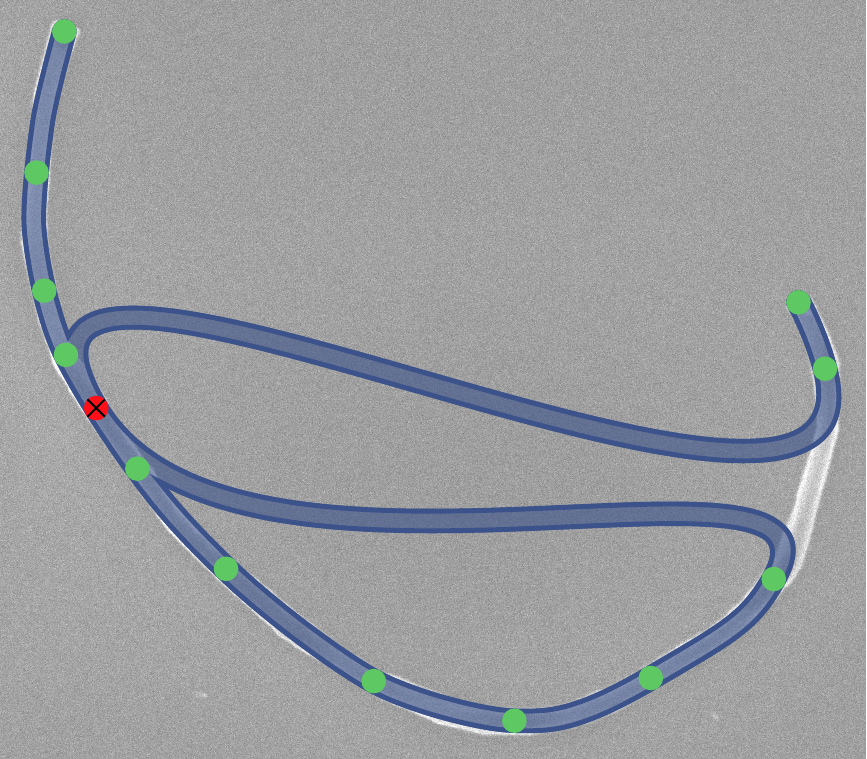}
	\caption{Illustration of the effect of a misplaced keypoint on a keypoint-based fiber mask.}
	\label{fig:MisplacedKeypoint}
\end{figure}

The creation of fiber masks (see \cref{sec:Method-FibeRCNN}) on the basis of fiber widths and keypoints is susceptible to misplaced keypoints, i.e. a single misplaced keypoint can lead to large errors (see \cref{fig:MisplacedKeypoint}). However, as mentioned in \cref{sec:Method-FibeRCNN}, the retrieval of redundant information about the detected fiber instances enables the use of error detection and even correction strategies.

\subsubsection{Error Detection}
\label{sec:Method-ErrorDetection}
Since the fiber length prediction is much easier than the keypoint prediction, it is much more precise (see \cref{sec:Results-Application}). Therefore, errors during the keypoint detection can easily be detected by comparing the length of the cubic spline interpolation of the predicted keypoints, to the length predicted by the fiber length regression head. Under the assumption, that the fiber length predictions are significantly more reliable than the keypoint predictions, this yields a possibility to reliably quantify the overall keypoint prediction quality.

An alternative error detection strategy is the comparison of the keypoint-based fiber masks with the masks output by the mask segmentation head, by calculating their \gls{IoU}\footnote{Intersection over union is a measure to determine the likeness of a pair of two-dimensional objects, with respect to their size, shape and position (see also \cref{fig:IoU}). As the name implies, it is defined as the ratio of the area of intersection and the area of the union of the two objects.}. However, since the latter are usually quite ragged, i.e. imperfect themselves (see \cref{fig:Motivation}), only relative statements about the prediction quality can be made: a higher agreement of both masks indicates a better prediction. However, in contrast to the fiber length based error detection strategy, this gives no absolute measure for the prediction quality.

\subsubsection{Error Correction}
\label{sec:Method-ErrorCorrection}
As illustrated in \cref{fig:MisplacedKeypoint}, misplaced keypoints can lead to large errors during the detection of fibers. We therefore propose an error correction strategy, hereby named keypoint pruning. During the keypoint pruning, individual keypoints are removed and it is tested whether the removal improves the detection quality (see \cref{app:alg:KeypointPruning}). As measures for the quality, the strategies presented in \cref{sec:Method-ErrorDetection}, i.e. fiber mask \gls{IoU} and length deviation, are used. 

A drawback of keypoint pruning is its high computational cost, due to the large number of keypoint combinations to be tested and the repetitive calculations of the fiber mask \gls{IoU}. Fortunately, it is possible to retroactively apply the error correction to an already trained \frcnn/, so that the training speed is not impeded by the error correction.

\subsection{Training}
\label{sec:Method-Training}
The schedule, used for the training of \frcnn/\footnote{Whenever \mrcnn/ was used as a comparison during the design and evaluation of \frcnn/, both networks were trained using the same learning rate schedule, to maintain the comparability between both models.}, is based on the well-established \name{$3\times$ training schedule}~\cite{He.2017,He.2018}, which was developed to train \mrcnn/ and \krcnn/ on the \gls{COCO} data set~\cite{Lin.2014}. To adapt it to the simpler learning task and speed up the training, its step size and the training duration were reduced by a factor of \num{10}. 

\begin{figure}
	\centering
	\includegraphics[width=\figurewidth]{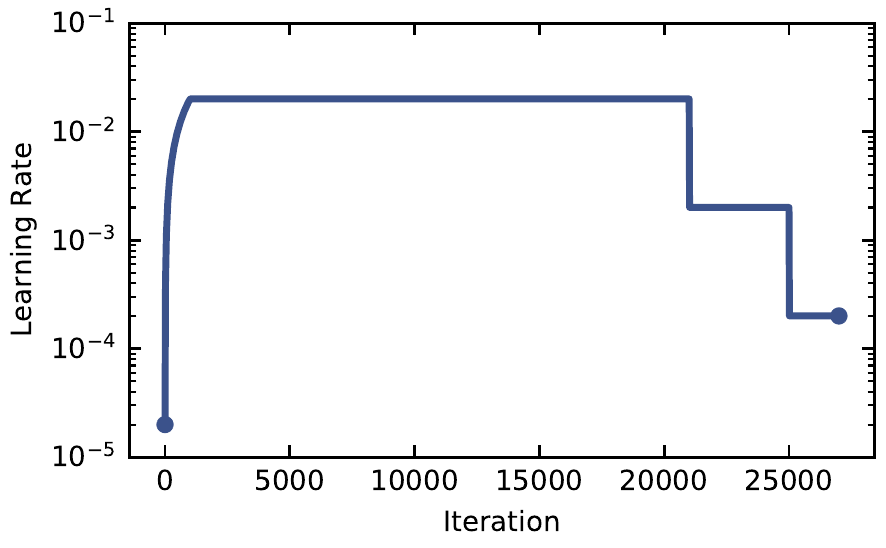}
	\caption{Learning rate schedule.}
	\label{fig:LearningRateSchedule}
\end{figure}

\cref{fig:LearningRateSchedule} depicts the learning rate schedule. The base learning rate is $\gls{symb:baseLearningRate}=\num{0.02}$. However, the training begins with a warm-up period of \num{1000} iterations, during which the learning rate is increased linearly from $\num{0.001} \cdot \gls{symb:baseLearningRate}$ to \gls{symb:baseLearningRate}. After \num{21000} and \num{25000} iterations respectively, the learning rate is reduced by a factor of \num{10}. The training ends after \num{27000} iterations. A complete overview of all training hyperparameters can be found in \cref{tab:TrainingHyperparameters}.
\begin{table}
	\centering
	\caption{Training hyperparameters.}	
	\label{tab:TrainingHyperparameters}
	\begin{tabularx}{\linewidth}{CZ}
		\toprule\toprule
		         Solver           & \gls{SGDM}~\cite{Murphy.2012}         \\
		   Base Learning Rate     & \num{0.02}                            \\
		        Momentum          & \num{0.9}                             \\
		     Warm-Up Factor       & \num{0.001}                           \\
		     Warm-Up Period       & \num{1000} iterations                 \\
		Learning Rate Drop Steps  & \num{21000} \& \num{25000} iterations \\
		Learning Rate Drop Factor & \num{0.1}                             \\
		        Duration          & \num{27000} iterations                \\
		       Batch Size         & \num{64} (\num{16} per \gls{GPU})     \\ \bottomrule
	\end{tabularx}
\end{table}%
To further speed up the training, transfer learning was utilized, by initializing the weights of the feature extraction network, with weights of a \name{ResNet-50}-based \krcnn/, which was trained on the \gls{COCO} data set according to the \name{$3\times$ training schedule} and is included with the \name{detectron2} framework.

The training was carried out on a dedicated \gls{GPU} server (see \cref{app:tab:Software,app:tab:Hardware}).

	\graphicspath{{./figs/results/}}
\section{Results}
\label{sec:Results}
There are two kinds of results from the studies carried out for this publication. Firstly, there are the results and insights, produced during the design of the \frcnn/ architecture (see \cref{sec:Results-ArchitectureDesign}). Secondly, there are the results and their implications with respect to the use of the \frcnn/ architecture for imaging particle analysis applications (see \cref{sec:Results-Application}).\footnote{The source code of the \frcnn/ architecture, the final model and the data sets, used for its training and testing, are available via the following link:\\ \url{https://github.com/maxfrei750/FibeR-CNN/releases/v1.0} \\Additionally, the training and test data sets are part of the \name{BigParticle.Cloud} (\url{https://bigparticle.cloud}).}

\paragraph{Statistics Note} Each error bar in this section represents the \SI{95}{\percent} \gls{CI} of a result, based on \num{3} repetitions, using varying random seeds during the training of the respective neural network.

\subsection{Architecture Design}
\label{sec:Results-ArchitectureDesign}
For the systematic design of the \frcnn/ architecture, an additive approach was used. Starting with a basic implementation, as close as possible to \mrcnn/ and \krcnn/, the architecture was improved and enhanced piece by piece. After each addition, the new model's quality was compared to that of the previous model version. 
\enlargethispage{1\baselineskip}
As quality measures, \glspl{AP} at multiple \gls{symb:intersectionoverUnion} thresholds were used, according to the \gls{COCO} evaluation scheme~\cite{COCOConsortium.2019}:
\begin{itemize}
	\item \gls{symb:meanAveragePrecision}: mean of \glspl{AP} with \gls{IoU} thresholds in the range from \SIrange{50}{95}{\percent} with increments of \SI{5}{\percent}
	\item \gls{symb:averagePrecision50}: \gls{AP} at an \gls{IoU} threshold of \SI{50}{\percent}
	\item \gls{symb:averagePrecision75}: \gls{AP} at an \gls{IoU} threshold of \SI{75}{\percent}
\end{itemize}

As test data, a collection of all available real test data sets was used. In case of conflicting quality indications from the different \glspl{AP}, the \gls{mAP} was used as basis for the final decision.

\begin{figure}
	\centering
	\includegraphics[width=\figurewidth]{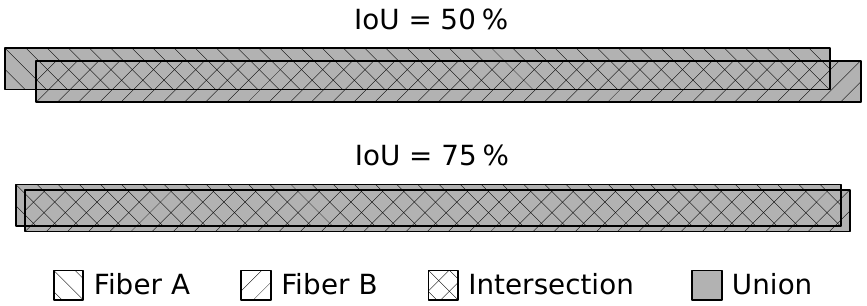}
	\caption{Illustration of the \glsentryfull{IoU} metric.}
	\label{fig:IoU}
\end{figure}

\paragraph{Average Precision}
Since the \gls{AP} is such a central concept for the evaluation of the architecture design, it shall be briefly elaborated upon. According to \name{Padilla et al.} \cite{Padilla.2020}, object detections can be grouped into four categories: 
\begin{itemize}
	\item \emph{\gls{FP}}: erroneous detection that does not encompass a sought-after object
	\item \emph{\gls{TP}}: correct detection that encompasses a sought-after object
	\item \emph{\gls{FN}}: sought-after object that has not been detected
	\item \emph{\gls{TN}}: detection that has not been detected because there was no sought after object
\end{itemize}
To evaluate, which of these four categories a prediction belongs to, it is necessary to define a criterion to match pairs of detections and ground truths. A common criterion is the \gls{IoU} (see \cref{fig:IoU}). Detections and ground truths that feature an \gls{IoU} greater than or equal to a certain threshold (e.g. $\gls{symb:intersectionoverUnion}\geq\SI{50}{\percent}$, which yields \gls{symb:averagePrecision50} or $\gls{symb:intersectionoverUnion}\geq\SI{75}{\percent}$, which yields \gls{symb:averagePrecision75}), are defined to be matches. For one-class detections like the application at hand, matching pairs of detections and ground truths are \glspl{TP}. If there is no matching detection for a ground truth, then it is counted as a \gls{FN}. Contrarily, if there is no matching ground truth for a detection, then it is a \gls{FP}. If there are multiple matches for a ground truth, then the first match is counted as \gls{TP}, while the rest is counted as \glspl{FN}.

Two basic metrics for object detection applications are precision \cite{Padilla.2020}:
\begin{equation}
	\text{precision} = 
		\frac{\text{\# \glspl{TP}}}{\text{\# \glspl{TP} + \# \glspl{FP}}} = 
		\frac{\text{\# \glspl{TP}}}{\text{\# detections}},
\end{equation}
i.e. the probability of the detector to yield a true positive, and recall \cite{Padilla.2020}:
\begin{equation}
	\text{recall} = 
		\frac{\text{\# \glspl{TP}}}{\text{\# \glspl{TP} + \# \glspl{FN}}} = 
		\frac{\text{\# \glspl{TP}}}{\text{\# ground truths}},
\end{equation}
i.e. the chance of the detector to detect all ground truths.

When using an \gls{R-CNN}, every detection comes with a score, which quantifies the confidence of the \gls{R-CNN} in the detection. By setting a threshold for this score, the number of predictions can be effectively controlled. A higher threshold yields fewer detections, which are more likely to be correct and therefore results in a higher precision. Contrarily, a lower threshold yields more detections, which are less likely to be correct and therefore results in a higher recall.

\begin{figure}
	\centering
	\includegraphics[width=\figurewidth]{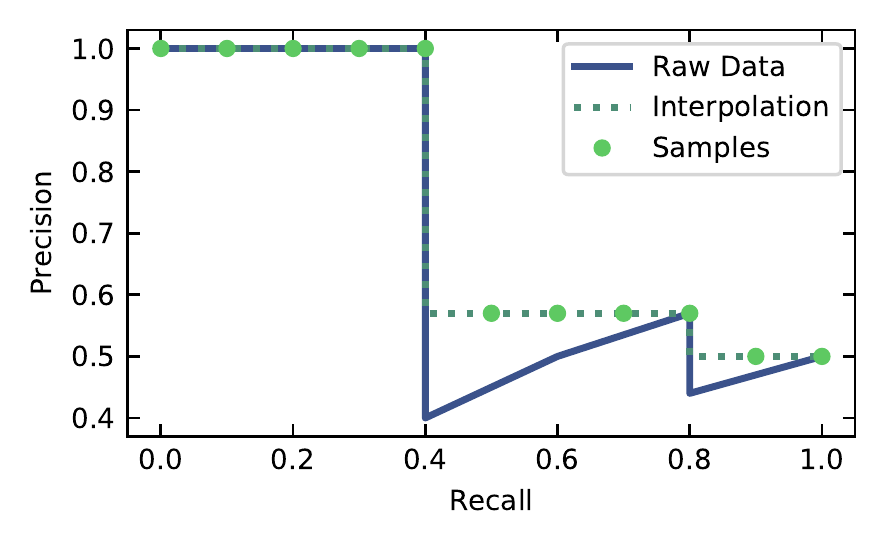}
	\caption{Illustration of a precision–recall curve.}
	\label{fig:PrecisionRecallCurve}
\end{figure}

This tradeoff between recall and precision can be represented by a precision–recall curve (see \cref{fig:PrecisionRecallCurve}). Due to its saw-tooth shape, precision–recall curves are often interpolated, by assigning each recall value the maximum precision value that can be found to the right side of it. After the interpolation, the \gls{AP} is determined by sampling a fixed number\footnote{For the studies at hand this number is \num{101}, in agreement with the \gls{COCO} evaluation scheme \cite{COCOConsortium.2019}.} of precision values at uniformly and linearly spaced recall intervals (see \cref{fig:PrecisionRecallCurve}) and calculating their average. 

Ultimately, the \gls{mAP} is the mean of multiple \glspl{AP} that result from the use of different \gls{IoU} thresholds, usually in the range from \SIrange{50}{90}{\percent} with increments of \SI{5}{\percent} \cite{COCOConsortium.2019}.

\subsubsection{Baseline}
\label{sec:Results-Baseline}
\graphicspath{{./figs/results/architecture-design/}}
As starting point for the architecture design, it was necessary to establish baselines, against which all subsequent experiments could be evaluated. For the evaluation at hand, two such baselines were established: 
\begin{enumerate}
	\item A default \mrcnn/ implementation without input augmentation (see \cref{sec:Method-InputAugmentation}).
	\item A basic \frcnn/ implementation without mask segmentation head (see \cref{sec:Method-KeyPointRCNN}), keypoint ordering (see \cref{sec:Method-KeypointOrdering} ), input augmentation (see \cref{sec:Method-InputAugmentation}) or error correction (see \cref{sec:Method-ErrorCorrection}).
\end{enumerate}

Wherever possible, common hyperparameters (e.g. batch size) of the two baseline models were set to identical values to maximize the comparability between both models. 

\begin{figure}
	\centering
	\includegraphics[width=\figurewidth]{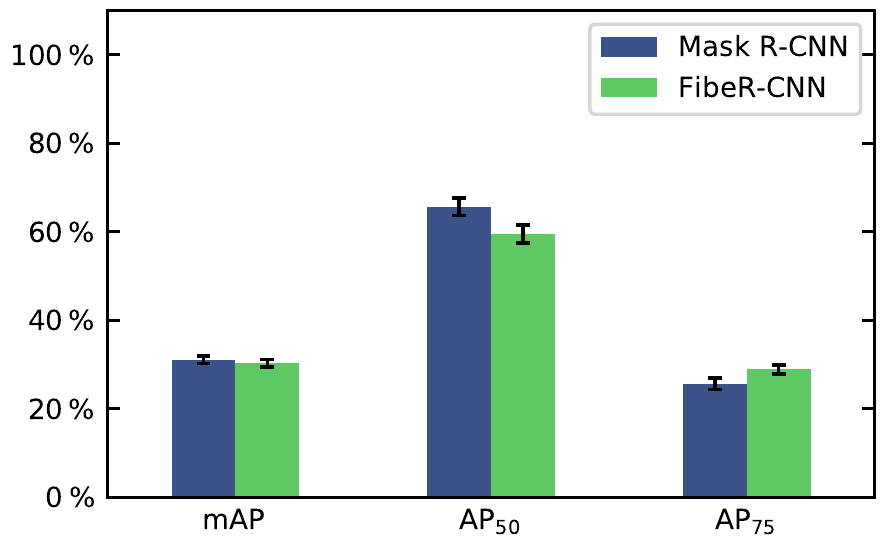}
	\caption{\Glsentryfull{AP} baselines for \mrcnn/ and \frcnn/ across an aggregation of all real test sets.}
	\label{fig:Results-Baseline}
\end{figure}

In \cref{fig:Results-Baseline}, the \glspl{AP} of the \mrcnn/ and the \frcnn/ baseline models are being compared. The performance of both models is quite similar with respect to \gls{symb:meanAveragePrecision}, with the \mrcnn/ model slightly outperforming the \frcnn/ model. Interestingly, \frcnn/ performs better than \mrcnn/ for higher \gls{IoU} thresholds, while \mrcnn/ performs better than \frcnn/ for lower \gls{IoU} thresholds.

\subsubsection{Mask Segmentation Head}
\label{sec:Results-MaskLoss}
As first addition, a mask segmentation head was added to the baseline \frcnn/ model. 

\begin{figure}
	\centering
	\includegraphics[width=\figurewidth]{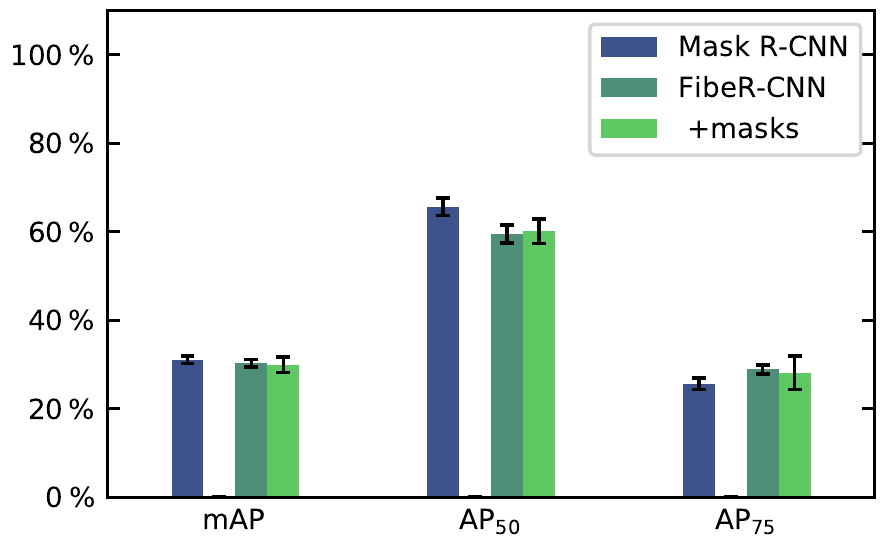}
	\caption{Influence of an additional mask segmentation head on the \glsentryfullpl{AP} of \frcnn/ across an aggregation of all real test sets. \mrcnn/ baseline as reference.}
	\label{fig:Results-MaskLoss}
\end{figure}

As can be seen in \cref{fig:Results-MaskLoss}, this does hardly affect its performance. However, the mask segmentation head was kept for the subsequent experiments, to allow the use of error detection and correction (see \cref{sec:Method-ErrorDetectionAndCorrection,sec:Results-ErrorCorrection}).

\subsubsection{Keypoint Ordering}
\label{sec:Results-KeypointOrdering}
As next addition, keypoint ordering according to the \enquote{top to bottom, left to right} rule, stated in \cref{sec:Method-KeypointOrdering} was implemented and evaluated. 

\begin{figure}
	\centering
	\includegraphics[width=\figurewidth]{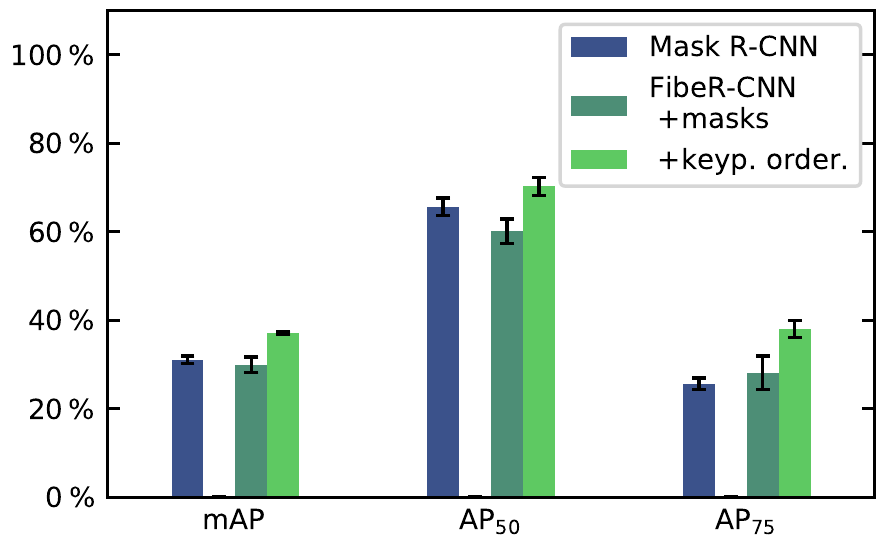}
	\caption{Influence of keypoint ordering on the \glsentryfullpl{AP} of \frcnn/ (with all previous additions) across an aggregation of all real test sets. \mrcnn/ baseline as reference.}
	\label{fig:Results-KeypointOrdering}
\end{figure}

\cref{fig:Results-KeypointOrdering} depicts a comparison of the \glspl{AP}, resulting from the evaluation. Keypoint ordering significantly improves the performance of \frcnn/, so that for the first time, it is able to surpass the \mrcnn/ baseline model's \glspl{AP} for all tested \gls{IoU} thresholds. 

The reason for this effect is presumably, that the start and end points of fibers are too similar to be distinguished, based on the intensity gradients in the input images alone. This leads to situations, where \frcnn/ chooses an end point of a fiber as both start \emph{and} end point, which leads to large errors with respect to the \gls{AP}, as explained in \cref{sec:Method-ErrorDetectionAndCorrection} and illustrated in \cref{fig:MisplacedKeypoint}. Therefore, it is essential to ensure a spatially consistent keypoint order, so that \frcnn/ can make use of additional spatial information in the image.

\subsubsection{Input Augmentation}
\label{sec:Results-InputAugmentation}
\begin{figure}
	\centering
	\includegraphics[width=\figurewidth]{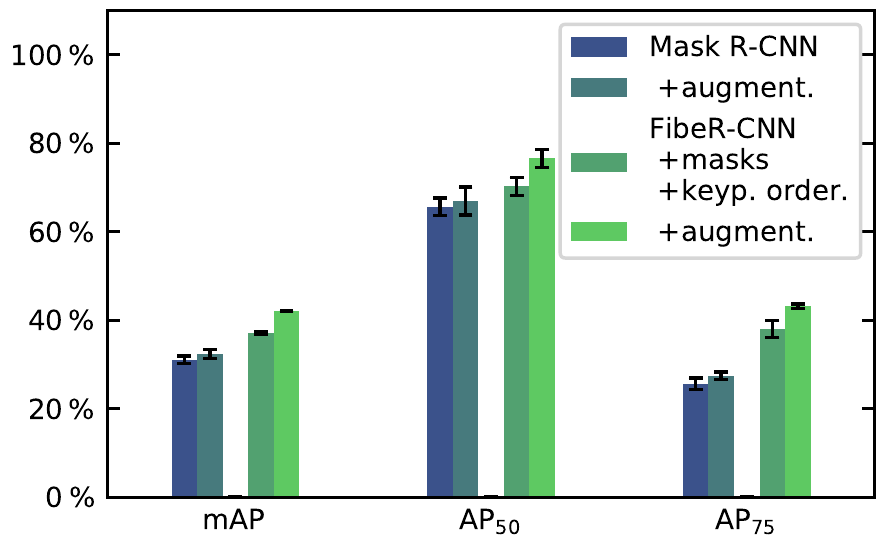}
	\caption{Influence of input augmentation on the \glsentryfullpl{AP} of \frcnn/ (with all previous additions) across an aggregation of all real test sets. \mrcnn/ baseline and \mrcnn/ with additional input augmentation as references.}
	\label{fig:Results-InputAugmentation}
\end{figure}

As third extension, input augmentation was added to the \frcnn/ architecture. Since \mrcnn/ can usually profit from input augmentation, it was also added to the baseline \mrcnn/ model (see \cref{sec:Results-Baseline}), to maintain fairness with respect to the model comparison (see \cref{fig:Results-InputAugmentation}).

Both \mrcnn/ and \frcnn/ profit from the use of input augmentation. However, for \frcnn/ the effect is more distinct, thereby making its \glspl{AP} surpass those of \mrcnn/ even further than with the previous model version.

\subsubsection{Error Correction}
\label{sec:Results-ErrorCorrection}
\begin{figure}
	\centering
	\includegraphics[width=\figurewidth]{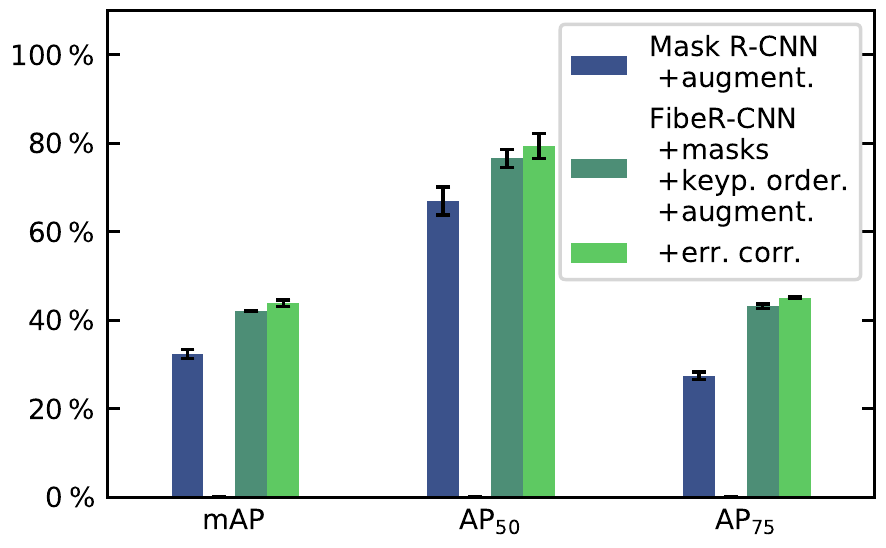}
	\caption{Influence of error correction on the \glsentryfullpl{AP} of \frcnn/ (with all previous additions) across an aggregation of all real test sets. \mrcnn/ with additional input augmentation as reference.}
	\label{fig:Results-ErrorCorrection}
\end{figure}

As fourth and final extension to the \frcnn/ architecture, error correction was tested (see \cref{sec:Method-ErrorCorrection}). As expected, error correction yields increased \glspl{AP} (see \cref{fig:Results-ErrorCorrection}). However, the improvement is rather small and comes at a high computational cost (see \cref{sec:Method-ErrorCorrection}). 
The reason for the minor effect of the error correction is that the predictions quality of the \frcnn/ architecture without error correction is of rather binary nature: It is either excellent or bad, but much less often mediocre. Consequently, excellent results are hardly improved by the error correction, because there are only few errors. Bad results, however, are not improved enough to advance into the territory of $\gls{symb:intersectionoverUnion} \geq \SI{50}{\percent}$ and are therefore -- even with error correction -- still not factored into the surveyed quality metrics.

\subsubsection{Summary}
\label{sec:Results-Summary}
\begin{figure*}
	\centering
	\includegraphics[width=2\figurewidth]{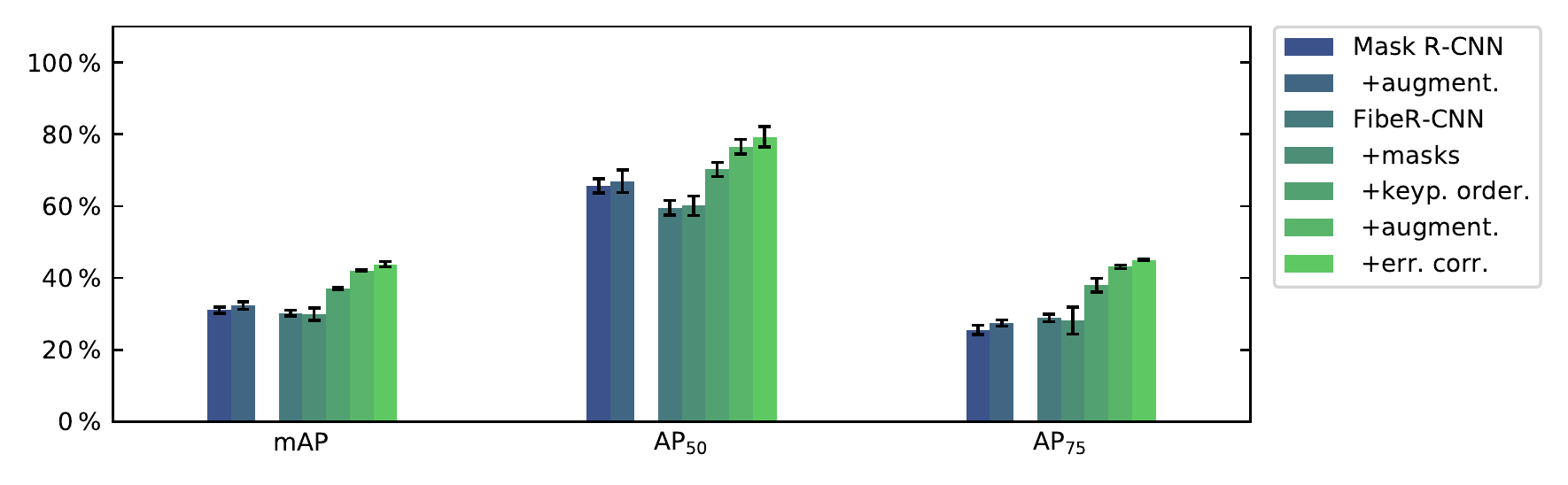}
	\caption{\Glsentryfullpl{AP} across an aggregation of all real test sets achieved by each of the tested architectures.}
	\label{fig:Results-Summary}
\end{figure*}

\cref{fig:Results-Summary} summarizes the \glspl{AP} of all tested \mrcnn/ and \frcnn/ model variants. By systematically improving the \frcnn/ architecture, its \gls{symb:meanAveragePrecision}, \gls{symb:averagePrecision50} and \gls{symb:averagePrecision75} were increased by \SI{14}{\pp}\footnote{percentage points} (\SI{95}{\percent}\,$\text{\glsentryshort{CI}} = \SIrange[range-phrase ={,}\,]{12}{15}{\pp}$), \CIpp{20}{15}{25} and \CIpp{16}{15}{17} respectively, compared to the \frcnn/ baseline model. With these improvements, its \gls{symb:meanAveragePrecision}, \gls{symb:averagePrecision50} and \gls{symb:averagePrecision75} surpass those of the best tested \mrcnn/ model by \CI{11}{10}{13}, \CI{12}{6}{18} and \CI{18}{17}{19} respectively. 

The final version of \frcnn/, which was used for all subsequent experiments, included all presented extensions, i.e. a mask segmentation head, keypoint ordering, input augmentation and error correction.

\subsection{Training Data Supplementation}
\label{sec:Results-TrainingDataSupplementation}
\graphicspath{{./figs/results/training-data-supplementation/}}
For many applications, the accuracy of \glspl{CNN} scales excellently with an increasing training data set size. Therefore, it was examined, whether the supplementation of the training data with synthetic images (\texttt{[?l|?c|?o]}$_\text{synth.}$; see \cref{sec:Data-ImageSynthesis} and \cref{tab:DataSetComparisonCombined}) was beneficial for the \glspl{AP} achieved by \frcnn/.

\begin{figure}
	\centering
	\includegraphics[width=\figurewidth]{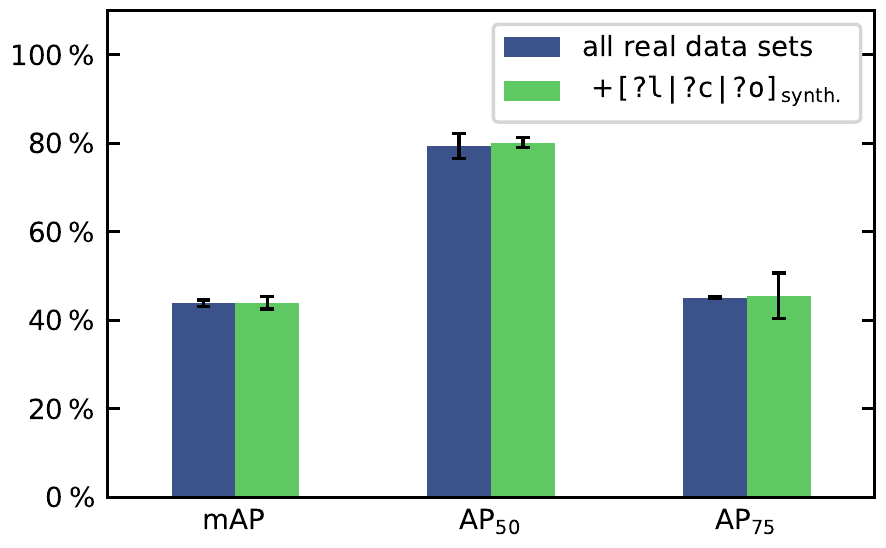}
	\caption{Influence of training data supplementation on the \glsentryfullpl{AP} of \frcnn/ across an aggregation of all real test sets. \frcnn/ trained on all available real training data sets as reference.}
	\label{fig:Results-TrainingDataSupplementation}
\end{figure}

\cref{fig:Results-TrainingDataSupplementation} compares the \glspl{AP} of \frcnn/ with and without training data supplementation. Within the margin of error, training data supplementation does not have any influence on the \glspl{AP} of \frcnn/.

\subsection{Lazy Annotation}
Apart from trying to improve the performance of \frcnn/ by using synthetic images, just like semiautomatic annotation (see \cref{sec:Data-SemiautomaticAnnotation}), they may also be used to circumvent the need for a manual annotation. Naturally, it is to be expected that the resulting \glspl{AP} will be lower than those achieved with the complete set of real training data used before. Nevertheless, it is interesting to examine what \glspl{AP} can be achieved with a \enquote{lazy} annotation.

\begin{figure}
	\centering
	\includegraphics[width=\figurewidth]{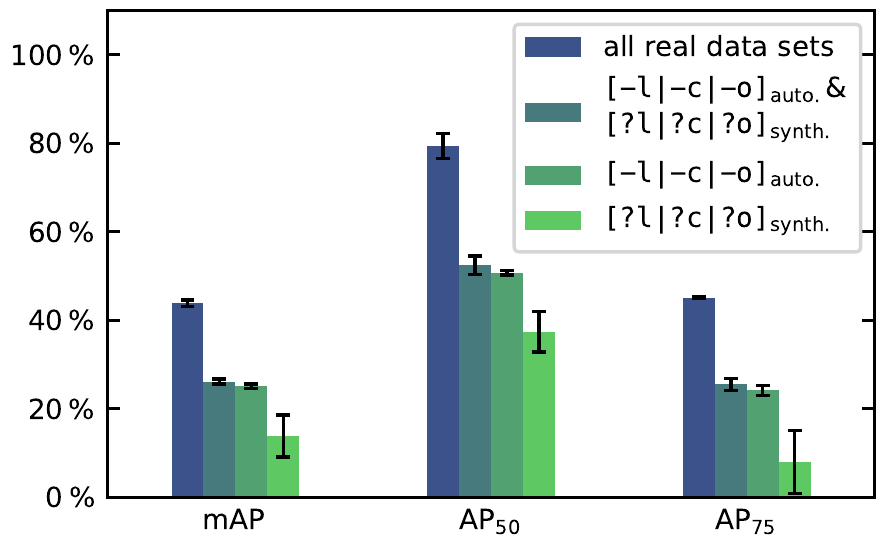}
	\caption{\Glsentryfullpl{AP} of \frcnn/s, trained on different data sets, across an aggregation of all real test sets. \frcnn/ trained on all available real training data sets as reference.}
	\label{fig:Results-LazyAnnotation}
\end{figure}

\cref{fig:Results-LazyAnnotation} compares the \glspl{AP} of \frcnn/, trained with only synthetic data (\texttt{[?l|?c|?o]}$_\text{synth.}$), only semiautomatically annotated data (\texttt{[-l|-c|-o]}$_\text{auto.}$) and a combination of both. 

The \frcnn/ model, trained only on synthetic data, yields a poor performance in comparison to the other tested models. This supports the supposition that the utilized synthetic images lack the necessary realism. 

While better than only synthetic data, the use of only semiautomatic data still falls short of the use of the complete set of real training data, available for this publication. This observation is indeed plausible, because only fibers with simple shapes and neither loops, clutter nor overlaps can be annotated semiautomatically. Therefore, the set of semiautomatically annotated images does not cover the domain of the utilized test data sufficiently. 

This also explains, why the combined set of both \enquote{lazily} annotated data sets performs slightly better than its individual components: While the semiautomatically annotated set contributes the necessary realism, the synthetic data set contributes the necessary complexity. However, also the combined data set still lacks the quality of the complete set of real training data.

\subsection{Application}
\label{sec:Results-Application}
While the previous two sections concentrated on ways to improve the precision of \frcnn/ or to reduce the effort to manually produce annotations, this section will discuss what tasks \frcnn/ can be used for, how good it performs on these tasks and what possible inhibiting factors for a successful application to real world problems are.

\subsubsection{Detection Quality}
\label{sec:Results-DetectionQuality}
\begin{figure*}
	\setlength{\tabcolsep}{0mm} %
	\centering
	\begin{tabularx}{\textwidth}{XXXX}
		\image{_loops_-clutter_-overlaps_1} & \image{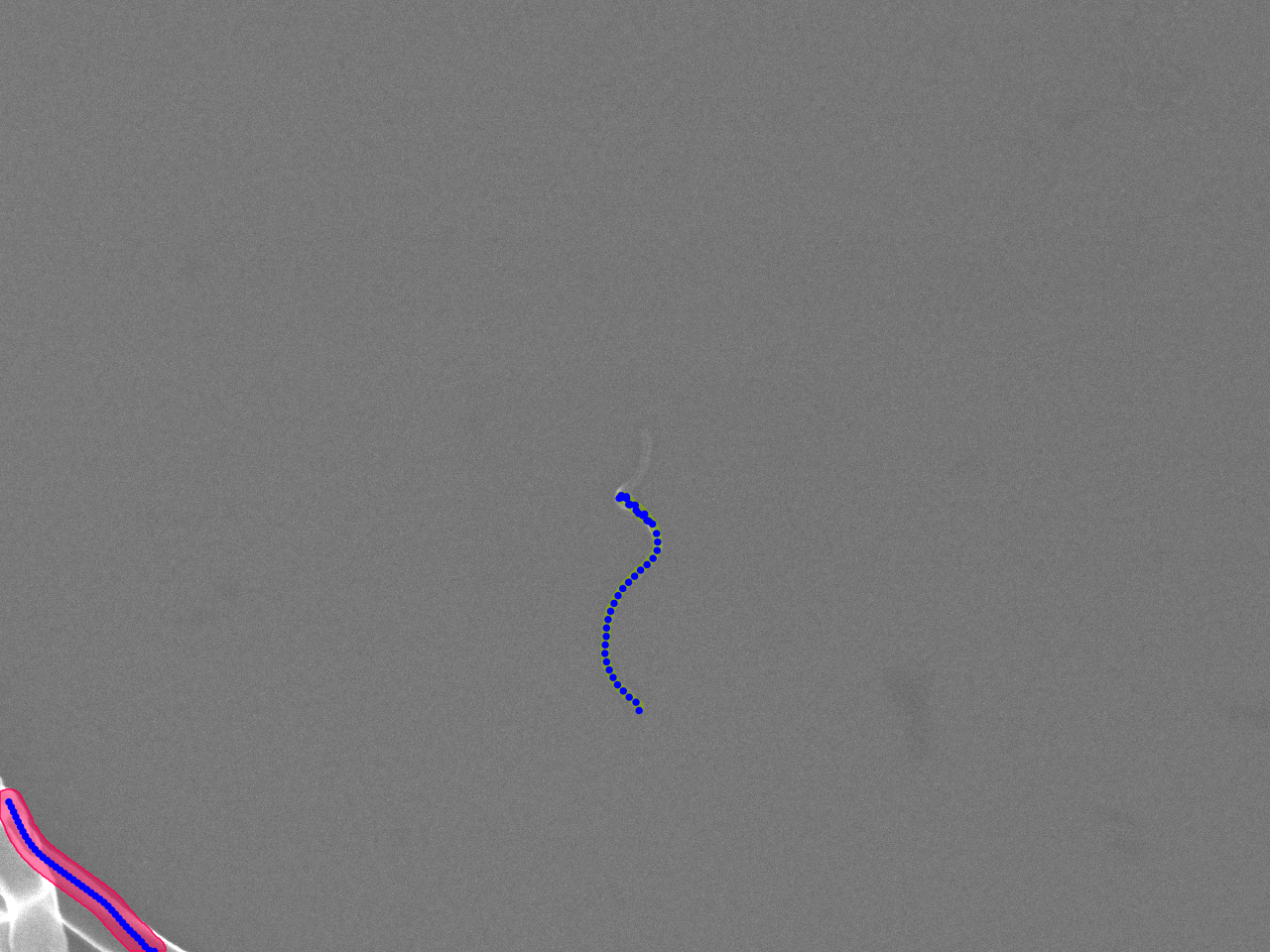} & \image{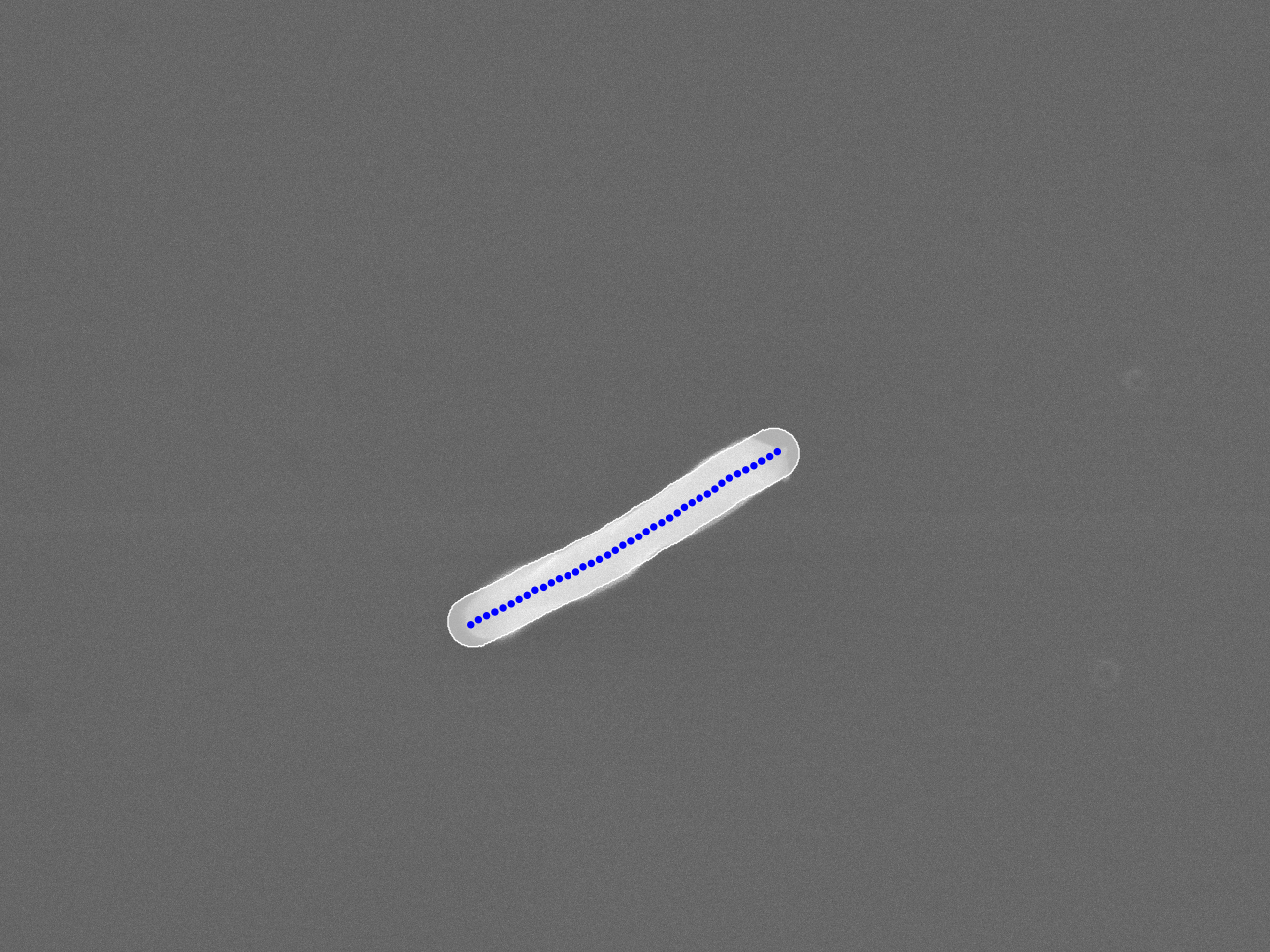} & \image{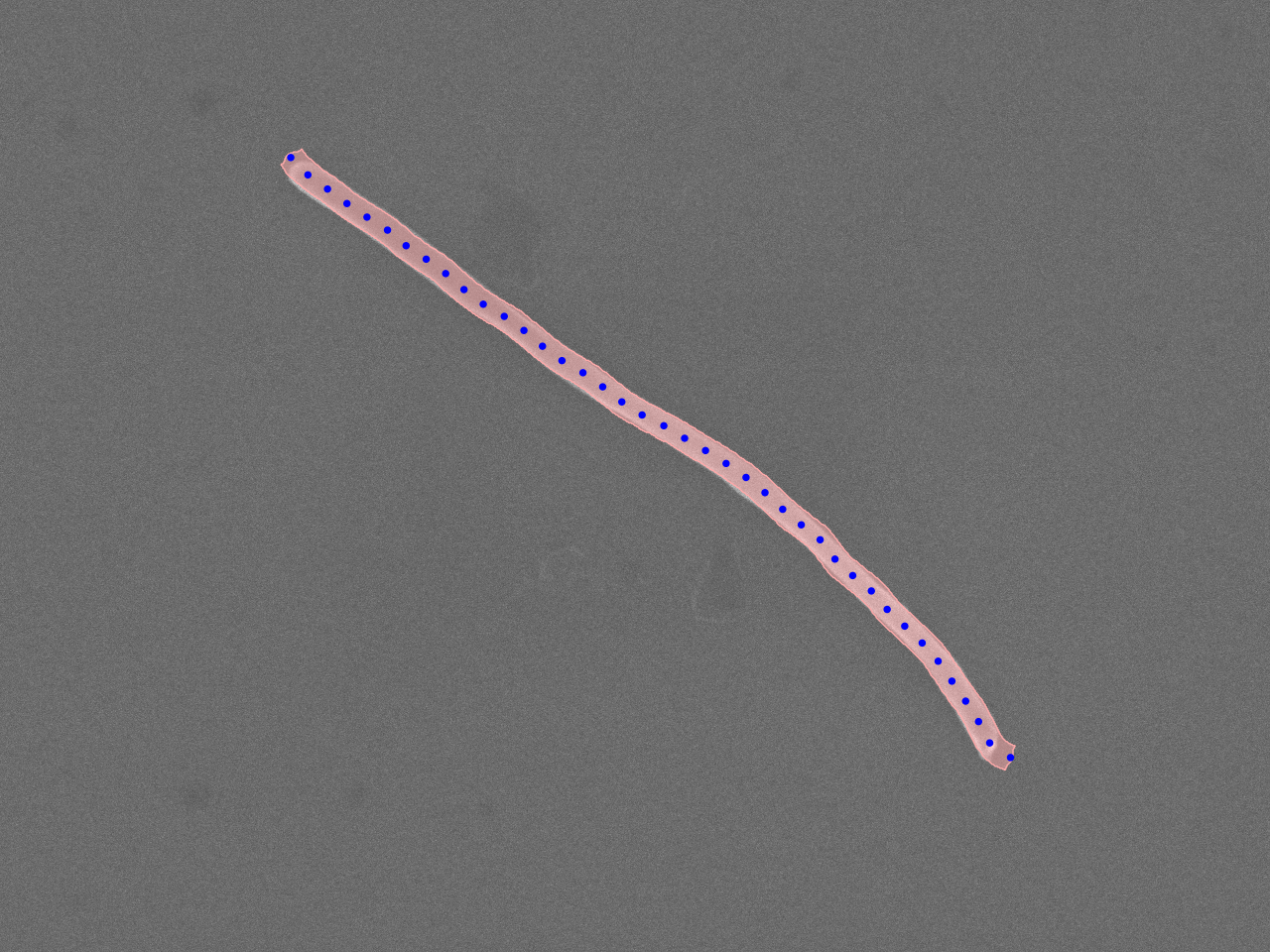}   \\
		                                                       \multicolumn{4}{c}{\texttt{[-l|-c|-o]}}                                                          \\
		\image{_loops_-clutter_+overlaps_1} & \image{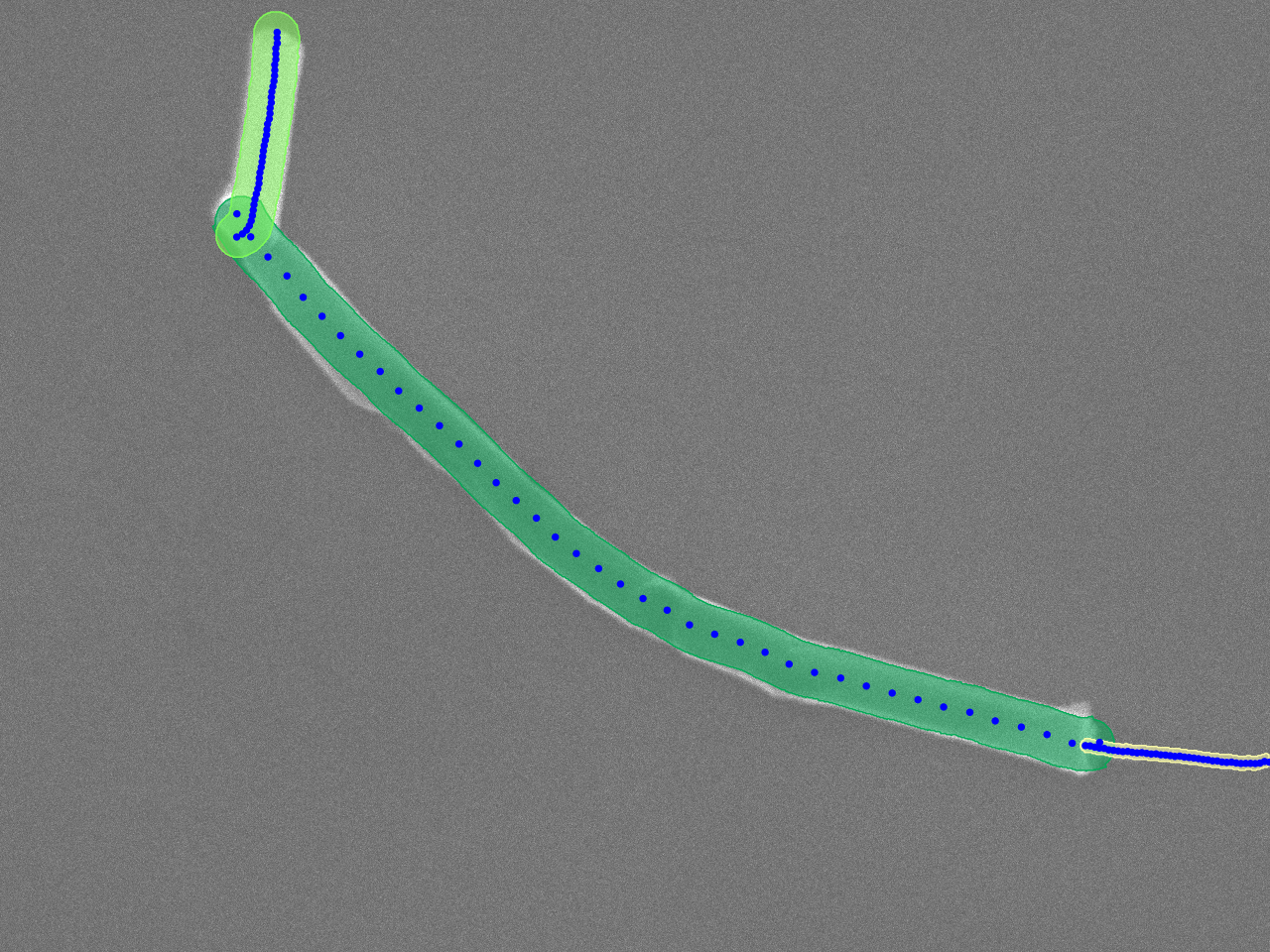} & \image{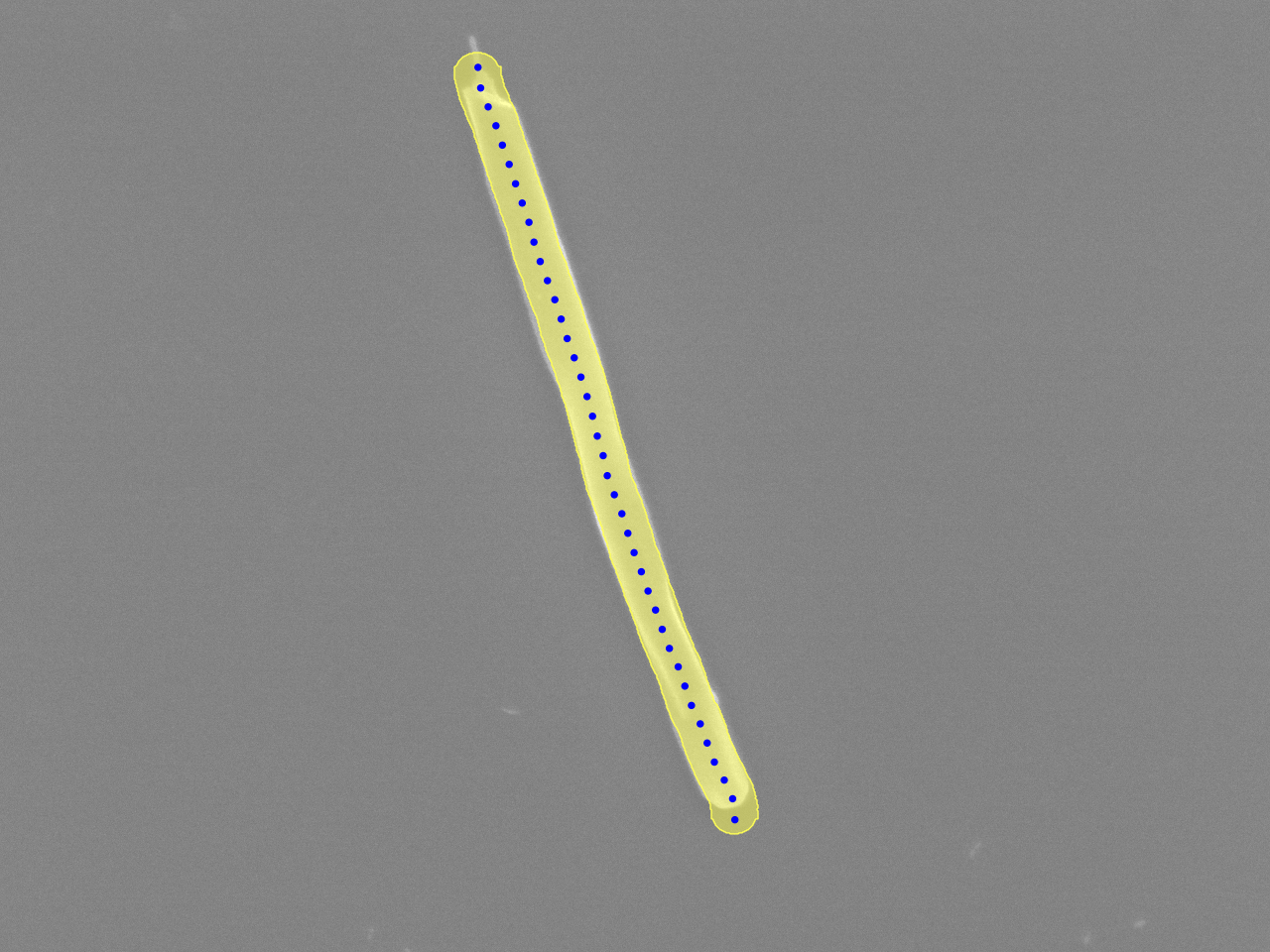} & \image{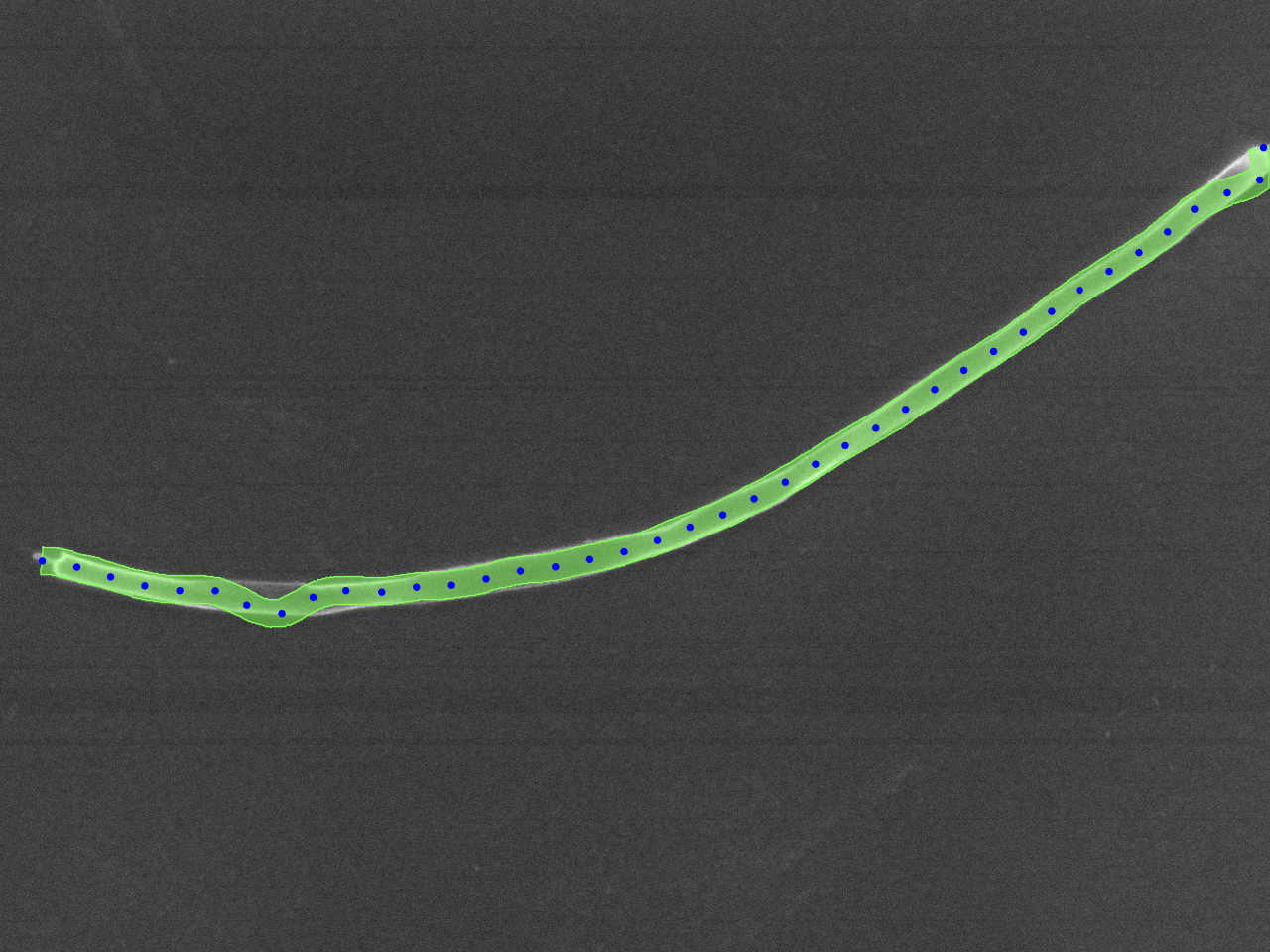}   \\
		                                                       \multicolumn{4}{c}{\texttt{[-l|-c|+o]}}                                                          \\
		\image{_loops_+clutter_-overlaps_1} & \image{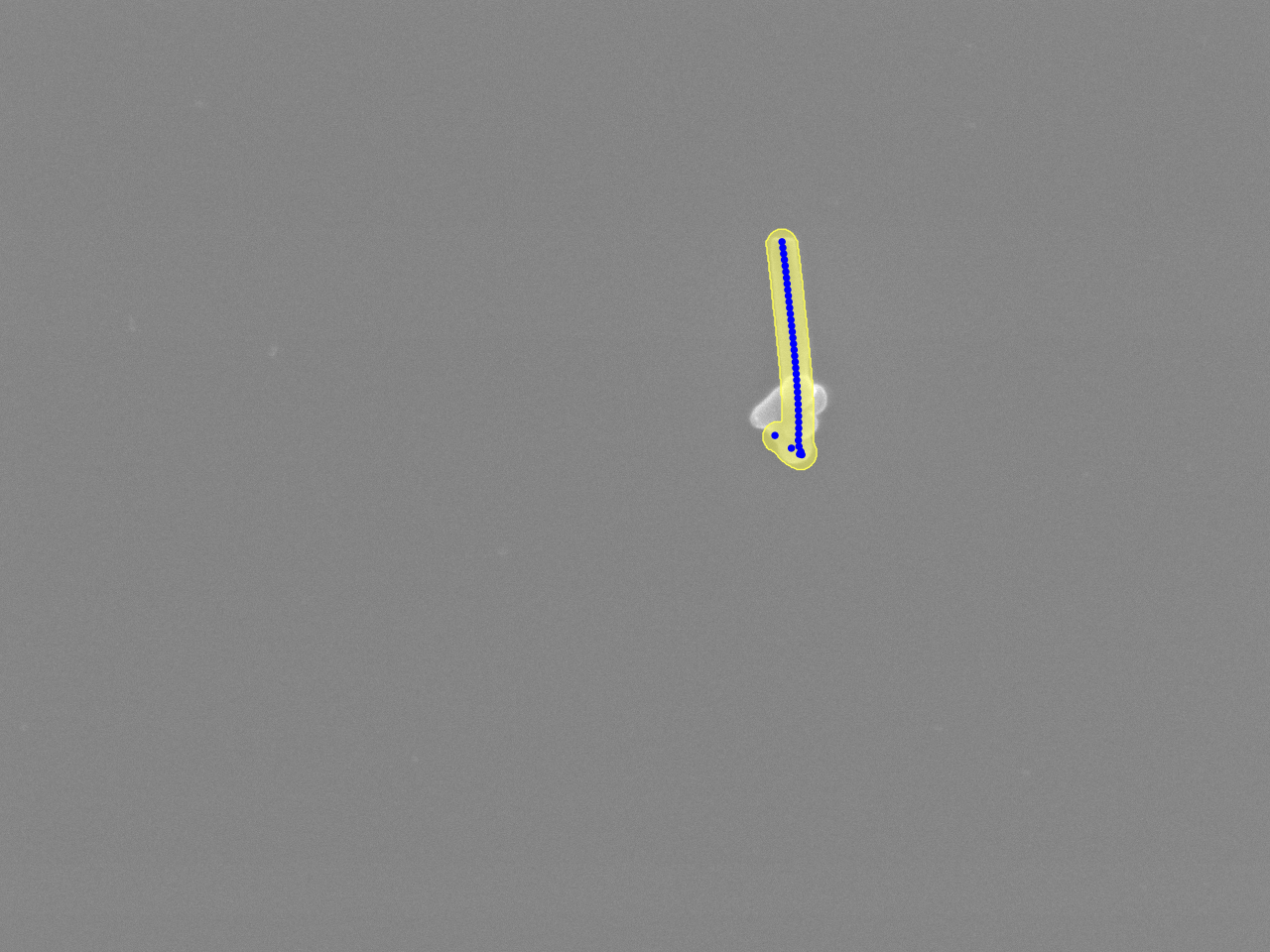} & \image{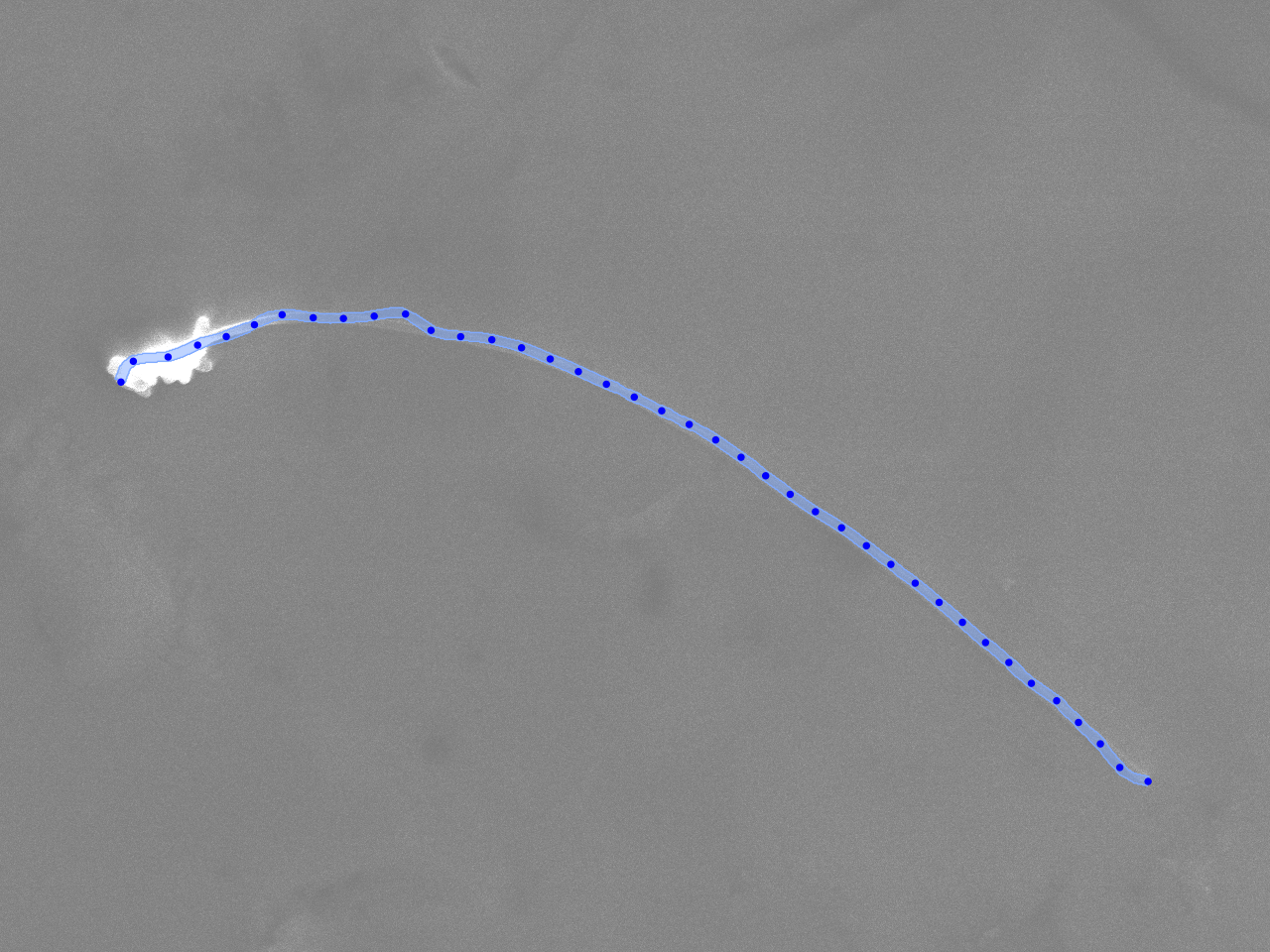} & \image{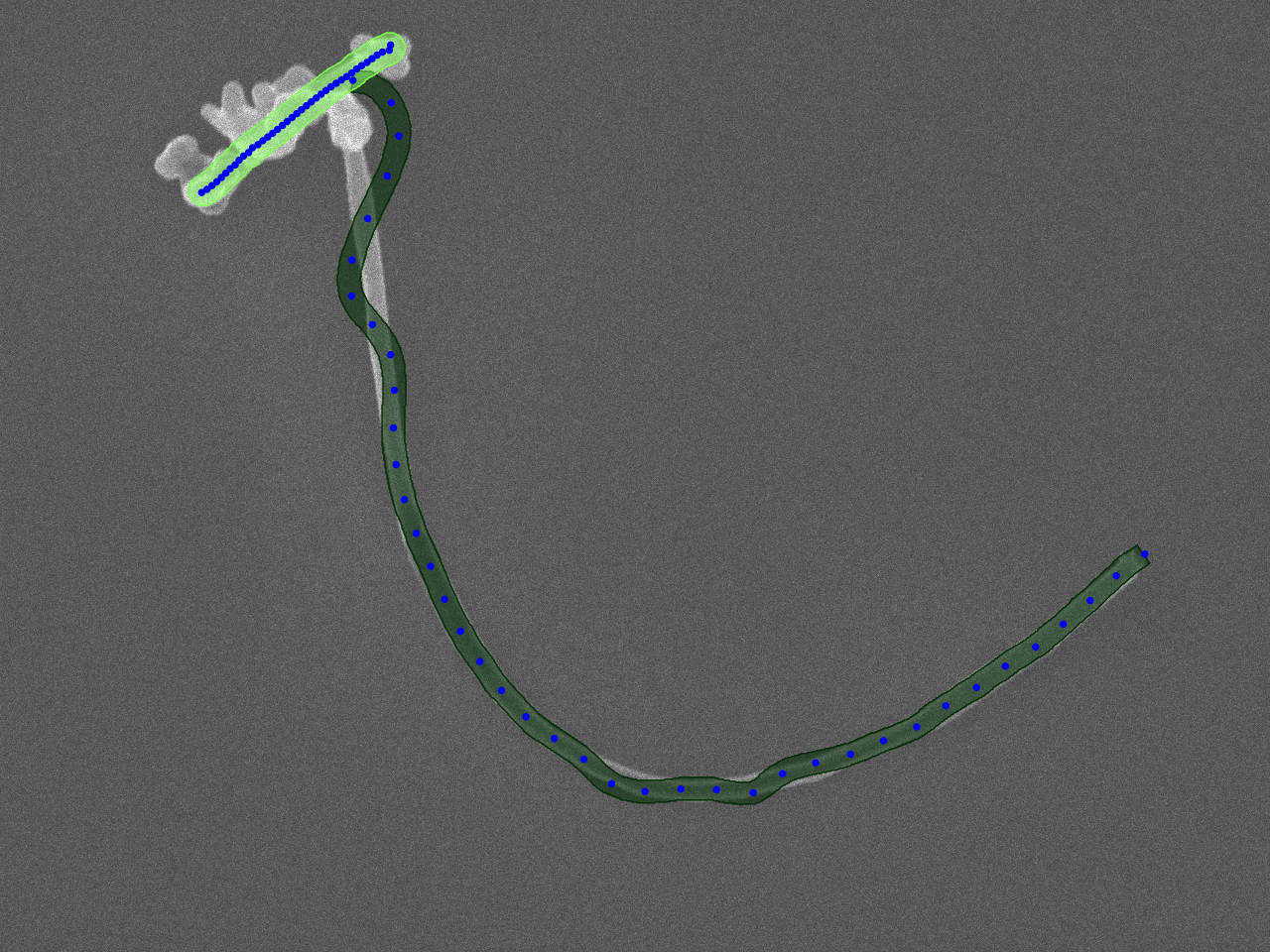}   \\
		                                                       \multicolumn{4}{c}{\texttt{[-l|+c|-o]}}                                                          \\
		\image{_loops_+clutter_+overlaps_1} & \image{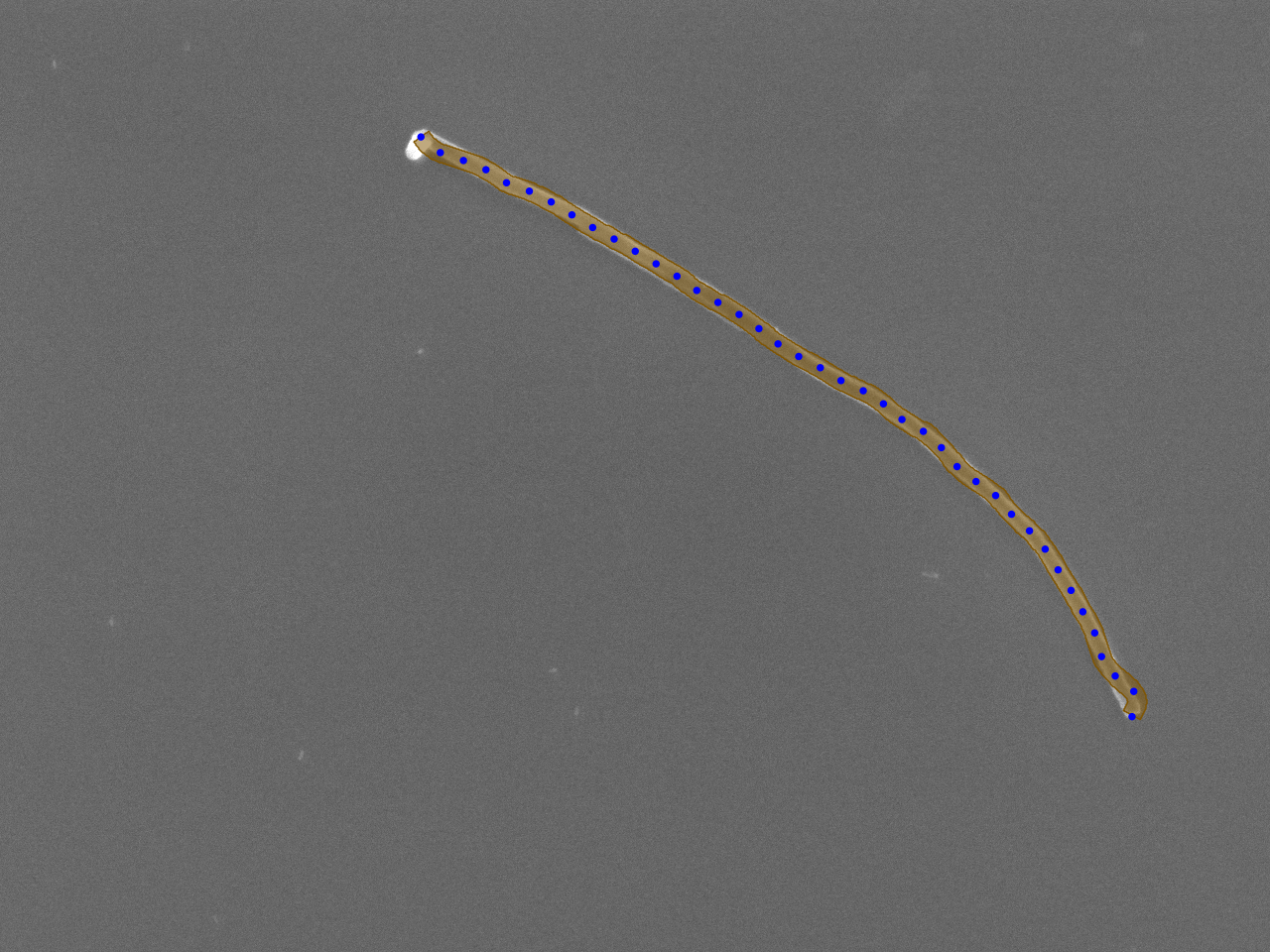} & \image{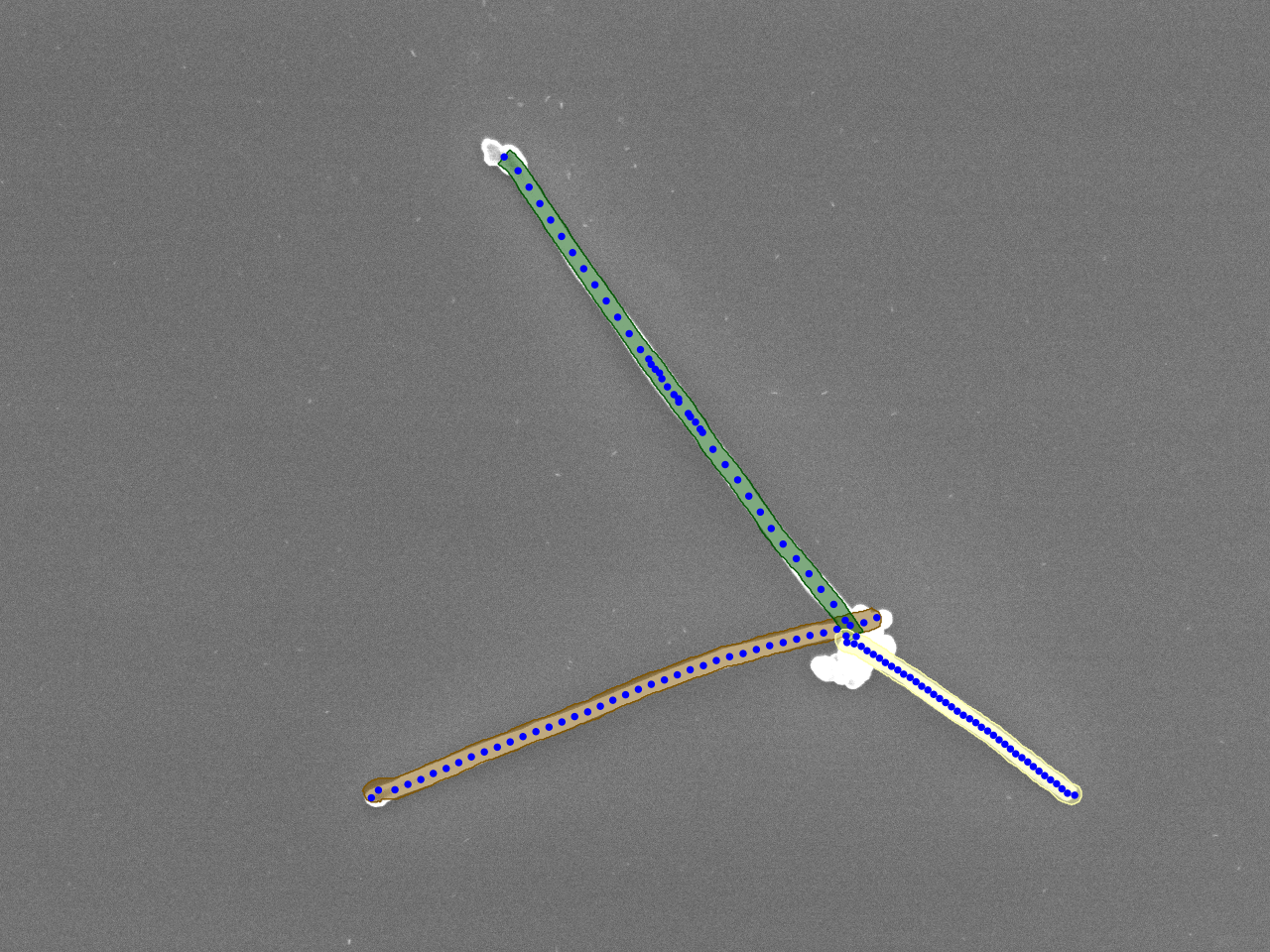} & \image{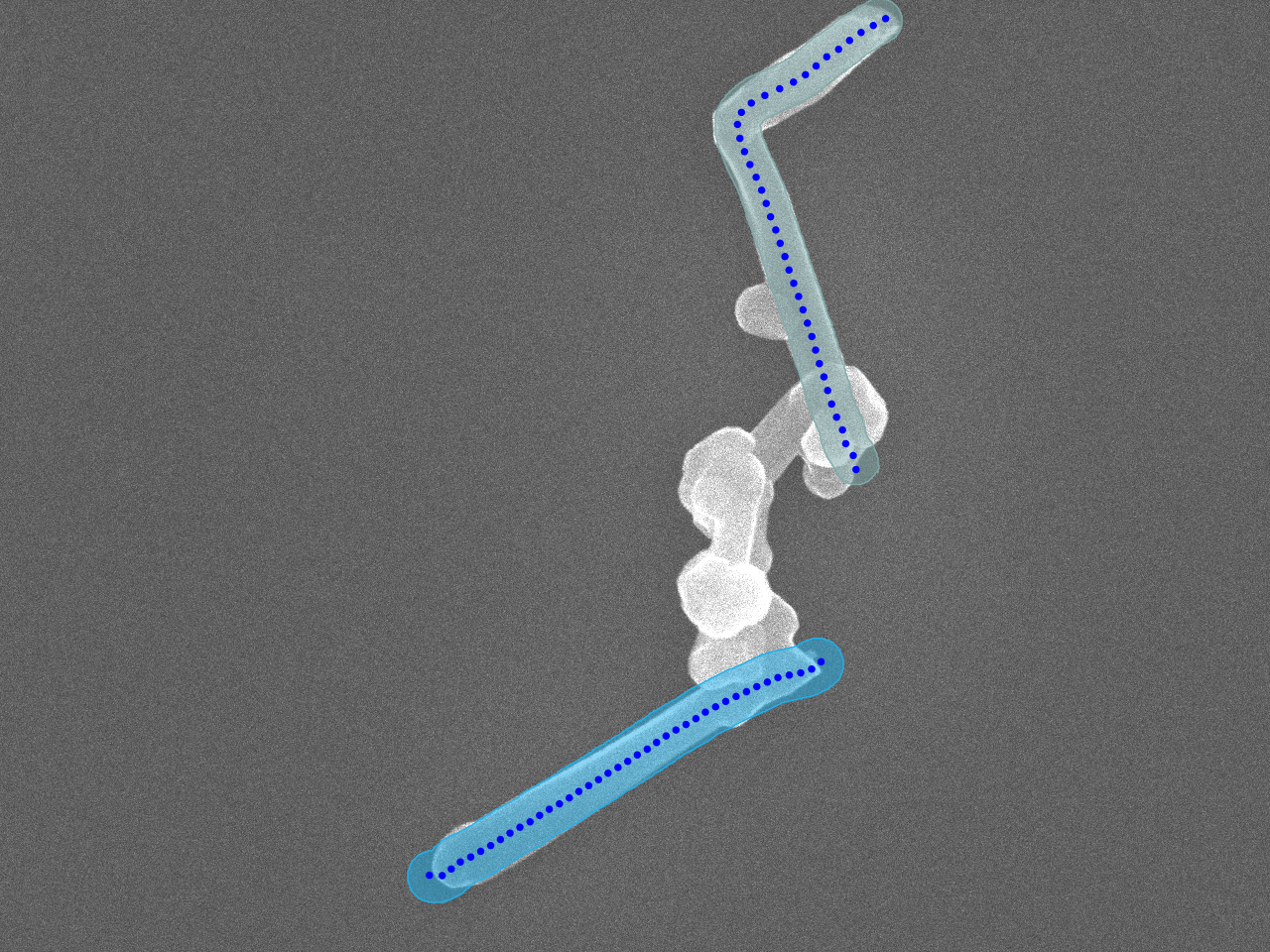}   \\
		                                                       \multicolumn{4}{c}{\texttt{[-l|+c|+o]}}                                                          \\
		\image{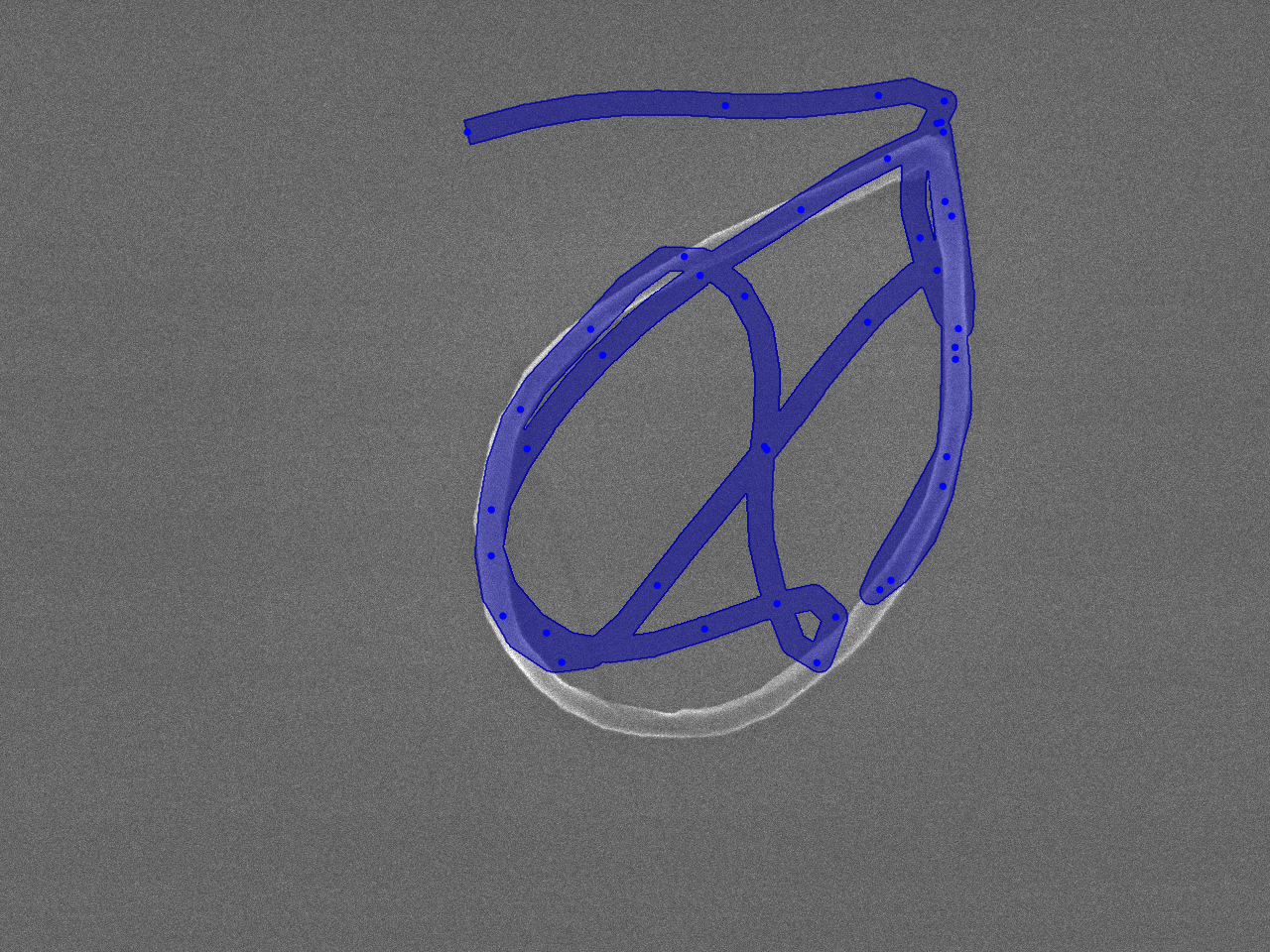} & \image{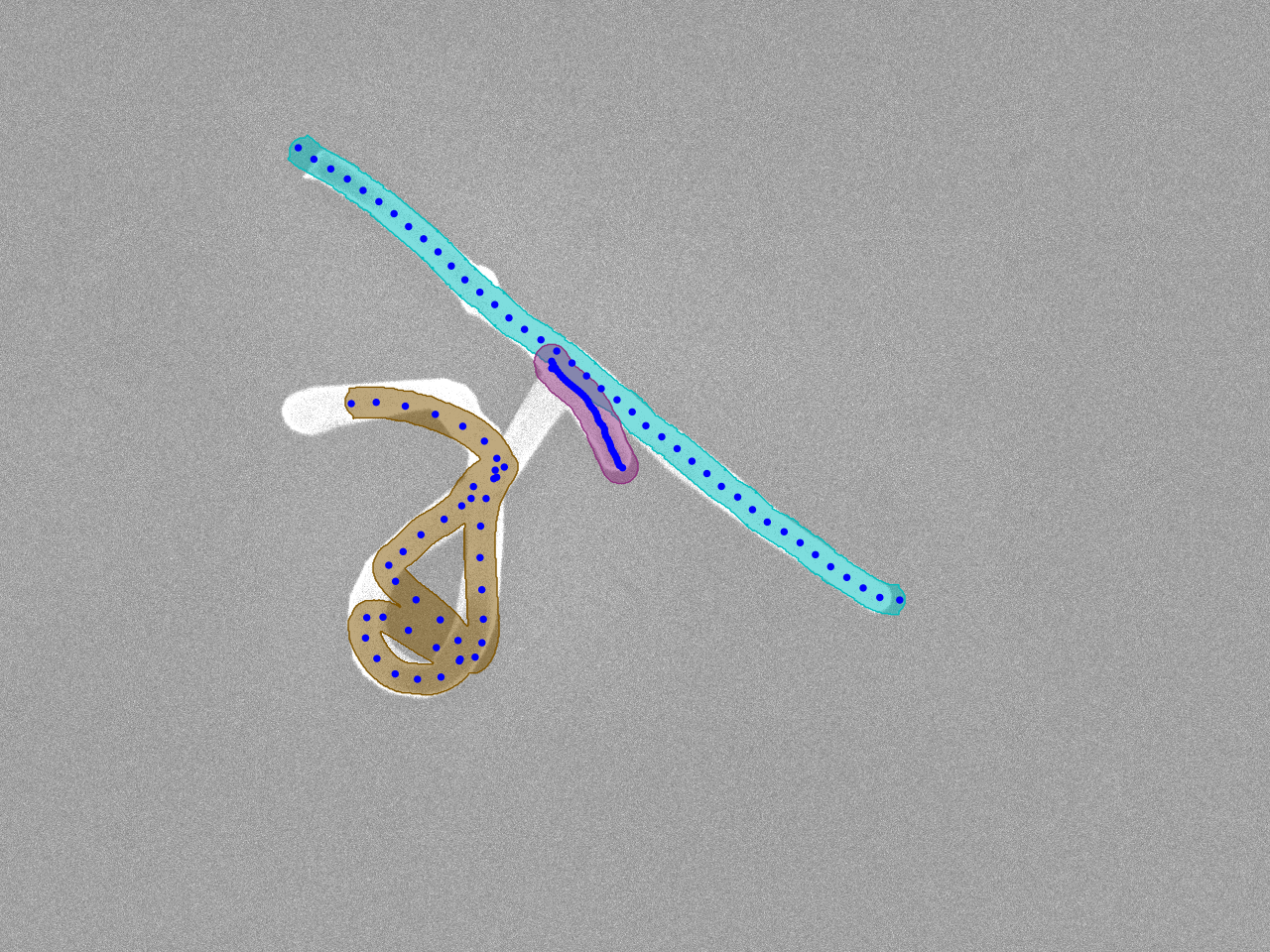} & \image{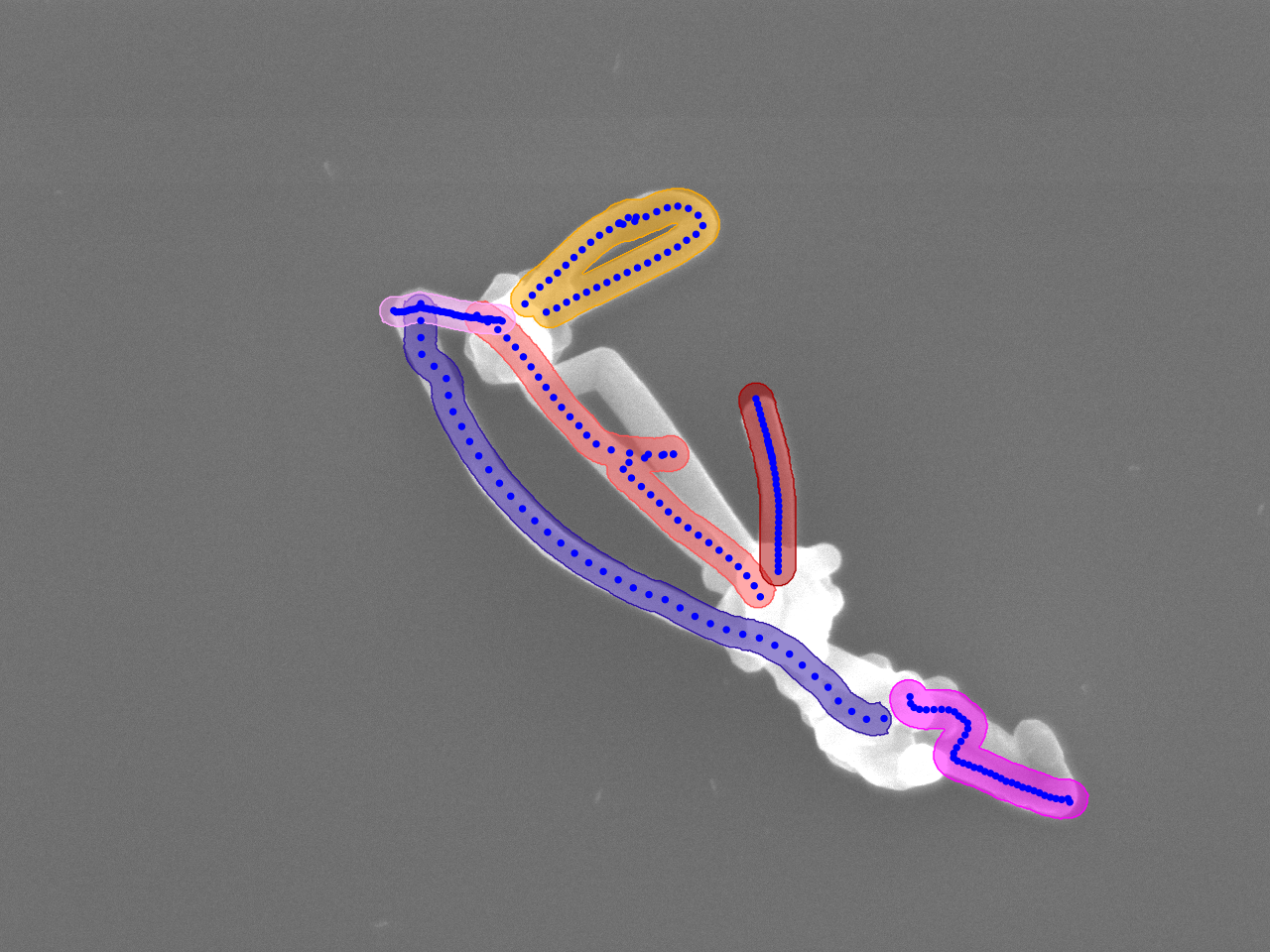} & \image{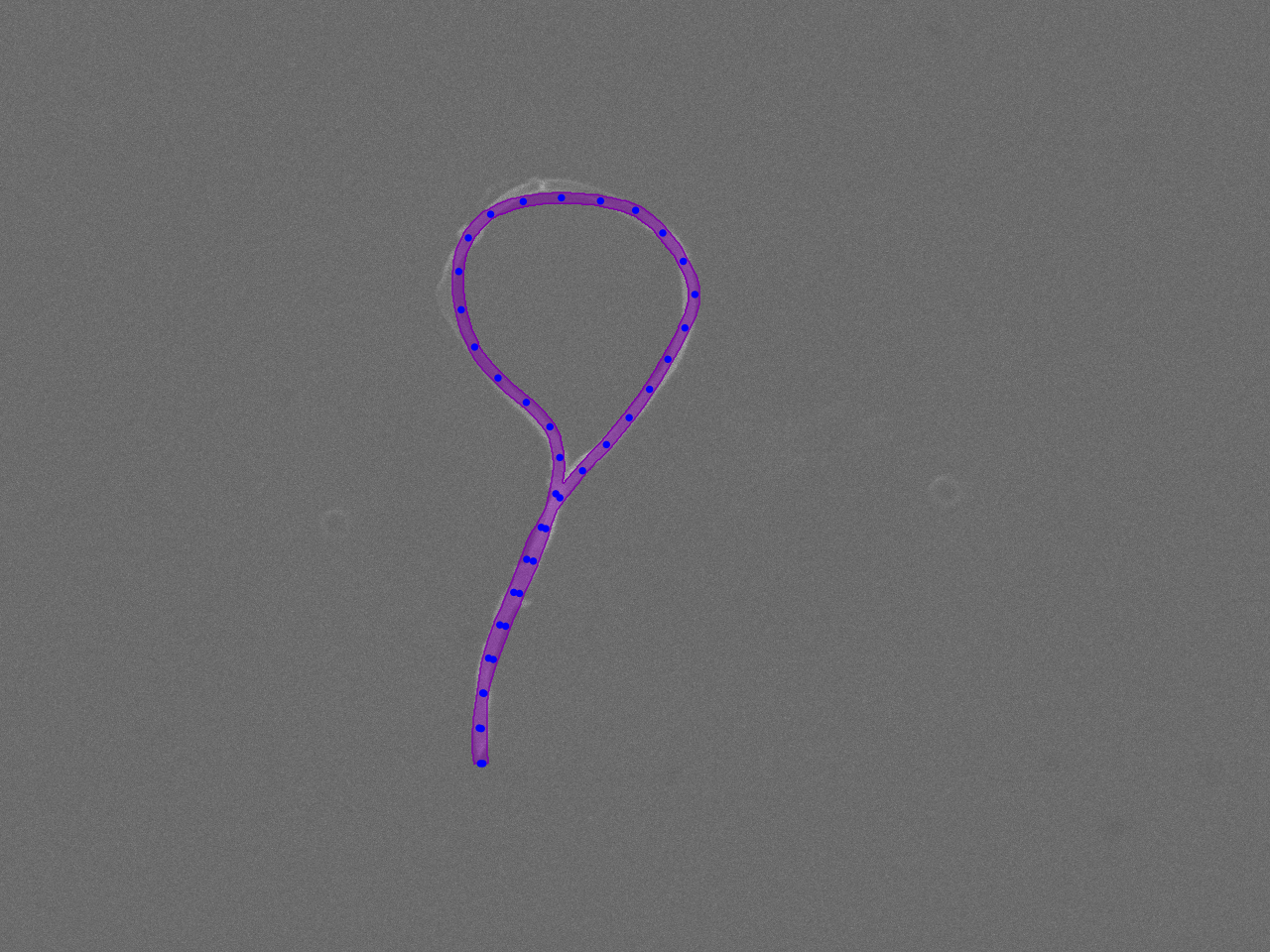}   \\
		                                                       \multicolumn{4}{c}{\texttt{[+l|?c|?o]}}                                                          \\
		\image{_loops_-clutter_-overlaps__automatic__1} & \image{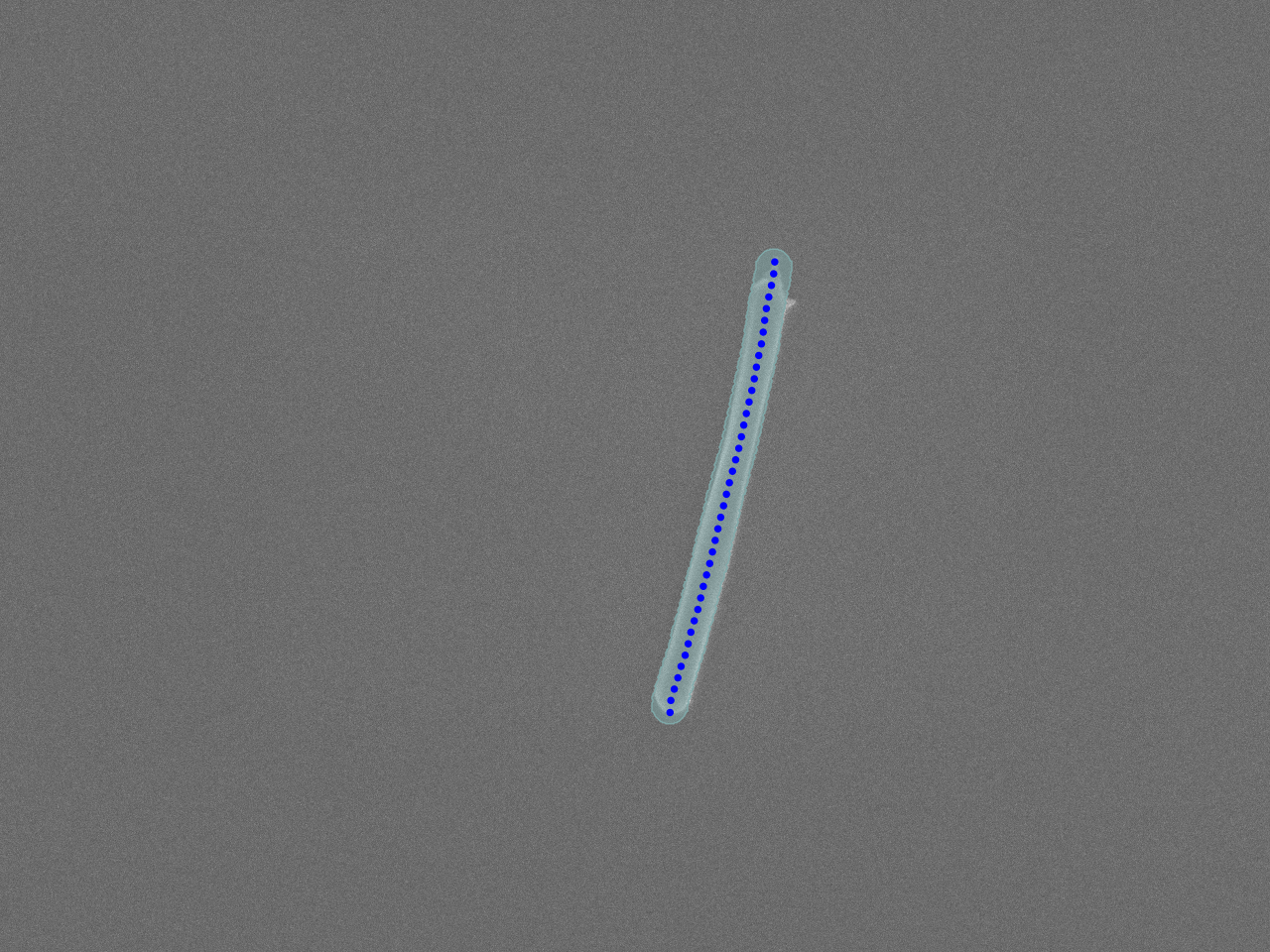} & \image{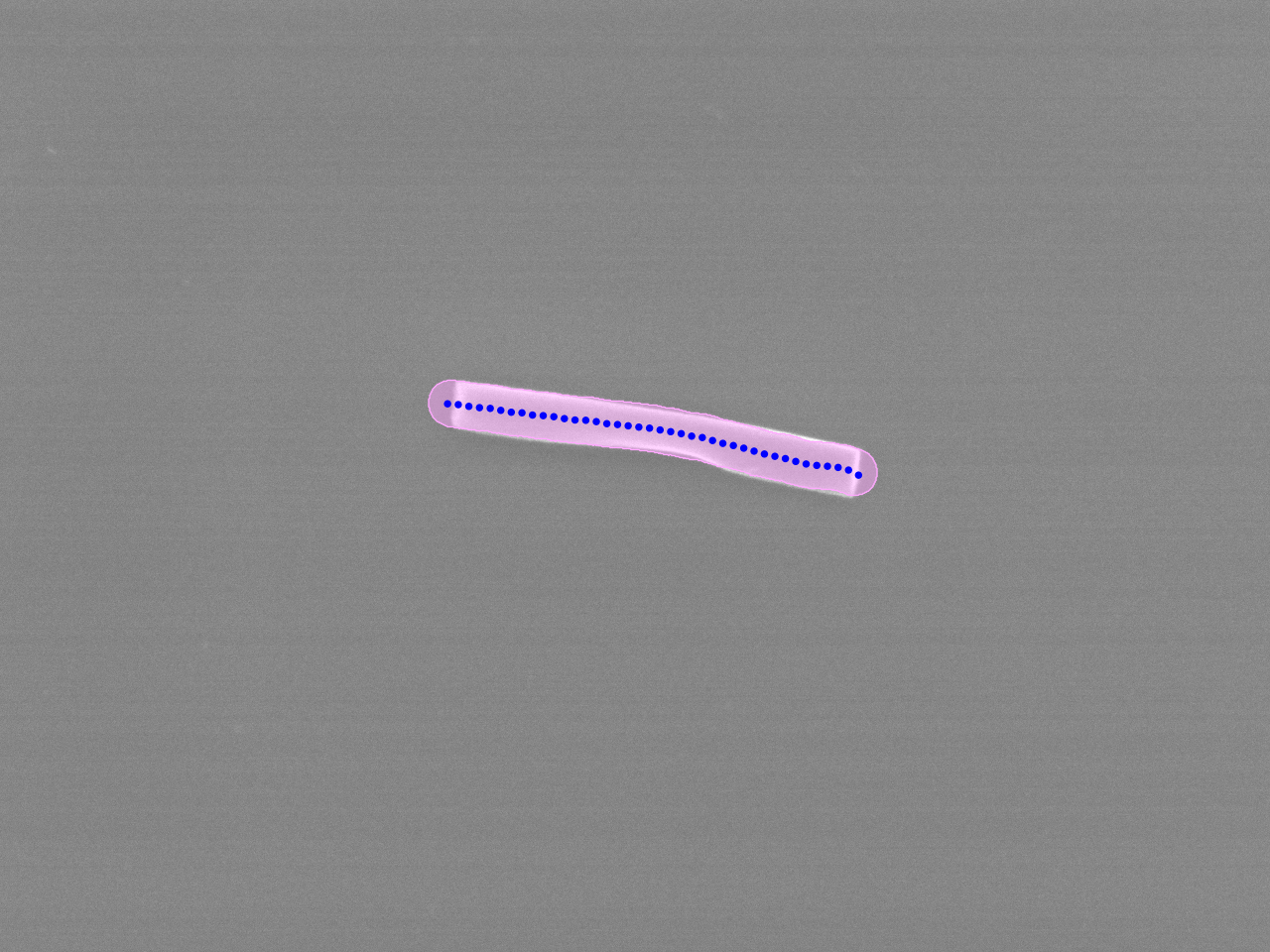} & \image{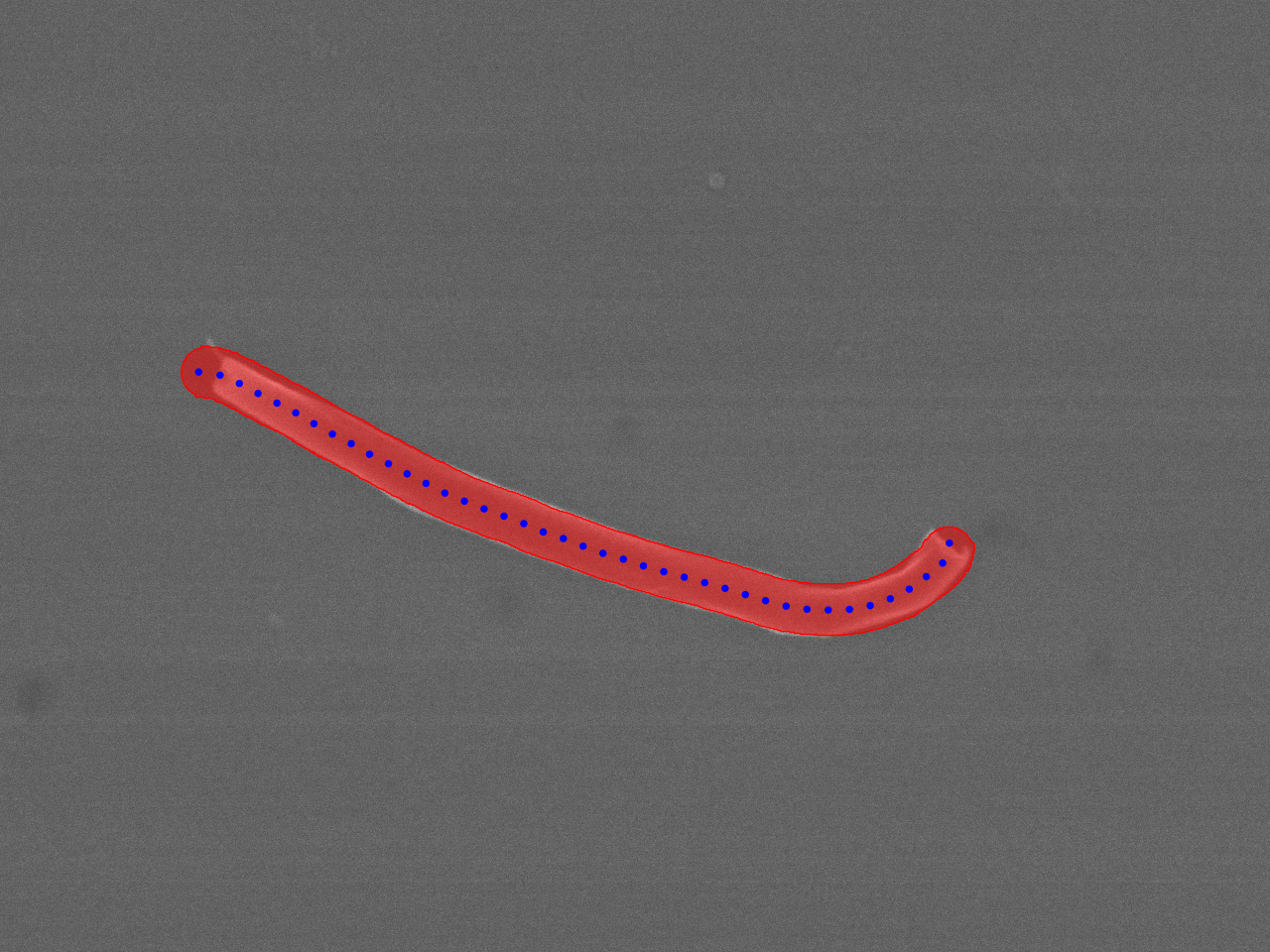}   \\
		                                               \multicolumn{4}{c}{\texttt{[-l|-c|-o]}$_\text{auto.}$}                                                 
	\end{tabularx}
	\caption{Example detections for each of the real test data sets (\identifierlegend/).}
	\label{fig:DetectionQuality}
\end{figure*}

To get a first, qualitative impression on the capabilities of \frcnn/, it is helpful to visually inspect a selection of example detections. \cref{fig:DetectionQuality} shows four randomly chosen detections for each of the six test data sets (see \cref{sec:Data-DataSets}). 
\nImageColumns4\relax 
\setlength{\spacerwidth}{2mm}
\setlength{\imagewidth}{(\textwidth-\spacerwidth*\nImageColumns)/\nImageColumns}
\setlength{\spacerheight}{1cm}
\graphicspath{{./figs/results/detection-quality/}}
\graphicspath{{./figs/results/}}

Fibers with neither loops, clutter nor overlaps (\texttt{[-l|-c|-o]} and \texttt{[-l|-c|-o]}$_\text{auto.}$) are not challenging for \frcnn/ and even fibers that could not be segmented semiautomatically, are detected reliably. Also, the presence of clutter (\texttt{[-l|+c|-o]}), overlaps (\texttt{[-l|-c|+o]}) or a combination of both (\texttt{[-l|+c|+o]}) does not impede the detection quality too much, as long as the degree of instance-instance or instance-clutter overlap is not too high and the size difference between overlapping fibers is not too small.

Contrarily, loops (\texttt{[+l|?c|?o]}) pose a greater challenge for \frcnn/, especially, when being combined with overlaps and clutter. Still, some loops can be detected flawlessly by \frcnn/. 

\subsubsection{Mean Average Precision}
\label{sec:Results-MeanAveragePrecision}
\begin{figure}
	\centering
	\includegraphics[width=\figurewidth]{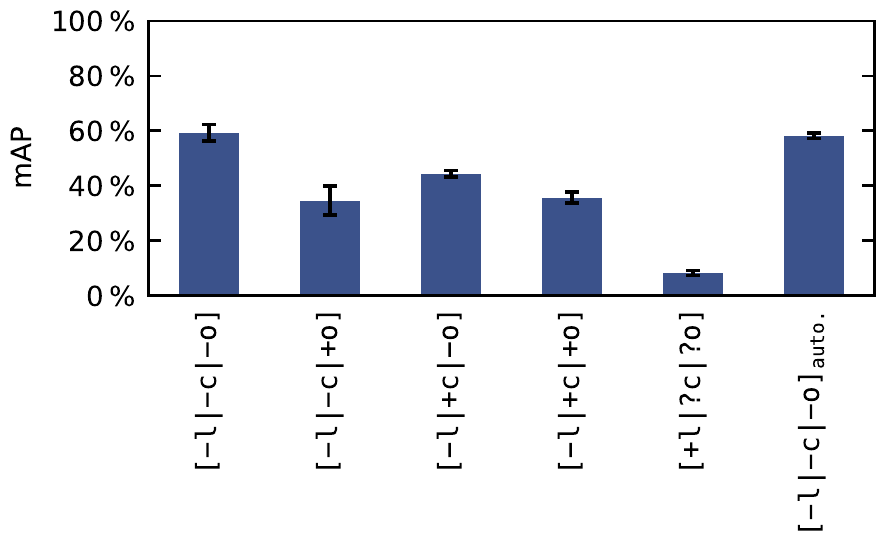}
	\caption{\Glsentryfullpl{mAP} of the fiber mask predictions of \frcnn/ for the different test data subsets.}
	\label{fig:Results-FiberMask}
\end{figure}

\graphicspath{{./figs/results/fibermask/}}
\cref{fig:Results-FiberMask} depicts the \glsentryshortpl{mAP} achieved by \frcnn/ for each of the real test data sets. Just as indicated by the example detections presented in \cref{sec:Results-DetectionQuality} (see \cref{fig:DetectionQuality}), loops (\texttt{[-l|-c|+o]}) pose the largest challenge to \frcnn/, while individual, isolated fibers (\texttt{[-l|-c|-o]} and \texttt{[-l|-c|-o]}$_\text{auto.}$) are much less problematic. Furthermore, the presence of clutter (\texttt{[-l|+c|-o]}) impedes the \gls{symb:meanAveragePrecision} less than the presence of overlaps (\texttt{[-l|-c|+o]} and \texttt{[-l|+c|+o]}).

\subsubsection{Fiber Width and Length Measurement}
\label{sec:Results-FiberWidthAndLengthMeasurement}
\graphicspath{{./figs/results/fiberwidth-and-length/}}
One way to evaluate the abilities of \frcnn/ for applications featuring the measurement of fiber widths and lengths, is to carry out an instance based accuracy assessment, i.e. that predictions errors of the fiber width and length are determined for each instance. To perform this analysis, it is necessary to match predicted instances with ground truth instances. As criterion for this matching process the bounding box \gls{IoU} was used, with an $\gls{symb:intersectionoverUnion} \ge 0.5$ indicating a match. After the matching, the percentage error \gls{symb:percentageError} for each match can be calculated:
\begin{equation}
\gls{symb:percentageError}{}_{,\gls{symb:index}}=\frac{\gls{symb:prediction}_{\gls{symb:index}}-\gls{symb:target}_{\gls{symb:index}}}{\gls{symb:target}_{\gls{symb:index}}} \cdot \SI{100}{\percent},
\end{equation}
where \gls{symb:index} is the instance index, \gls{symb:target} is the target value as determined via manual analysis and \gls{symb:prediction} is the prediction of \frcnn/.

To characterize the prediction errors across multiple instances, e.g. across a complete data set, the \gls{MAPE} can be used, which is defined as~\cite{Hyndman.2006}:
\begin{equation}
\gls{symb:meanAbsolutePercentageError} = \frac{1}{\gls{symb:numberOfDates}}\cdot\sum_{\gls{symb:index}=1}^{\gls{symb:numberOfDates}}|\gls{symb:percentageError} {}_{,\gls{symb:index}}|,
\end{equation}
where \gls{symb:index} is the instance index and \gls{symb:numberOfDates} is the number of dates, i.e. instances.

\begin{figure}
	\centering
	\includegraphics[width=\figurewidth]{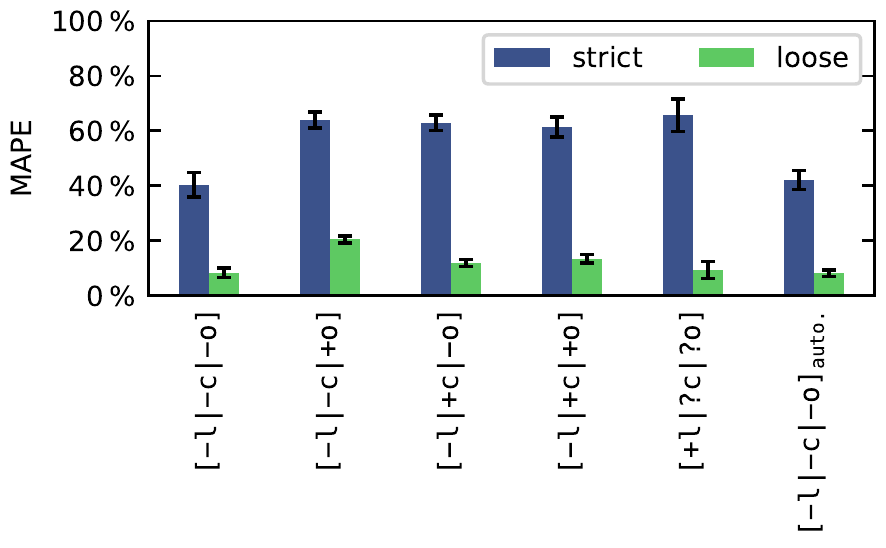}
	\caption{\Glsentryfullpl{MAPE} of the fiber width predictions of \frcnn/ for the different test data subsets.}
	\label{fig:Results-FiberWidth}
\end{figure}

\begin{figure}
	\centering
	\includegraphics[width=\figurewidth]{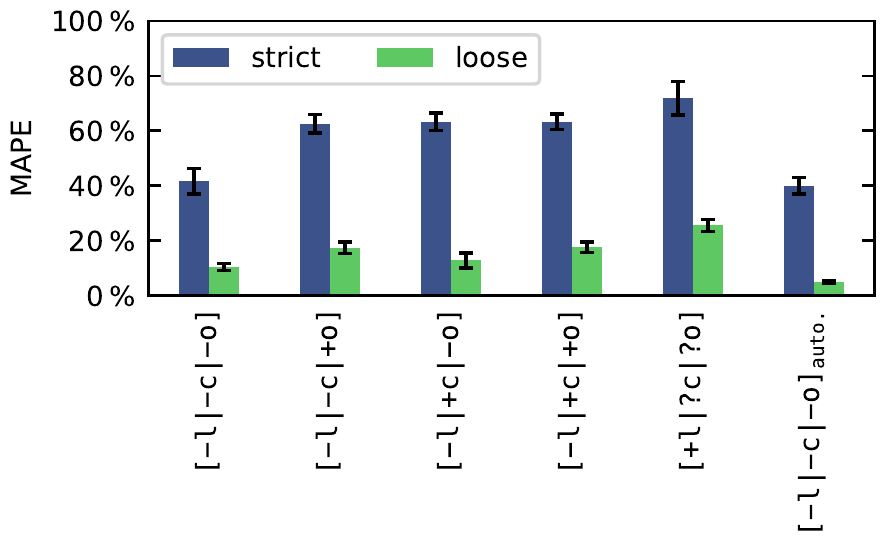}
	\caption{\Glsentryfullpl{MAPE} of the fiber length predictions of \frcnn/ for the different test data subsets.}
	\label{fig:Results-FiberLength}
\end{figure}

To reflect the requirements of different applications, two possibilities to handle non-matched instances were examined:
\begin{itemize}
	\item \emph{strict}: Non-matched instances were accounted for with $\gls{symb:percentageError} {}_{,\gls{symb:index}} = \SI{100}{\percent}$. This definition should be used for security-relevant applications and applications where the focus lies on the reliable detection of individual fibers (e.g. workplace risk assessments).
	\item \emph{loose}: Non-matched instances were accounted for with $\gls{symb:percentageError} {}_{,\gls{symb:index}} = \SI{0}{\percent}$. This definition can be used for non-security-relevant applications or applications where the fiber width or length distribution of an ensemble is of greater interest than the detection of individual fibers.
\end{itemize}
\cref{fig:Results-FiberWidth,fig:Results-FiberLength} show the \emph{strict} and \emph{loose} \glspl{MAPE} of \frcnn/ with respect to the fiber width and length prediction, respectively, for the different real data subsets. In general, the fiber width prediction is more accurate than the fiber length prediction. In agreement with the analysis of the fiber mask prediction accuracy (see \cref{sec:Results-DetectionQuality,sec:Results-MeanAveragePrecision}), also for the fiber width and length prediction, fibers featuring neither loops, clutter nor overlap (\texttt{[-l|-c|-o]} and \texttt{[-l|-c|-o]}$_\text{auto.}$) are least challenging. 
Interestingly, while loops (\texttt{[+l|?c|?o]}) are hardest for \frcnn/ with respect to the fiber mask and length prediction, they are much easier with respect to the fiber width prediction. This observation is indeed plausible, considering the fact that for a fiber with a constant width, a partial understanding of the fiber structure suffices to make a correct prediction, while for the correct fiber length prediction, it is necessary to understand the fiber's structure as a whole.

\begin{figure}
	\centering
	\includegraphics[width=\figurewidth]{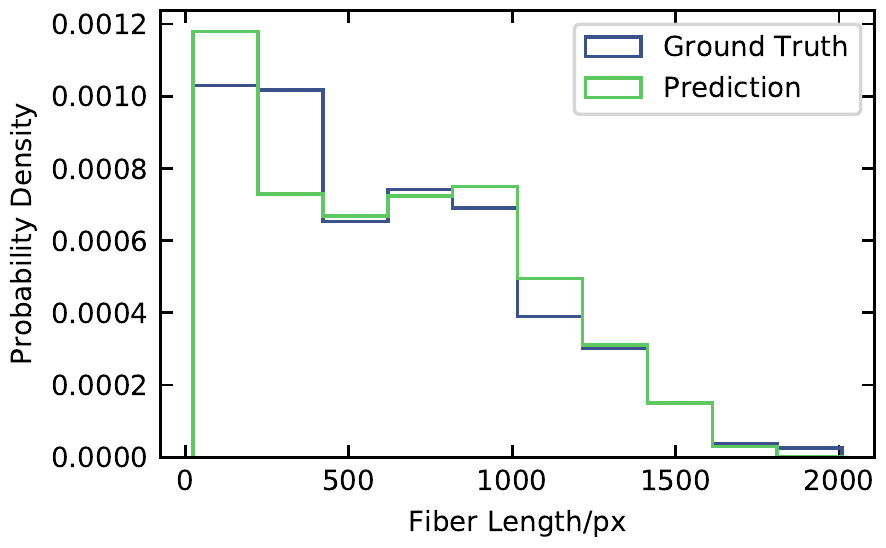}
	\caption{Probability density distributions of the fiber length predictions of \frcnn/ and the corresponding ground truths for an aggregation of all real test sets.}
	\label{fig:Results-FiberLengthDistribution}
\end{figure}

\begin{figure}
	\centering
	\includegraphics[width=\figurewidth]{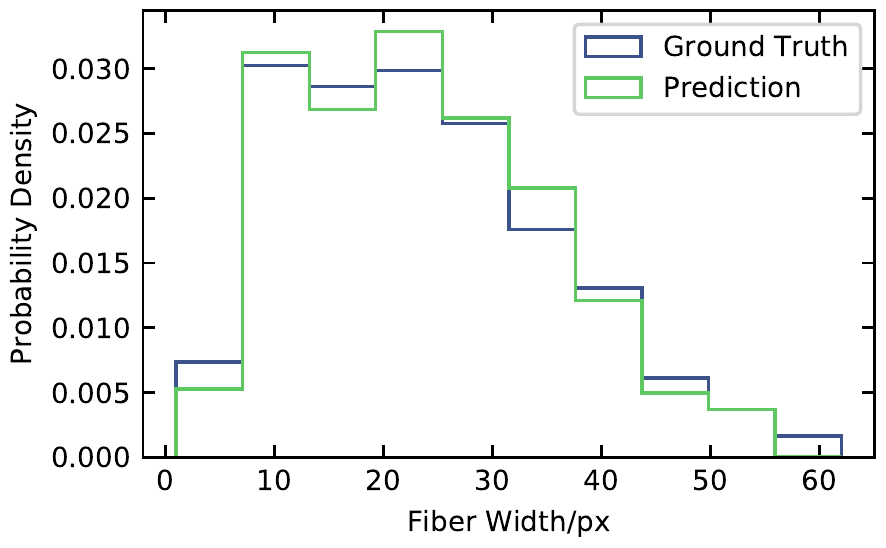}
	\caption{Probability density distributions of the fiber width predictions of \frcnn/ and the corresponding ground truths for an aggregation of all real test sets.}
	\label{fig:Results-FiberWidthDistribution}
\end{figure}

Another important factor with respect to the application of \frcnn/ to fiber width and length measurement tasks, is its ability to reconstruct the underlying length and width distributions of an ensemble of fibers. \cref{fig:Results-FiberLengthDistribution,fig:Results-FiberWidthDistribution} compare the fiber width and length distributions of the ensemble of all real test data sets to the respective predictions of \frcnn/.\footnote{For the histogram creation, each prediction was weighted with its objectness score, i.e. its certainty, as assigned by the \gls{ROI} extraction head of \frcnn/ (see \cref{sec:Method-RCNNs}).} Both the fiber width and the fiber length distribution are reconstructed accurately by \frcnn/, with no significant biases towards individual size classes. As indicated by the previous studies conducted within this section, the fiber width prediction is more reliable than the fiber length prediction.
To quantify the accuracy of the prediction of fiber width and length distributions by \frcnn/, the \name{Kullback}–\name{Leibler} divergence -- a common measure for the divergence of two probability distributions \gls{symb:probabilityDistributionP} and \gls{symb:probabilityDistributionQ} -- can be used~\cite{MacKay.2016}:
\begin{equation}
\label{eq:KullbackLeiblerDivergence}
\gls{symb:kullbackLeiblerDivergence}(\gls{symb:probabilityDistributionP} \parallel \gls{symb:probabilityDistributionQ}) = \sum_{x} \gls{symb:probabilityDistributionP}(x) \log\left(\frac{\gls{symb:probabilityDistributionP}(x)}{\gls{symb:probabilityDistributionQ}(x)}\right),
\end{equation}
where, for the case at hand, \gls{symb:probabilityDistributionQ} and \gls{symb:probabilityDistributionP} are the probability distributions of the predicted fiber widths (or lengths) and the associated ground truths, respectively.\footnote{For the calculation, bins where either $\gls{symb:probabilityDistributionP}(d)=0$ or $\gls{symb:probabilityDistributionQ}(d)=0$ were excluded.} The \name{Kullback}–\name{Leibler} divergence yields values which range from \num{0} for identical to \num{1} for completely diverging probability distributions.

For the probability distributions at hand, \cref{eq:KullbackLeiblerDivergence} yields the following \name{Kullback}–\name{Leibler} divergences:
\begin{itemize}
	\item Fiber width: $\gls{symb:kullbackLeiblerDivergence} = \num{0.006}$
	\item Fiber length: $\gls{symb:kullbackLeiblerDivergence} = \num{0.031}$
\end{itemize}
both of which indicate extremely high degrees of similarity and further support the hypothesis that \frcnn/ can predict fiber widths more reliably than fiber lengths.

	\graphicspath{{./figs/conclusion-and-outlook/}}
\section{Conclusion and Outlook}
\label{sec:ConclusionAndOutlook}
Within this publication, the well-known \mrcnn/ architecture was extended to yield an improved method for image-based fiber analysis. To do so, it was combined with a keypoint regression head, originally used for human pose estimation, to identify the \enquote{spine} of the analyzed fibers and supplemented with two novel heads for fiber width and fiber length predictions. 

For the training and validation of the new architecture, a large data set of more than \num{1650} annotated \gls{SEM} images, featuring approximately \num{2600} \glspl{CNT}, divided into six subsets of varying difficulty, was used. With semiautomatic annotation and image synthesis, two possibilities to supplement the manually annotated data -- or even avoid the laborious task of manual annotation as a whole -- were explored. Unfortunately, neither of these strategies did yield \glspl{AP} comparable to those achieved with manually annotated data. Improved methods for the creation of more realistic synthetic training data may therefore be explored in the future.

The design of the novel \frcnn/ architecture was optimized systematically, following an additive approach. Starting with a most basic implementation, the architecture was enhanced stepwise and reevaluated after each addition, based on the \glspl{AP} achieved on a test set of \num{401} fibers. The largest improvements were achieved by the introduction of a systematic ordering of ground truth keypoints during the training  and the utilization of input augmentation. While the introduction of keypoint pruning as an error correction mechanism resulted in a significant improvement, it was not as beneficial as initially anticipated. Therefore, the study of other error correction strategies is advisable for future research.

In the course of the architecture design, the \gls{symb:meanAveragePrecision} was improved considerably, compared to both the \frcnn/ and the \mrcnn/ baseline model.

To find possible weak spots and examine possible ways to improve \frcnn/ in the future, its \gls{symb:meanAveragePrecision} was determined for a number of test data sets, featuring different kinds of inhibiting factors (overlap, clutter and loops). The evaluation showed that loops and large amounts of fiber overlap are especially challenging for \frcnn/. Future research should therefore concentrate on these types of fibers.

Unlike \mrcnn/, \frcnn/ can not only be used for the instance segmentation of fiber images but also for the prediction of fiber width and length distributions. The evaluation, based on the complete set of available test data, yielded excellent reconstruction capabilities with respect to the underlying fiber width and length distributions, with the predicted fiber length and width probability distributions featuring only small deviations from the associated ground truths.

All in all, the \frcnn/ architecture provides an effective tool for automatic image-based fiber shape analysis. It is likely that future research concerning \frcnn/ in particular and \gls{R-CNN} architectures in general will improve its reliability and precision even further.
	
\section*{Acknowledgment}
The authors gratefully acknowledge the support via the project \name{"20226 N -- Deep Learning Particle Detection"} of the \name{DECHEMA} research foundation, which was funded by the \name{German Federation of Industrial Research Associations~(AiF)} within the program for \name{Industrial Corporate Research~(IGF)} of the \name{Federal Ministry for Economic Affairs and Energy~(BMWi)} based on a decision of the \name{German Bundestag}. All authors declare that they have no competing interests.

Special thank goes to the \name{Institute of Energy and Environmental Technology~(IUTA)} for providing the utilized \glsentrylong{SEM} images. 

No carbon dioxide was emitted due to the training of neural networks for this publication, thanks to the use of renewable energy.

	\glsaddallunused
	\printglossary[type=\acronymtype]
	\printglossary[type=symbolslist]
	\bibliographystyle{elsarticle-num}
	\bibliography{bibliography}
	\beginappendix
\onecolumn
\section*{Appendix}

\begin{algorithm*}[!htb]
	\caption{Keypoint pruning algorithm.}
	\label{app:alg:KeypointPruning}
	\begin{algorithmic}[1]
		\Require{\gls{var:keypoints}\Comment{fiber keypoints, as predicted by the keypoint regression head}}
		\Require{\gls{var:fiberwidth}\Comment{fiber width, as predicted by the fiber width regression head}}
		\Require{\gls{var:fiberlength}\Comment{fiber length, as predicted by the fiber length regression head}}
		\Require{\gls{var:mask}\Comment{fiber mask, as predicted by the mask segmentation head}}
		\Statex
		\Function{PruneKeypoints}{$\gls{var:keypoints}, \gls{var:fiberwidth}, \gls{var:fiberlength}, \gls{var:mask}$}
			\State $\gls{var:numberofkeypoints} \gets \Call{GetNumber}{\gls{var:keypoints}}$
			\State $\gls{var:splinemask} \gets \Call{GetSplineMask}{\gls{var:keypoints}, \gls{var:fiberwidth}}$\Comment{spline fit through \gls{var:keypoints} with width \gls{var:fiberwidth}}
			\State $\gls{var:iou} \gets \Call{CalculateIoU}{\gls{var:splinemask}, \gls{var:mask}}$\Comment{intersection over union}
			\State $\gls{var:splinelengtherror} \gets \Call{GetSplineLengthError}{\gls{var:keypoints}, \gls{var:fiberlength}}$\Comment{see \cref{alg:func:GetSplineLengthError}}
			\\
			\State $\gls{var:segments} \gets \Call{GetSegments}{\gls{var:keypoints}}$\Comment{list of pairs of adjacent keypoints}\label{alg:line:marker}
			\State $\gls{var:segments} \gets \Call{OrderSegmentsByLength}{\gls{var:segments}}$\Comment{misplaced keypoints cause long segments; should be checked first}
			\For{\gls{var:segment} in \gls{var:segments}}
				\For{\gls{var:keypoint} in \gls{var:segment}}
					\State $\gls{var:keypoints}\_new \gets \Call{RemoveKeypoint}{\gls{var:keypoint}, \gls{var:keypoints}}$
					\State $\gls{var:splinemask}\_new \gets \Call{GetSplineMask}{\gls{var:keypoints}\_new, \gls{var:fiberwidth}}$
					\State $\gls{var:iou}\_new \gets \Call{CalculateIoU}{\gls{var:splinemask}\_new, \gls{var:mask}}$
					\State $\gls{var:splinelengtherror}\_new \gets \Call{GetSplineLengthError}{\gls{var:keypoints}\_new, \gls{var:fiberlength}}$\Comment{see \cref{alg:func:GetSplineLengthError}}
					
					\If{$\gls{var:iou}\_new \ge \gls{var:iou}$ and $\gls{var:splinelengtherror}\_new \le \gls{var:splinelengtherror}$}\Comment{check for improvement}
					\State $\gls{var:keypoints} \gets \gls{var:keypoints}\_new$
					\State $\gls{var:splinemask} \gets \gls{var:splinemask}\_new$
					\State $\gls{var:iou} \gets \gls{var:iou}\_new$
					\State $\gls{var:splinelengtherror} \gets \gls{var:splinelengtherror}\_new$
					\State \Goto{alg:line:marker}\Comment{start over with improved keypoints}
					\EndIf
				\EndFor
			\EndFor
			\\
			\State $\gls{var:keypoints} \gets \Call{UniformSplineInterpolation}{\gls{var:keypoints}, \gls{var:numberofkeypoints}}$\Comment{restore original number of keypoints}
			\Return $\gls{var:keypoints}$
		\EndFunction
		\Statex
		\Function{GetSplineLengthError}{$\gls{var:keypoints}, \gls{var:fiberlength}$}\label{alg:func:GetSplineLengthError}
			\State $\gls{var:splinelength} \gets \Call{GetSplineLength}{\gls{var:keypoints}}$\Comment{integrate length of spline fit through \gls{var:keypoints}}         
			\Return $\left\rvert1-\frac{\gls{var:splinelength}}{\gls{var:fiberlength}}\right\rvert$
		\EndFunction
	\end{algorithmic}
\end{algorithm*}

\begin{table}
	\centering
	\caption{Relevant hardware of the utilized \glsentryshort{GPU} server.}	
	\label{app:tab:Hardware}
	\begin{tabularx}{\figurewidth}{Yc}
		\toprule\toprule
		Mainboard  &              Supermicro X11DPG-QT              \\
		CPU        &            2 $\times$ Intel Xeon Gold 5118            \\
		GPU        &         4 $\times$ NVIDIA GeForce RTX 2080 Ti         \\
		RAM        &  12 $\times$ \SI{8}{\giga\byte} DDR4 PC2666 ECC reg.  \\
		SSD (OS)   & Micron SSD 5100 PRO \SI{960}{\giga\byte}, SATA \\
		SSD (data) &  Samsung SSD 960 EVO \SI{1}{\tera\byte}, M.2   \\ \bottomrule
	\end{tabularx}
\end{table}

\begin{table}
	\centering
	\caption{Relevant software of the utilized \glsentryshort{GPU} server.}	
	\label{app:tab:Software}
	\begin{tabularx}{\figurewidth}{YZ}
		\toprule\toprule
		OS (host)        & Ubuntu 18.04.3 LTS \\
		OS (docker)      & Ubuntu 18.04.3 LTS \\
		Python (docker)  & 3.6.9              \\
		PyTorch (docker) & 1.4.0              \\ \bottomrule
	\end{tabularx}
\end{table}%

\end{document}